\documentclass[12pt]{article}
\usepackage[utf8]{inputenc}
\usepackage[svgnames]{xcolor}
\usepackage{jmlr2e}
\usepackage[a4paper, margin=1.5cm, bottom=1in]{geometry}
\usepackage{tikz}
\usepackage[framemethod=tikz]{mdframed}
\usepackage[ruled,vlined]{algorithm2e}
\usepackage{amsmath}
\usepackage{bm}
\usepackage{amssymb}
\usepackage{amsfonts}
\usetikzlibrary{bayesnet,arrows,shapes,automata,backgrounds,petri,positioning}
\usetikzlibrary{decorations.pathmorphing}
\usetikzlibrary{decorations.shapes}
\usetikzlibrary{decorations.text}
\usetikzlibrary{decorations.fractals}
\usetikzlibrary{decorations.footprints}
\usetikzlibrary{decorations.pathreplacing}
\usetikzlibrary{shadows}
\usetikzlibrary{calc}
\usetikzlibrary{spy}
\usepackage{multirow}
\usepackage{graphicx}
\usepackage{booktabs}
\usepackage{array}
\usepackage{pifont}
\usepackage{natbib}
\usepackage{stackengine}
\usepackage{subfigure}
\usepackage{float}

\colorlet{questionColor}{DarkOrange}
\colorlet{exampleColor}{Silver}
\colorlet{answerColor}{LimeGreen}
\colorlet{commentColor}{DodgerBlue}
\colorlet{remarkColor}{FireBrick}
\colorlet{proofColor}{MediumPurple}
\def\questionSymbol{{\textbf{Q}}}
\def\exampleSymbol{{\textbf{E}}}
\def\answerSymbol{{\textbf{A}}}
\def\commentSymbol{{\textbf{C}}}
\def\remarkSymbol{{\textbf{R}}}
\def\proofSymbol{{\textbf{P}}}
\let\svendmdframed\endmdframed
\makeatletter
\def\endmdframed{\svendmdframed\unskip}
\makeatother

\def\delequal{\mathrel{\ensurestackMath{\stackon[1pt]{=}{\scriptstyle\Delta}}}}

\tikzset{commentsymbol/.style={rectangle, draw=commentColor, text=commentColor, fill=white, scale=1, overlay}}
\tikzset{examplesymbol/.style={rectangle, draw=exampleColor, text=exampleColor, fill=white, scale=1, overlay}}
\tikzset{answersymbol/.style={rectangle, draw=answerColor, text=answerColor, fill=white, scale=1, overlay}}
\tikzset{questionsymbol/.style={rectangle, draw=questionColor, text=questionColor, fill=white, scale=1, overlay}}
\tikzset{remarksymbol/.style={rectangle, draw=remarkColor, text=remarkColor, fill=white, scale=1, overlay}}
\tikzset{proofsymbol/.style={rectangle, draw=proofColor, text=proofColor, fill=white, scale=1, overlay}}

\mdfdefinestyle{answer}{
hidealllines=true, leftline=true, skipabove=0pt, skipbelow=0pt, innertopmargin=0.7em, innerbottommargin=0.7em,
innerrightmargin=0.7em, rightmargin=0.7em, innerleftmargin=1.4em, leftmargin=0.7em, middlelinewidth=.2em,
linecolor=answerColor, fontcolor=black, backgroundcolor=answerColor!15,
firstextra={\path let \p1=(P), \p2=(O) in (\x2,0.5 * \y1) node[answersymbol] {\answerSymbol};},
secondextra={\path let \p1=(P), \p2=(O) in (\x2,0.5 * \y1) node[answersymbol] {\answerSymbol};},
middleextra={\path let \p1=(P), \p2=(O) in (\x2,0.5 * \y1) node[answersymbol] {\answerSymbol};},
singleextra={\path let \p1=(P), \p2=(O) in (\x2,0.5 * \y1) node[answersymbol] {\answerSymbol};}
}

\mdfdefinestyle{comment}{
hidealllines=true, leftline=true, skipabove=0pt, skipbelow=0pt, innertopmargin=0.7em, innerbottommargin=0.7em,
innerrightmargin=0.7em, rightmargin=0.7em, innerleftmargin=1.4em, leftmargin=0.7em, middlelinewidth=.2em,
linecolor=commentColor, fontcolor=black, backgroundcolor=commentColor!15,
firstextra={\path let \p1=(P), \p2=(O) in (\x2,0.5 * \y1) node[commentsymbol] {\commentSymbol};},
secondextra={\path let \p1=(P), \p2=(O) in (\x2,0.5 * \y1) node[commentsymbol] {\commentSymbol};},
middleextra={\path let \p1=(P), \p2=(O) in (\x2,0.5 * \y1) node[commentsymbol] {\commentSymbol};},
singleextra={\path let \p1=(P), \p2=(O) in (\x2,0.5 * \y1) node[commentsymbol] {\commentSymbol};}
}

\mdfdefinestyle{question}{
hidealllines=true, leftline=true, skipabove=0pt, skipbelow=0pt, innertopmargin=0.7em, innerbottommargin=0.7em,
innerrightmargin=0.7em, rightmargin=0.7em, innerleftmargin=1.4em, leftmargin=0.7em, middlelinewidth=.2em,
linecolor=questionColor, fontcolor=black, backgroundcolor=questionColor!15,
firstextra={\path let \p1=(P), \p2=(O) in (\x2,0.5 * \y1) node[questionsymbol] {\questionSymbol};},
secondextra={\path let \p1=(P), \p2=(O) in (\x2,0.5 * \y1) node[questionsymbol] {\questionSymbol};},
middleextra={\path let \p1=(P), \p2=(O) in (\x2,0.5 * \y1) node[questionsymbol] {\questionSymbol};},
singleextra={\path let \p1=(P), \p2=(O) in (\x2,0.5 * \y1) node[questionsymbol] {\questionSymbol};}
}

\mdfdefinestyle{remark}{
hidealllines=true, leftline=true, skipabove=0pt, skipbelow=0pt, innertopmargin=0.7em, innerbottommargin=0.7em,
innerrightmargin=0.7em, rightmargin=0.7em, innerleftmargin=1.4em, leftmargin=0.7em, middlelinewidth=.2em,
linecolor=remarkColor, fontcolor=black, backgroundcolor=remarkColor!15,
firstextra={\path let \p1=(P), \p2=(O) in (\x2,0.5 * \y1) node[remarksymbol] {\remarkSymbol};},
secondextra={\path let \p1=(P), \p2=(O) in (\x2,0.5 * \y1) node[remarksymbol] {\remarkSymbol};},
middleextra={\path let \p1=(P), \p2=(O) in (\x2,0.5 * \y1) node[remarksymbol] {\remarkSymbol};},
singleextra={\path let \p1=(P), \p2=(O) in (\x2,0.5 * \y1) node[remarksymbol] {\remarkSymbol};}
}

\mdfdefinestyle{example}{
hidealllines=true, leftline=true, skipabove=0pt, skipbelow=0pt, innertopmargin=0.7em, innerbottommargin=0.7em,
innerrightmargin=0.7em, rightmargin=0.7em, innerleftmargin=1.4em, leftmargin=0.7em, middlelinewidth=.2em,
linecolor=exampleColor, fontcolor=black, backgroundcolor=exampleColor!15,
firstextra={\path let \p1=(P), \p2=(O) in (\x2,0.5 * \y1) node[examplesymbol] {\exampleSymbol};},
secondextra={\path let \p1=(P), \p2=(O) in (\x2,0.5 * \y1) node[examplesymbol] {\exampleSymbol};},
middleextra={\path let \p1=(P), \p2=(O) in (\x2,0.5 * \y1) node[examplesymbol] {\exampleSymbol};},
singleextra={\path let \p1=(P), \p2=(O) in (\x2,0.5 * \y1) node[examplesymbol] {\exampleSymbol};}
}

\mdfdefinestyle{proof}{
hidealllines=true, leftline=true, skipabove=0pt, skipbelow=0pt, innertopmargin=0.7em, innerbottommargin=0.7em,
innerrightmargin=0.7em, rightmargin=0.7em, innerleftmargin=1.4em, leftmargin=0.7em, middlelinewidth=.2em,
linecolor=proofColor, fontcolor=black, backgroundcolor=proofColor!15,
firstextra={\path let \p1=(P), \p2=(O) in (\x2,0.5 * \y1) node[proofsymbol] {\proofSymbol};},
secondextra={\path let \p1=(P), \p2=(O) in (\x2,0.5 * \y1) node[proofsymbol] {\proofSymbol};},
middleextra={\path let \p1=(P), \p2=(O) in (\x2,0.5 * \y1) node[proofsymbol] {\proofSymbol};},
singleextra={\path let \p1=(P), \p2=(O) in (\x2,0.5 * \y1) node[proofsymbol] {\proofSymbol};}
}

\makeatletter
\g@addto@macro\bfseries{\boldmath}
\makeatother

\newcommand{\kl}[2]{D_{\mathrm{KL}} \left[ \left. \left. #1 \right|\right| #2 \right] }
\DeclareMathOperator*{\argmin}{arg\,min}

\makeatletter
\newcommand*\bigcdot{\mathpalette\bigcdot@{.5}}
\newcommand*\bigcdot@[2]{\mathbin{\vcenter{\hbox{\scalebox{#2}{$\m@th#1\bullet$}}}}}
\makeatother

\def\pgf@sh@@knotanchor#1#2{%
	\anchor{#2 north west}{%
		\csname pgf@anchor@knot #1@north west\endcsname%
		\pgf@x=#2\pgf@x%
		\pgf@y=#2\pgf@y%
	}%
	\anchor{#2 north east}{%
		\csname pgf@anchor@knot #1@north east\endcsname%
		\pgf@x=#2\pgf@x%
		\pgf@y=#2\pgf@y%
	}%
	\anchor{#2 south west}{%
		\csname pgf@anchor@knot #1@south west\endcsname%
		\pgf@x=#2\pgf@x%
		\pgf@y=#2\pgf@y%
	}%
	\anchor{#2 south east}{%
		\csname pgf@anchor@knot #1@south east\endcsname%
		\pgf@x=#2\pgf@x%
		\pgf@y=#2\pgf@y%
	}%
	\anchor{#2 north}{%
		\csname pgf@anchor@knot #1@north\endcsname%
		\pgf@x=#2\pgf@x%
		\pgf@y=#2\pgf@y%
	}%
	\anchor{#2 east}{%
		\csname pgf@anchor@knot #1@east\endcsname%
		\pgf@x=#2\pgf@x%
		\pgf@y=#2\pgf@y%
	}%
	\anchor{#2 west}{%
		\csname pgf@anchor@knot #1@west\endcsname%
		\pgf@x=#2\pgf@x%
		\pgf@y=#2\pgf@y%
	}%
	\anchor{#2 south}{%
		\csname pgf@anchor@knot #1@south\endcsname%
		\pgf@x=#2\pgf@x%
		\pgf@y=#2\pgf@y%
	}%
}

\begin{document}

\title{Structure learning with Temporal Gaussian Mixture for model-based Reinforcement Learning}

\author{\name Théophile Champion \email txc314@student.bham.ac.uk \\
       \addr University of Birmingham, School of Computer Science,\\
       Birmingham B15 2TT, United Kingdom
       \AND
       \name Marek Grze\'s \email m.grzes@kent.ac.uk \\
       \addr University of Kent, School of Computing\\
       Canterbury CT2 7NZ, United Kingdom
       \AND
       \name Howard Bowman \email h.bowman@bham.ac.uk \\
       \addr University of Birmingham, School of Psychology and School of Computer Science,\\
       Birmingham B15 2TT, United Kingdom\\
       University College London, Wellcome Centre for Human Neuroimaging (honorary)\\
       London WC1N 3AR, United Kingdom
}
       
\editor{\textbf{TO BE FILLED}} %TODO

\maketitle

\begin{abstract} %
Model-based reinforcement learning refers to a set of approaches capable of sample-efficient decision making, which create an explicit model of the environment. This model can subsequently be used for learning optimal policies. In this paper, we propose a temporal Gaussian Mixture Model composed of a perception model and a transition model. The perception model extracts discrete (latent) states from continuous observations using a variational Gaussian mixture likelihood. Importantly, our model constantly monitors the collected data searching for new Gaussian components, i.e., the perception model performs a form of structure learning \citep{structure_learning_AI,friston2018bayesian,neacsu2022structure} as it learns the number of Gaussian components in the mixture. Additionally, the transition model learns the temporal transition between consecutive time steps by taking advantage of the Dirichlet-categorical conjugacy. Both the perception and transition models are able to forget part of the data points, while integrating the information they provide within the prior, which ensure fast variational inference. Finally, decision making is performed with a variant of Q-learning which is able to learn Q-values from beliefs over states. Empirically, we have demonstrated the model's ability to learn the structure of several mazes: the model discovered the number of states and the transition probabilities between these states. Moreover, using its learned Q-values, the agent was able to successfully navigate from the starting position to the maze's exit.
\end{abstract}
\vspace{0.5cm}

\section{Introduction}

Model-based reinforcement learning is a theory describing how an agent should interact with its environment. More precisely, the agent maintains a model of its environment, which is composed of observations, states, and actions. When new observations become available, the agent needs to estimate the most likely states of the environment. This process is generally referred to as perception or inference, and can be implemented by minimizing the variational free energy, also known as the negative evidence lower bound in machine learning \citep{blei2017variational}.

Variational inference is a form of approximate inference, where the true posterior is approximated by the variational distribution. Specifically, inference is generally rendered tractable by restricting the approximate posterior to a class of distributions, where each distribution corresponds to a different set of parameter values. Inference then refers to the optimization of these parameters such that the variational free energy is minimized. 

An important family of distributions frequently used in variational inference is the exponential family \citep{holland1981exponential}, which contains most of the well-known distributions. For example, the Gaussian, Bernoulli, and Dirichlet distributions are all members of this family. Importantly, all the distributions within the exponential family can be mapped to the same functional form, which can be used to derive modular inference algorithms such as variational message passing \citep{VMP_TUTO}.

In these previous approaches, the generative model's structure is known in advance, i.e., the number of latent states is known by the agent. In practice, this means that the model needs to be designed by a domain expert and is not learned by the agent. However, for some applications, the model may not be available, or experts may only have a partial understanding of the problem. In this case, learning the model becomes essential. Recent work has focused on problems where observations are discrete \citep{structure_learning_AI,friston2018bayesian,neacsu2022structure}. In contrast, this paper addresses the case where observations are continuous.

Throughout this paper, we assume that the data is distributed according to a mixture of multivariate Gaussian distributions. Then, each component of the mixture is associated with a latent state, and we aim to learn how these states change over time. The key challenge is to identify the correct number of components. This requires the model to add and remove components when necessary. While the Gaussian Mixture Model (GMM) is able to prune superfluous components, we also need a way to increase the number of components as new clusters of data points are discovered. To this end, our agent constantly monitors the data for new clusters of data points, and new Gaussian components are added to the mixtures when such clusters are identified.

While learning the model structure and inferring the current state using variational inference is useful, it does not prescribe which actions should be taken. Thus, the next step consists of comparing the quality of different policies, where in reinforcement learning, the quality of a policy is the discounted sum of rewards. Ideally, we would like to compare all possible policies, and select the policy with the highest quality. Unfortunately, the number of valid policies grows exponentially with the time horizon of planning. Specifically, if the agent can choose one of $A$ actions, for $T$ time steps, then the number of possible policies is: $A^T$.

This exponential explosion renders an exhaustive search intractable. Instead, algorithms such as Monte Carlo tree search \citep{browne2012survey} can be used to efficiently explore the space of valid policies \citep{BTAI_empirical,CHAMPION2022295,BTAI_3MF,BTAI_BF}. An alternative approach is the Q-learning algorithm \citep{Mnih2015}, where instead of searching the space of policies using a search tree, the agent learns the value of performing each action in each state. However, Q-learning requires states to be observable. Therefore, we propose a variant of the Q-learning algorithm which can learn from beliefs over states.

In section \ref{sec:perception}, we present the variational Gaussian mixture model, which is used to learn the number of components and perform inference of the latent variables. Section \ref{sec:transition} explains how the temporal transition can be learned by taking advantage of a categorical-Dirichlet model. Then, in Section \ref{sec:forgetting}, we describe an optimization which allows the agent to forget part of the data points by integrating the information they provided into the prior. By forgetting part of the data, the agent reduces the amount of memory required, while speeding up the inference process. Next, in Section \ref{sec:planning}, we describe how Q-learning can be adapted to work with beliefs over states. Finally, in Section \ref{sec:experiments} we empirically validate our approach, before summarizing and concluding the paper in Section \ref{sec:conclusion}.

\section{Perception Model} \label{sec:perception}

In this section, we discuss the theory and implementation details of our perception model. Specifically, we describe how mean shift clustering can be used to initialize the parameters of a variational Gaussian mixture (VGM) model. Then, we describe the VGM generative model, variational distribution and update equations.

\subsection{Mean Shift Clustering}

\begin{figure}[H]
	\centering
	\subfigure[]{
		\begin{tikzpicture}[scale=1,spy using outlines]
			\node (image) {\includegraphics[width=0.40\textwidth]{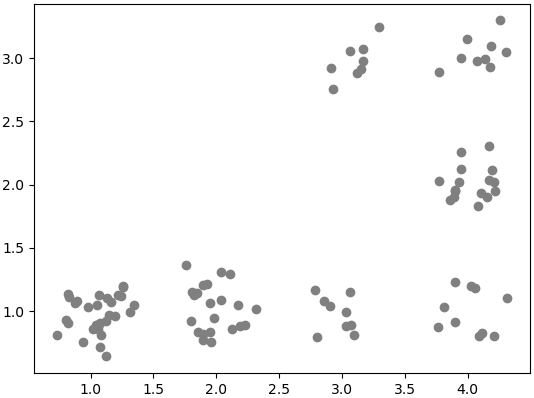}};
			
			\coordinate (A1) at (-1.185, -0.97);
			\coordinate (A2) at (-0.9285, -1.177);
			\coordinate (B1) at (-0.9, -1.2);
			\coordinate (B2) at (-0.765, -1.425);
			
			\draw [line width=0.3mm, -stealth, red!40!black] (A1) -- (A2);
			\draw [line width=0.3mm, -stealth, red!60!black] (B1) -- (B2);

			\draw[thick, red!40!black, opacity=1] (-1.185, -0.97) circle (20pt);
			\filldraw[thick, red!40!black, opacity=1] (-1.185, -0.97) circle (1pt);
			
			\filldraw[thick, red!60!black, opacity=1] (-0.9, -1.2) circle (1pt);
			\draw[thick, red!60!black, opacity=1] (-0.9, -1.2) circle (20pt);
			
			\filldraw[thick, red!80!black, opacity=1] (-0.75, -1.45) circle (1pt);
			\draw[thick, red!80!black, opacity=1] (-0.75, -1.45) circle (20pt);
						
			\spy [blue,draw,height=4.3cm,width=4.8cm,magnification=1.8,connect spies] on ($(B1)$) in node at ($(B1) + (0,3.5)$);

		\end{tikzpicture}\label{fig:mean_shift_1}} 
	\subfigure[]{\includegraphics[width=0.40\textwidth]{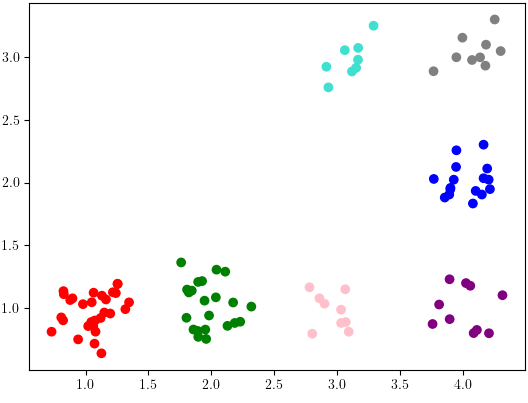}\label{fig:mean_shift_2}}
	\vspace{-0.25cm}
	\caption{(a) shows how the centroids are initialized and updated, while (b) illustrates how the data points are clustered (different colors indicate different clusters).}
	\label{fig:k_means}
\end{figure}
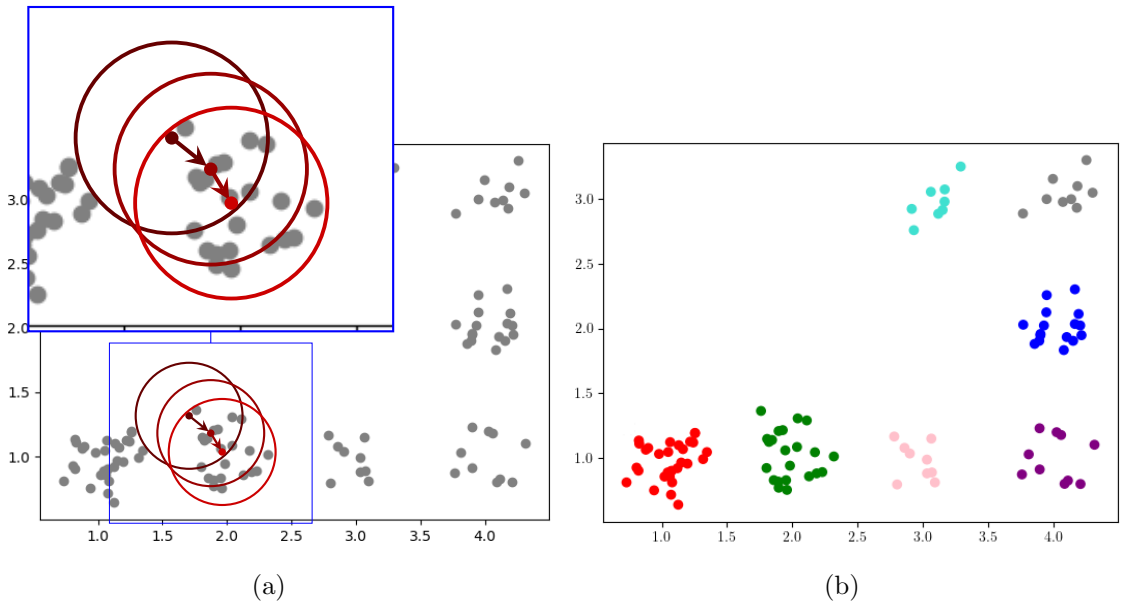

Given a dataset $\bm{X} = \{\bm{x}_1, ..., \bm{x}_N\}$, the mean shift algorithm \citep{carreira2015review} aims to find clusters of data points. While the mean shift algorithm does not require the number of clusters to be known, the user must specify the bandwidth parameter $\theta_b$, which defines the window's radius (see below). For each data point $\bm{x}_n$, the mean shift algorithm initializes a spherical window of radius $\theta_b$ centred at $\bm{x}_n$, then the window's centroid $\bm{m}_n$ is set to the mean of the data points within the window, i.e.,
\begin{align}
	\bm{m}_n = \frac{1}{|\mathcal{W}_{\theta_b}|} \sum_{n \in \mathcal{W}_{\theta_b}} \bm{x}_n, \label{eq:mean-shift}
\end{align}
where $\mathcal{W}_{\theta_b}$ is the set of data point indices within the window of radius $\theta_b$, and $|\mathcal{W}_{\theta_b}|$ corresponds to the number of data points in this window. The above update equation is iterated until convergence of the window's centroid position, at which point the window is centred on a part of the space with high density. The points whose window's centroid land close to each other are then grouped within the same cluster. This procedure will output $K$ clusters, we let $\bm{\hat{r}}_{nk}$ be one if the $n$-th data point is associated to the $k$-th cluster, and zero otherwise. Figure \ref{fig:k_means} illustrates the mean shift algorithm. 

\subsection{Variational Gaussian mixture} \label{ssec:gm}

While the mean-shift algorithm can identify clusters of data points, this algorithm does not scale well with the number of data points, i.e., for each additional data point Equation \eqref{eq:mean-shift} must be iterated until convergence. Thus, we need to find a way to discard data points without losing the information they contain. As will become apparent later, this can be achieved using a variational Gaussian mixture (VGM) model \citep{bishop2006pattern}. In the following sub-sections, we discuss the generative model, variational distribution, variational free energy and update equations of the variational Gaussian mixture.

\subsubsection{Generative model} \label{sssec:generative_model}

The generative model assumes that each data point $\bm{x}_n$ is sampled from one of $K$ Gaussian components. For each data point, $\bm{z}_{nk}$ is a binary random variable equal to one if the $k$-th component is responsible for the $n$-th data point, and zero otherwise. We denote by $\bm{z}_{n}$ the one-of-$K$ (binary) random vector (i.e., a vector containing binary random variables) describing which component is responsible for the $n$-th data point, and let $\bm{Z} = \{\bm{z}_1, ..., \bm{z}_N\}$ be the set of random vectors associated with all the data points. 

\begin{figure}[H]
	\begin{center}
		\begin{tikzpicture}[scale=1]
			\node[latent, yshift=0.5cm] (D) {$\bm{D}$};
			\node[latent, below=of D, yshift=0.5cm] (Z) {$\bm{Z}$};
			\node[obs, below=of Z, yshift=0.5cm] (X) {$\bm{X}$};
			\node[latent, right=of Z, xshift=-0.5cm] (lambda) {$\bm{\Lambda}$};
			\node[latent, below=of lambda, yshift=0.5cm] (mu) {$\bm{\mu}$};
			\edge {D} {Z}
			\edge {Z} {X}
			\edge {lambda} {mu}
			\edge {lambda} {X}
			\edge {mu} {X}
		\end{tikzpicture}
	\end{center}
	\vspace{-0.5cm}
	\caption[Variational Gaussian Mixture]{This figure illustrates the graphical model of the variational Gaussian mixture. The latent variables are represented by a white circle with the variable's name at the center, and the observed variables are depicted similarly but with a gray background. Finally, arrows connect the predictor variables to their target variable.}
	\label{fig:vgm_gm}
\end{figure}
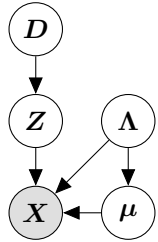

\begin{figure}[H]
	\begin{center}
		\includegraphics[width=0.40\textwidth]{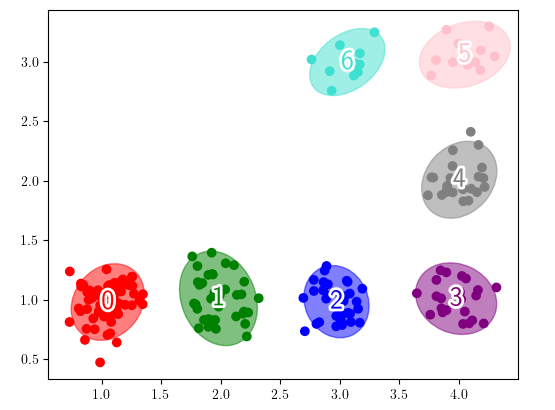}
	\end{center}
	\vspace{-0.5cm}
	\caption[Variational Gaussian Mixture Prior]{This figure illustrates the variational Gaussian mixture prior, when its parameters are initialized using the mean shift algorithm.}
	\label{fig:mean_shift_prior}
\end{figure}

Additionally, each Gaussian component is parameterized by its mean vector $\bm{\mu}_k$ and precision matrix $\bm{\Lambda}_k$. For conciseness, the set of all mean vectors and precision matrices are denoted by $\bm{\mu}$ and $\bm{\Lambda}$, respectively. Using the aforementioned notation, the probability of the data (i.e., $\bm{X}$), given the latent variables (i.e., $\bm{Z}$) and the component parameters (i.e., $\bm{\mu}$ and $\bm{\Lambda}$) is:
\begin{align}
	P(\bm{X} | \bm{Z}, \bm{\mu}, \bm{\Lambda}) &= \prod_{n=1}^N \prod_{k=1}^K \mathcal{N}(\bm{x}_n; \bm{\mu}_k, \bm{\Lambda}_k^{-1})^{\bm{z}_{nk}}.	\label{eq:likelihood_factor}
\end{align}
Moreover, the prior over latent variables is defined as a categorical distribution parameterized by the mixing coefficients $\bm{D}$, where $\bm{D}_k$ is the fraction of the data for which the $k$-th component is responsible, i.e., the larger $\bm{D}_k$ is, the more prevalent the $k$-th component will be. More formally:
\begin{align}
	P(\bm{Z} | \bm{D}) &= \prod_{n=1}^N \text{Cat}(\bm{z}_n ; \bm{D}). \label{eq:Z_prior}
\end{align}
Finally, the VGM model takes advantage of conjugate priors by defining the prior over $\bm{D}$, $\bm{\mu}$ and $\bm{\Lambda}$ as a Dirichlet, Gaussian and Wishart distributions, respectively. More precisely:
\begin{align}
	P(\bm{\mu} | \bm{\Lambda}) &= \prod_{k=1}^K \mathcal{N}(\bm{\mu}_k; \bm{m}_k, (\beta_k\bm{\Lambda}_k)^{-1}), \label{eq:mu_prior} \\
	P(\bm{\Lambda}) &= \prod_{k=1}^K \mathcal{W}(\bm{\Lambda}_k; \bm{W}_k, v_k), \label{eq:lambda_prior} \\
	P(\bm{D}) &= \text{Dir}(\bm{D}; d), \phantom{\prod^K} \label{eq:D_prior}
\end{align}
where $d$ is a vector containing concentration parameters of the prior over $\bm{D}$, $\bm{W}_k$ and $v_k$ are the scale matrix and degrees of freedom of the Wishart distribution over $\bm{\Lambda}_k$, $\bm{m}_k$ is the mean vector of the prior over $\bm{\mu}_k$, and finally, $\beta_k$ is a parameter scaling the values of the precision matrix $\bm{\Lambda}_k$. Putting together all the factors defined in Equations \eqref{eq:likelihood_factor} to \eqref{eq:D_prior}, we see that the generative model factorizes as follows:
\begin{align}
	P(\bm{X}, \bm{Z}, \bm{\mu}, \bm{\Lambda}, \bm{D}) = P(\bm{Z} | \bm{D}) P(\bm{D}) P(\bm{\mu} | \bm{\Lambda}) P(\bm{\Lambda}) P(\bm{X} | \bm{Z}, \bm{\mu}, \bm{\Lambda}), \label{eq:gm_distribution_def}
\end{align}
which corresponds to the graphical model of Figure \ref{fig:vgm_gm}. Note, the VGM is parameterized by: $d$, $\bm{W}_k$, $v_k$, $\beta_k$, and $\bm{m}_k$, but \textit{how should we choose these parameters?} Experimentally, we set the parameters $d_k$ and $\beta_k$ to two times the number of states, i.e., $2K$. Additionally, we initialized $v_k$ to $2K + O - 0.99$, where $O$ is the dimensionality of the observations, i.e., $\bm{x}_n \in \mathbb{R}^O$. Moreover, $\bm{m}_k$ was set to the mean of the data points associated to the $k$-th component by the mean shift algorithm. Finally, according to Appendix B of \citep{bishop2006pattern}, the expected precision matrix under a Wishart distribution $P(\bm{\Lambda})$ is $\bm{W}_kv_k$, thus we let: 
\begin{align}
	\bm{W}_k = \frac{\bar{\bm{\Lambda}}_k}{v_k},
\end{align}
where $\bar{\bm{\Lambda}}_k$ is the empirical precision matrix of the $k$-th Gaussian component. The empirical precision matrix was computed by taking the inverse of the (empirical) covariance matrix. Figure \ref{fig:mean_shift_prior} illustrates the VGM obtained when its parameters are initialized based on the mean shift clusters using the aforementioned procedure.

\subsubsection{Empirical prior and variational distribution} \label{sss:ep_and_vd}

Now that the generative model has been laid out, we would like to take into account the information provided by the available data points. As mentioned previously, we want to forget part of the data points as time passes, thus, we assume that the dataset $\bm{X} = \{\bm{x}_1, ..., \bm{x}_N\}$ has been split into the data points to forget $\bm{X}^{'} = \{\bm{x}_i \,|\, i \in N^{'}\}$ and data points to keep $\bm{X}^{''} = \{\bm{x}_j \,|\, j \in N^{''}\}$, where $N^{'}$ and $N^{''}$ are the indices of the data points to forget and keep, respectively.

For now, we assume that $N^{'}$ and $N^{''}$ are given, and the process of constructing $N^{'}$ and $N^{''}$ will be detailed in Section \ref{sec:forgetting}. Since we want to forget the data points indexed by $\bm{X}^{'}$, we compute an empirical prior based on these data points alone. The empirical prior is a posterior distribution conditioned on the data points indexed by $\bm{X}^{'}$. Critically, because of conjugacy, the empirical prior has the same functional form as the prior, and it can be used as a prior when additional data points will become available.

As computing the exact posterior is intractable, we turn to the variational inference framework \citep{VI_TUTO}. More precisely, we approximate the true posterior using a structured mean-field approximation:
\begin{align}
	P(\bm{Z}^{'}, \bm{\mu}, \bm{\Lambda}, \bm{D} | \bm{X}^{'}) &\approx E(\bm{Z}^{'}, \bm{\mu}, \bm{\Lambda}, \bm{D}) \delequal E(\bm{Z}^{'}) E(\bm{D}) \prod_{k=1}^K E(\bm{\mu}_k | \bm{\Lambda}_k) E(\bm{\Lambda}_k),
\end{align}
where $\bm{Z}^{'}$ are the latent variables corresponding to the data points to forget, and to ensure conjugacy, the individual factors are defined as follows:
\begin{align}
	E(\bm{\mu} | \bm{\Lambda}) &= \prod_{k=1}^K \mathcal{N}(\bm{\mu}_k; \bm{\bar{m}}_k, (\bar{\beta}_k\bm{\Lambda}_k)^{-1}), \\
	E(\bm{\Lambda}) &= \prod_{k=1}^K \mathcal{W}(\bm{\Lambda}_k; \bm{\bar{W}}_k, \bar{v}_k), \\
	E(\bm{Z}^{'}) &= \prod_{n \in N^{'}} \text{Cat}(\bm{z}_n ; \bm{\hat{r}}_{n\,\bigcdot}), \\
	E(\bm{D}) &= \text{Dir}(\bm{D}; \bar{d}). \phantom{\prod}
\end{align}
Note, the empirical prior is parameterized by: $\bm{\hat{r}}_{nk}$, $\bar{d}$, $\bm{\bar{W}}_k$, $\bar{v}_k$, $\bar{\beta}_k$ and $\bm{\bar{m}}_k$. Similarly, we want to compute posterior beliefs using all the available data $\bm{X}$. Once again, we approximate the true posterior using a structured mean-field approximation: 
\begin{align}
	P(\bm{Z}, \bm{\mu}, \bm{\Lambda}, \bm{D} |\bm{X}) &\approx Q(\bm{Z},\bm{D},\bm{\mu},\bm{\Lambda}) \delequal Q(\bm{Z}) Q(\bm{D}) \prod_{k=1}^K Q(\bm{\mu}_k | \bm{\Lambda}_k) Q(\bm{\Lambda}_k) \label{eq:var_distribution_def}
\end{align}
where to ensure conjugacy, the individual factors are defined as follows:
\begin{align}
	Q(\bm{\mu} | \bm{\Lambda}) &= \prod_{k=1}^K \mathcal{N}(\bm{\mu}_k; \bm{\hat{m}}_k, (\hat{\beta}_k\bm{\Lambda}_k)^{-1}), \\
	Q(\bm{\Lambda}) &= \prod_{k=1}^K \mathcal{W}(\bm{\Lambda}_k; \bm{\hat{W}}_k, \hat{v}_k), \\
	Q(\bm{Z}) &= \prod_{n=1}^N \text{Cat}(\bm{z}_n ; \bm{\hat{r}}_{n\,\bigcdot}), \\
	Q(\bm{D}) &= \text{Dir}(\bm{D}; \hat{d}). \phantom{\prod}
\end{align}
Note, the variational distribution is parameterized by: $\bm{\hat{r}}_{nk}$, $\hat{d}$, $\bm{\hat{W}}_k$, $\hat{v}_k$, $\hat{\beta}_k$ and $\bm{\hat{m}}_k$. Note, the posterior parameters can be identified by the hat over each symbol, while the empirical prior parameters have a bar on top of each symbol. Also, the responsibilities $\bm{\hat{r}}_{nk}$ are identical in the empirical prior and posterior distribution, because they cannot be used in the generative model. Instead, $\bar{d}$ will replace $d$, which will update the prior over $\bm{D}$ and indirectly the prior over states $P(\bm{Z}|\bm{D})$. Importantly, the inference process will rely on iteratively updating the parameters of the empirical prior and variational distribution until convergence of the variational free energy.

\subsubsection{Variational free energy} \label{sssec:vfe}

The goal of variational inference is to make the variational distribution as close as possible to the true posterior. Mathematically speaking, we want:
\begin{align}
	Q^*(\bm{Z},\bm{D},\bm{\mu},\bm{\Lambda}) \,\, &= \argmin_{Q(\bm{Z},\bm{D},\bm{\mu},\bm{\Lambda})} \kl{Q(\bm{Z},\bm{D},\bm{\mu},\bm{\Lambda})}{P(\bm{Z},\bm{D},\bm{\mu},\bm{\Lambda}|\bm{X})}.
\end{align}
However, we do not know the true posterior, therefore, Bayes theorem is used to re-arrange the above expression as follows:
\begin{align}
	Q^*(\bm{Z},\bm{D},\bm{\mu},\bm{\Lambda}) \,\, &= \argmin_{Q(\bm{Z},\bm{D},\bm{\mu},\bm{\Lambda})} \kl{Q(\bm{Z},\bm{D},\bm{\mu},\bm{\Lambda})}{P(\bm{X}, \bm{Z},\bm{D},\bm{\mu},\bm{\Lambda})} + \ln P(\bm{X}) \\
	&= \argmin_{Q(\bm{Z},\bm{D},\bm{\mu},\bm{\Lambda})} \underbrace{\kl{Q(\bm{Z},\bm{D},\bm{\mu},\bm{\Lambda})}{P(\bm{X}, \bm{Z},\bm{D},\bm{\mu},\bm{\Lambda})}}_{\text{Variational Free Energy}},
\end{align}
where the $\ln P(\bm{X})$ term can be dropped, since its is a constant w.r.t. $Q(\bm{Z},\bm{D},\bm{\mu},\bm{\Lambda})$. The variational free energy is defined as the KL-divergence between the variational distribution and the generative model, and can be re-arranged as follow using Equation \eqref{eq:gm_distribution_def} and \eqref{eq:var_distribution_def}:
\begin{align}
	\bm{F} &= \kl{Q(\bm{Z},\bm{D},\bm{\mu},\bm{\Lambda})}{P(\bm{X},\bm{Z},\bm{D},\bm{\mu},\bm{\Lambda})} \nonumber \\
	&= \mathbb{E}_{Q(\bm{D})}[\ln Q(\bm{D})] + \mathbb{E}_{Q(\bm{\mu}, \bm{\Lambda})}[\ln Q(\bm{\mu} | \bm{\Lambda})] + \mathbb{E}_{Q(\bm{\Lambda})}[\ln Q(\bm{\Lambda})] + \mathbb{E}_{Q(\bm{Z})}[\ln Q(\bm{Z})] \nonumber \\
	&- \mathbb{E}_{Q(\bm{D})}[\ln P(\bm{D})] - \mathbb{E}_{Q(\bm{\mu}, \bm{\Lambda})}[\ln P(\bm{\mu} | \bm{\Lambda})] - \mathbb{E}_{Q(\bm{\Lambda})}[\ln P(\bm{\Lambda})] - \mathbb{E}_{Q(\bm{Z},\bm{D})}[\ln P(\bm{Z}|\bm{D})] \nonumber \\
	&- \mathbb{E}_{Q(\bm{Z}, \bm{\mu}, \bm{\Lambda})}[\ln P(\bm{X} | \bm{Z}, \bm{\mu}, \bm{\Lambda})],
\end{align}
where the expectations are computed as described in Appendix B.

\subsubsection{Update equations} \label{sssec:update_equations}

\begin{figure}[H]
	\centering
	\subfigure[]{\includegraphics[width=0.40\textwidth]{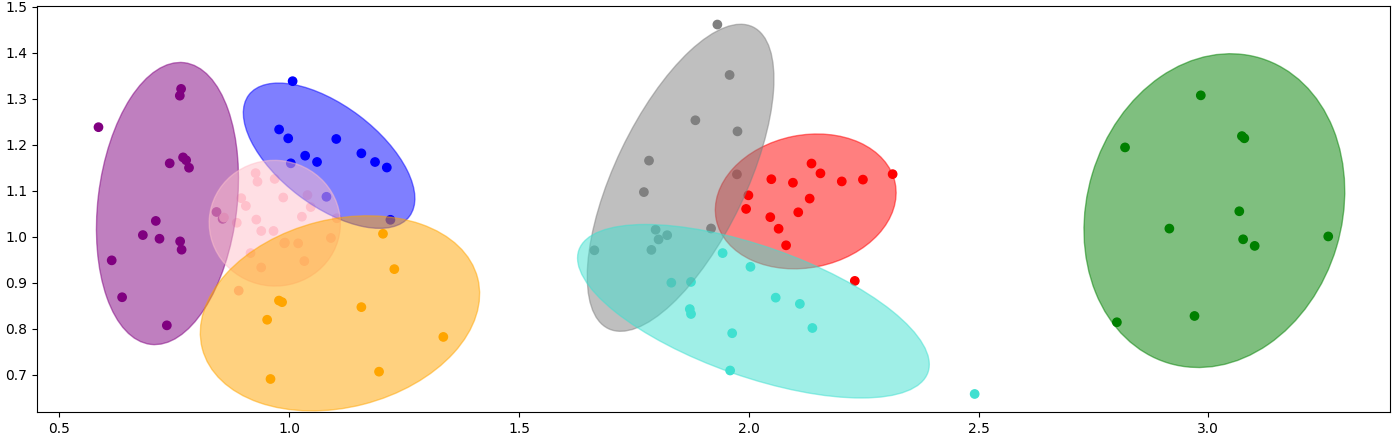}\label{fig:gm_step_0}} 
	\subfigure[]{\includegraphics[width=0.40\textwidth]{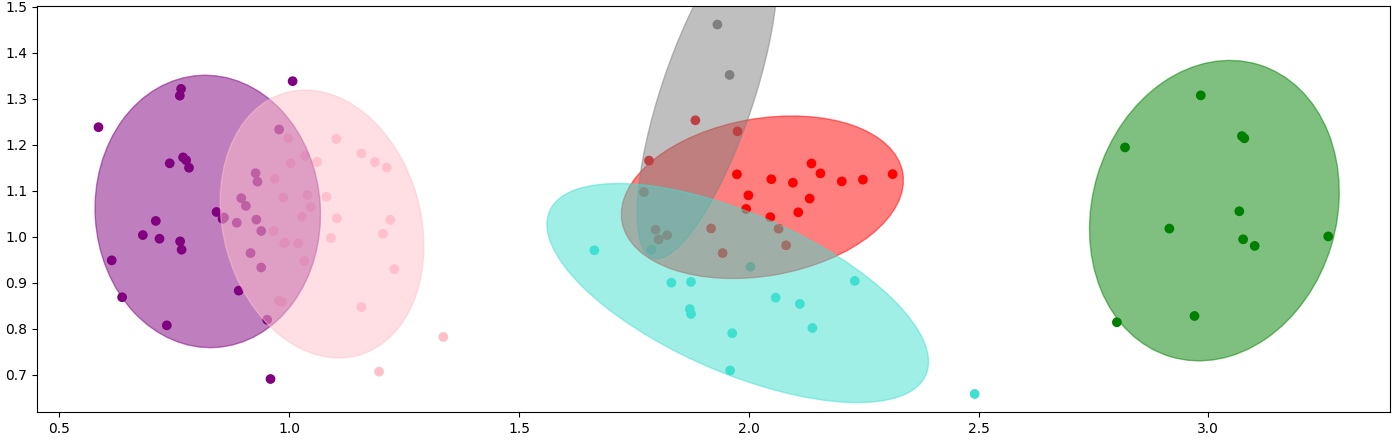}\label{fig:gm_step_5}}
	\subfigure[]{\includegraphics[width=0.40\textwidth]{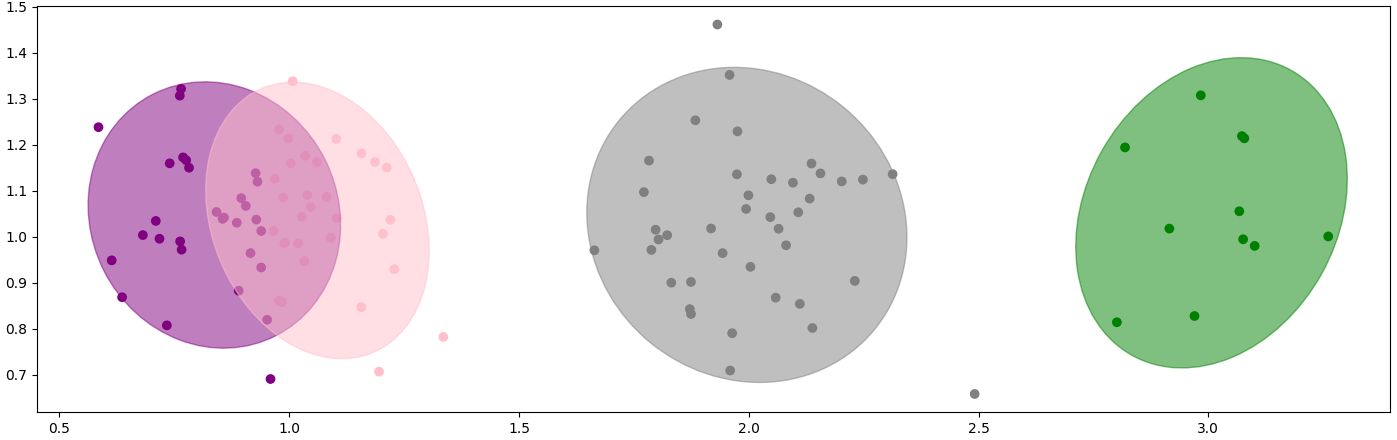}\label{fig:gm_step_7}}
	\subfigure[]{\includegraphics[width=0.40\textwidth]{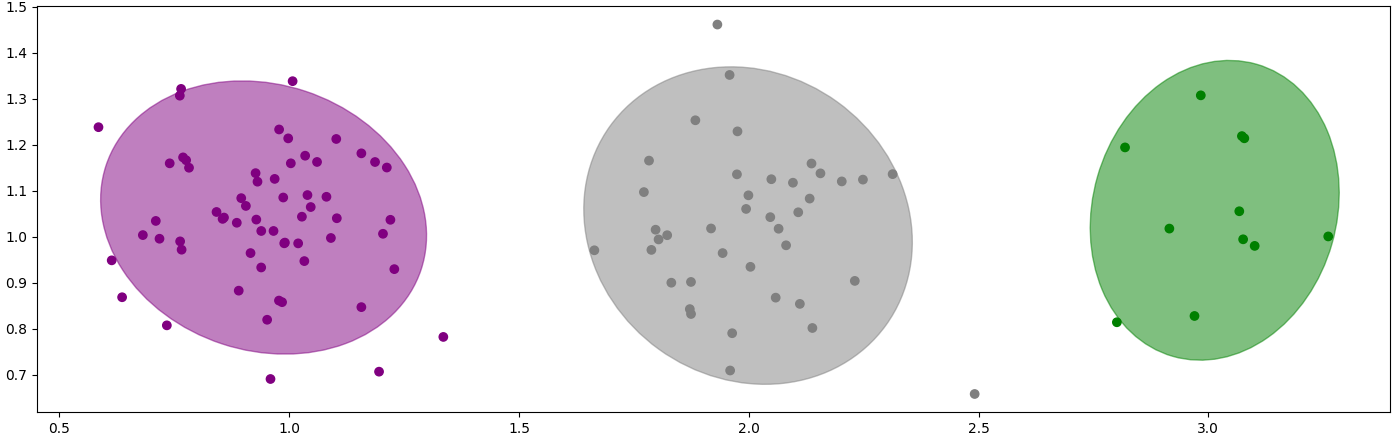}\label{fig:gm_step_10}} 		
	\vspace{-0.25cm}
	\caption{The variational Gaussian mixture: (a) before the optimization process, (b) after 5 steps of optimization, (c) after 7 steps of optimization, and (d) after 10 steps of optimization.
	}
	\label{fig:gm_steps}
\end{figure}

Let $\bm{Y} = \{\bm{Z},\bm{D},\bm{\mu},\bm{\Lambda}\}$ be the set of all latent variables, and $\bm{Y}_j$ be an arbitrary random variable. As explained by \citet{VMP_TUTO} and \citet{AI_VMP}, the variational free energy can be minimized by iteratively applying the following update equation for each latent variable $\bm{Y}_j$:
\begin{align}
	\ln Q^*(\bm{Y}_j) = \mathbb{E}_{i \neq j}[\ln P(\bm{X}, \bm{Y})] + c, \label{eq:VI_general_solution}
\end{align}
where $c$ is a constant, and the expectation is over all factors of $Q(\,\bigcdot\,)$ but $Q(Y_j)$. As shown in Appendix C, the update equations for the variational posterior of the VGM can be obtained from \eqref{eq:VI_general_solution}. Additionally, as explained in Appendix D, these equations can be re-arranged to provide update equations for both the empirical prior and variational distribution parameters. Interestingly, the final set of equations suggest that the empirical prior parameters should be updated as if we wanted to compute the posterior distribution given the data points to forget $\bm{X}^{'}$, and the prior parameters. Similarly, the variational posterior parameters need to be updated as if we wanted to compute the posterior distribution given the data points to keep $\bm{X}^{''}$, and the empirical prior parameters. More precisely, the empirical prior parameters $\bar{d}_k$, $\bar{v}_k$, and $\bar{\beta}_k$ should be updated as follows:
\begin{align}
	\bar{v}_k &= v_k + N_k^{'}, \\
	\bar{\beta}_k &= \beta_k + N_k^{'}, \\
	\bar{d}_{k} &= d_{k} + N_k^{'},
\end{align}
where $N_k^{'}$ is the number of data points to forget attributed to the $k$-th component:
\begin{align}
	N_k^{'} &= \sum_{n \in N^{'}} \bm{\hat{r}}_{nk}.
\end{align}
Similarly, the posterior parameters $\hat{d}_k$, $\hat{v}_k$, and $\hat{\beta}_k$ are obtained by simply adding $N_k^{''}$ to the associated empirical prior parameters $\bar{d}_k$, $\bar{v}_k$, and $\bar{\beta}_k$:
\begin{align}
	\hat{v}_k &= \bar{v}_k + N_k^{''}, \\
	\hat{\beta}_k &= \bar{\beta}_k + N_k^{''}, \\
	\hat{d}_{k} &= \bar{d}_{k} + N_k^{''},
\end{align}
where $N_k^{''}$ is the number of data points to keep attributed to the $k$-th component:
\begin{align}
	N_k^{''} &= \sum_{n \in N^{''}} \bm{\hat{r}}_{nk}.
\end{align}
Moreover, Appendix D derives the following update equations for the scale matrix $\bm{\bar{W}}_k$ and mean vector $\bm{\bar{m}}_k$ of the empirical prior:
\begin{align}
	\bm{\bar{W}}_k^{-1} &= \bm{W}_k^{-1} + N_k^{'} \bm{S}_k^{'} + \frac{\beta_kN_k^{'}}{\beta_k + N_k^{'}}(\bm{\bar{x}}_k^{'} - \bm{m}_k)(\bm{\bar{x}}_k^{'} - \bm{m}_k)^\top, \\
	\bm{\bar{m}}_k &= \frac{\beta_k\bm{m}_k + N_k^{'}\bm{\bar{x}}_k^{'}}{\beta_k + N_k^{'}},
\end{align}
where the empirical mean $\bm{\bar{x}}_k$ and covariance $\bm{S}_k$ of the data points in the $k$-th component are given by:
\begin{align}
	\bm{\bar{x}}_k^{'} &= \frac{1}{N_k^{'}}\sum_{n \in N^{'}} \bm{\hat{r}}_{nk}\bm{x}_n, \\
	\bm{S}_k^{'} &= \frac{1}{N_k^{'}}\sum_{n \in N^{'}} \bm{\hat{r}}_{nk}(\bm{x}_n - \bm{\bar{x}}_k{'})(\bm{x}_n - \bm{\bar{x}}_k{'})^\top.
\end{align}
Appendix D provides a derivation of the following update equations for the scale matrix $\bm{\hat{W}}_k$ and mean vector $\bm{\hat{m}}_k$ of the variational distribution:
\begin{align}
	\bm{\hat{W}}_k^{-1} &= \bm{\bar{W}}_k^{-1} + N_k^{''} \bm{S}_k^{''} + \frac{\bar{\beta}_k N_k^{''}}{\bar{\beta}_k + N_k^{''}}(\bm{\bar{x}}_k^{''} - \bm{\bar{m}}_k)(\bm{\bar{x}}_k^{''} - \bm{\bar{m}}_k)^\top, \\
	\bm{\hat{m}}_k &= \frac{\bar{\beta}_k\bm{\bar{m}}_k + N_k^{''}\bm{\bar{x}}_k^{''}}{\bar{\beta}_k + N_k^{''}},
\end{align}
where the empirical mean $\bm{\bar{x}}_k$ and covariance $\bm{S}_k$ of the data points in the $k$-th component are given by:
\begin{align}
	\bm{\bar{x}}_k^{''} &= \frac{1}{N_k^{''}}\sum_{n \in N^{''}} \bm{\hat{r}}_{nk}\bm{x}_n, \\
	\bm{S}_k^{''} &= \frac{1}{N_k^{''}}\sum_{n \in N^{''}} \bm{\hat{r}}_{nk}(\bm{x}_n - \bm{\bar{x}}_k^{''})(\bm{x}_n - \bm{\bar{x}}_k^{''})^\top.
\end{align}
Finally, the responsibilities $\bm{\hat{r}}_{nk}$ are updated as follows:
\begin{align}
	\bm{\hat{r}}_{nk} = \frac{\bm{\rho}_{nk}}{\sum_{k=1}^K \bm{\rho}_{nk}}, \label{eq:responsibilities}
\end{align}
where the logarithm of $\bm{\rho}_{nk}$ is given by:
\begin{align}
	\ln \bm{\rho}_{nk} &= \mathbb{E}_{Q(\bm{D})}[\ln \bm{D}_k] - \frac{K}{2} \ln 2 \pi + \frac{1}{2} \mathbb{E}_{Q(\bm{\Lambda}_k)}[\ln|\bm{\Lambda}_k|] - \frac{1}{2} \mathbb{E}_{Q(\bm{\mu}_k, \bm{\Lambda}_k)}[(\bm{x}_n - \bm{\mu}_k)^\top \bm{\Lambda}_k(\bm{x}_n - \bm{\mu}_k)],
\end{align}
and the expectations can be evaluated using \eqref{eq:wishart_ln_expectation}, \eqref{eq:wishart_gaussian_expectation_of_quadratic_form}, and \eqref{eq:dirichlet_ln_expectation}. Importantly, as shown in Figure \ref{fig:gm_steps}, iterating over the above update equations gives rise to a competition between the Gaussian components, which, in this example, ultimately leads to a Gaussian mixture with only three active components. Note that practically, the number of Gaussian components $K$ remains unchanged, but the responsibilities of the $k$-th component $\bm{\hat{r}}_{nk}$ can become zero for all $n$, i.e., the $k$-th component is not responsible for any data points and can be ignored. When this happens, the posterior parameters of this component revert to their prior counterparts.

\section{Transition Model} \label{sec:transition}

In this section, we explain how to learn the transition mapping between consecutive time steps using a categorical-Dirichlet model. When combined with the perception model from the last section, the resulting model is a temporal Gaussian mixture (TGM) that is able to learn the environment's structure. That is, a model which can learn the number of states and the transition probabilities between states given the action taken.

\subsection{Temporal Gaussian mixture} \label{ssec:tgm}

While the variational Gaussian mixture (from Section \ref{ssec:gm}) is able to learn a static model of data, it cannot model time series. A potential solution would be to duplicate the Gaussian mixture likelihood at each time step, and to define the transition probability between any two consecutive time steps, c.f., Figure \ref{fig:tgm_gm}. Unfortunately, as depicted in Figure \ref{fig:tgm}, adding such transition mapping interferes with the VGM's ability to learn one component for each cluster of data points. Put simply, we empirically found that learning the likelihood first, and using the posterior over latent variables to learn the transition mapping, works better than learning the likelihood and transition mappings jointly.

Therefore, we use a second generative model for the transition mapping. See Figure \ref{fig:aigm_gm} for an illustration of the model's structure. Note, the parameters of the variational distribution over $\bm{Z}_0$ and $\bm{Z}_1$ are computed by iterating over the update equations presented in Section \ref{sssec:update_equations}, and these parameters remain fixed during the inference process of the transition model.

More precisely, when the agent acts in the environment, it records the observations made $\bm{X}$, and the actions performed $\bm{A}$. Then, for each observation\footnote{Note, $\bm{x}_n$ is the $n$-th observation in the dataset, which corresponds to the observation received by the agent at time step $n$.} $\bm{x}_n \in \bm{X}$, the VGM is used to compute the optimal variational distribution over the corresponding latent variable $\bm{z}_n$. However, the transition model relies on two observed variables $\bm{X}_0$ and $\bm{X}_1$, i.e., it does not depend on $\bm{X}$. The trick is to consider any two consecutive observations $\bm{x}_n, \bm{x}_{n+1} \in \bm{X}$, as observations for $\bm{X}_0$ and $\bm{X}_1$, respectively. Consequently, the parameters of the variational distribution over $\bm{z}_n, \bm{z}_{n+1} \in \bm{Z}$ can be used to define the variational posterior over $\bm{Z}_0$ and $\bm{Z}_1$, respectively. Similarly, each time $\bm{x}_n$ and $\bm{x}_{n+1}$ are used as observations for $\bm{X}_0$ and $\bm{X}_1$, the action $\bm{a}_n$ becomes an observed value for $\bm{A}_0$.

Additionally, since $(\bm{x}_n, \bm{x}_{n+1}, \bm{a}_n, \bm{z}_n, \bm{z}_{n+1})$ are related to the variables $(\bm{X}_0, \bm{X}_1, \bm{A}_0, \bm{Z}_0, \bm{Z}_1)$, respectively, they will be denoted by $(\bm{x}_n^0, \bm{x}_n^1, \bm{a}_n^0, \bm{z}_n^0, \bm{z}_n^1)$. This simplifies the notation, for example, if we let $M$ be the number of data points available to train the transition model, then: $\bm{X}_0 = \{\bm{x}_1^0, ..., \bm{x}_{M}^0\}$, $\bm{X}_1 = \{\bm{x}_1^1, ..., \bm{x}_{M}^1\}$, $\bm{A}_0 = \{\bm{a}_1^0, ..., \bm{a}_{M}^0\}$, $\bm{Z}_0 = \{\bm{z}_1^0, ..., \bm{z}_{M}^0\}$, and $\bm{Z}_1 = \{\bm{z}_1^1, ..., \bm{z}_{M}^1\}$.

Finally, since we want to forget part of the data points as time passes, we assume that for all $\bm{Y}_t \in \{\bm{X}_0, \bm{X}_1, \bm{A}_0, \bm{Z}_0, \bm{Z}_1\}$, the $M$ data points have been split into a set of data points to forget $\bm{Y}^{'}_t = \{\bm{y}_i^t \,|\, i \in M^{'}\}$ and data points to keep $\bm{Y}^{''}_t = \{\bm{y}_j^t \,|\, j \in M^{''}\}$, where $M^{'}$ and $M^{''}$ are the indices of the data points to forget and keep, respectively. For example, if $\bm{Y}_t = \bm{X}_1$, then we have $\bm{X}^{'}_1 = \{\bm{x}_i^1 \,|\, i \in M^{'}\}$ and $\bm{X}^{''}_1 = \{\bm{x}_j^1 \,|\, j \in M^{''}\}$.

\subsubsection{Generative model}

\begin{figure}[H]
	\centering
	\subfigure[]{
	\begin{tikzpicture}[scale=1]
		\node[latent, yshift=0.5cm] (D) {$\bm{D}$};
		\node[obs, right=of D, xshift=-0.5cm] (A0) {$\bm{A}_0$};
		\node[latent, below=of D, yshift=0.5cm] (Z0) {$\bm{Z}_0$};
		\node[obs, below=of Z0] (X0) {$\bm{X}_0$};
		\node[latent, right=of Z0, xshift=-0.5cm, yshift=-0.5cm] (lambda) {$\bm{\Lambda}$};
		\node[latent, below=of lambda, yshift=0.5cm] (mu) {$\bm{\mu}$};
		\node[latent, right=of Z0, xshift=1cm] (Z1) {$\bm{Z}_1$};
		\node[latent, above=of Z1, yshift=-0.5cm] (B) {$\bm{B}$};
		\node[obs, below=of Z1] (X1) {$\bm{X}_1$};
		\edge {D} {Z0}
		\edge {B} {Z1}
		\edge {A0} {Z1}
		\edge {Z0} {Z1}
		\edge {Z0} {X0}
		\edge {Z1} {X1}
		\edge {lambda} {mu}
		\edge {lambda} {X0}
		\edge {mu} {X0}
		\edge {lambda} {X1}
		\edge {mu} {X1}
	\end{tikzpicture}
	\label{fig:tgm_gm}}
	\hspace{1cm}
	\subfigure[]{
	\begin{tikzpicture}[scale=1]
		\node[latent, yshift=0.5cm] (D) {$\bm{D}$};
		\node[latent, below=of D, yshift=0.5cm] (Z0) {$\bm{Z}_0$};
		\node[latent, right=of Z0, xshift=0.75cm] (Z1) {$\bm{Z}_1$};
		\node[latent, above=of Z1, yshift=-0.5cm] (B) {$\bm{B}$};
		\node[obs, left=of B, xshift=0.5cm] (A0) {$\bm{A}_0$};
		\edge {D} {Z0}
		\edge {B} {Z1}
		\edge {A0} {Z1}
		\edge {Z0} {Z1}
	\end{tikzpicture}
	\label{fig:aigm_gm}}
	\vspace{-0.25cm}
	\caption[Temporal model]{This figure illustrates the graphical model of (a) a partially observable Markov decision process with variational Gaussian mixture likelihood, and (b) a temporal model based on a categorical-Dirichlet model.}
	\label{fig:tm_gm}
\end{figure}
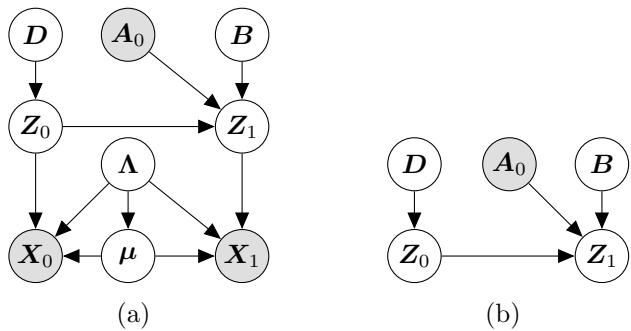

\begin{figure}[H]
	\vspace{-0.25cm}
	\centering
	\subfigure[]{
		{\includegraphics[width=0.7\textwidth]{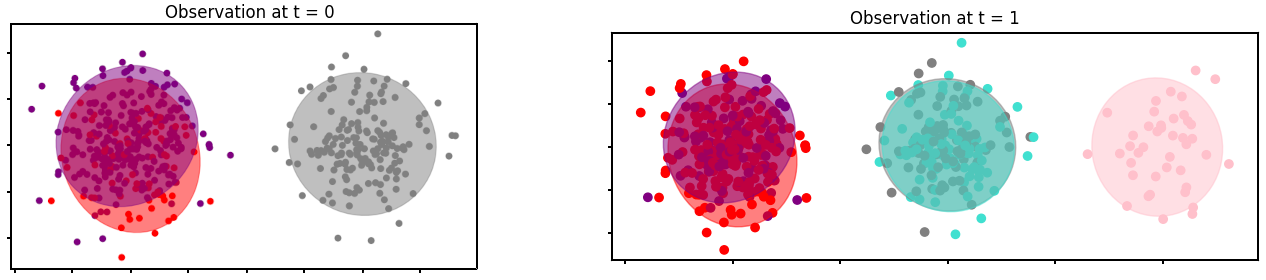}}
		\label{fig:tgm}
	}
	\subfigure[]{
		{\includegraphics[width=0.7\textwidth]{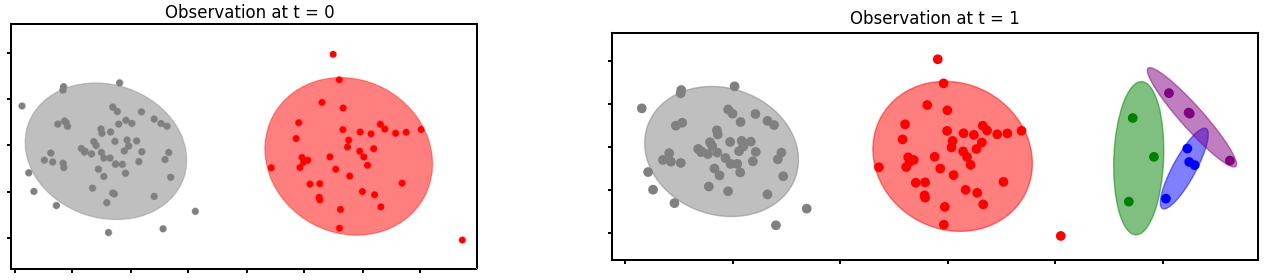}}
		\label{fig:aigm_100}
	}
	\subfigure[]{
		{\includegraphics[width=0.7\textwidth]{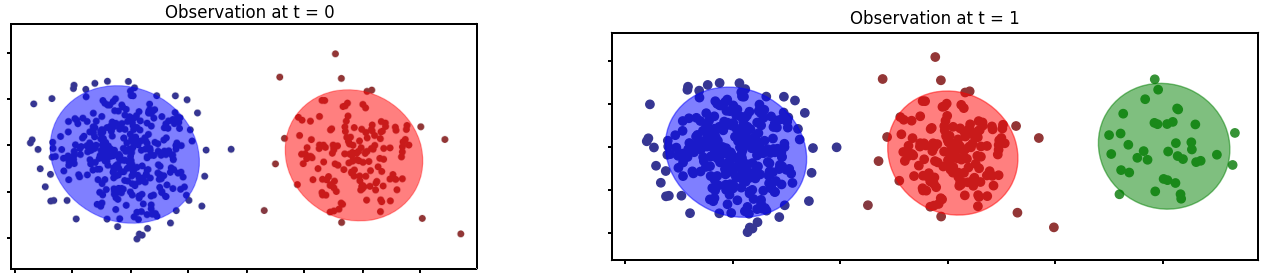}}
		\label{fig:aigm_500}
	}
	\vspace{-0.25cm}
	\caption{This figure illustrates: (a) the partially observable Markov decision process from Figure \ref{fig:tgm_gm}, (b) the categorical-Dirichlet model from Figure \ref{fig:aigm_gm} when provided with 100 data points, and (c) the categorical-Dirichlet model from Figure \ref{fig:aigm_gm} when provided with 500 data points.
	}
	\label{fig:tgm_vs_aigm}
\end{figure}

We now focus on the transition model definition. Like in Section \ref{sssec:generative_model}, the prior over $\bm{D}$ is a Dirichlet distribution, and the prior over $\bm{Z}_0$ is a Categorical distribution parameterized by $\bm{D}$:
\begin{align}
	P(\bm{D}) &= \text{Dir}(\bm{D}; d), \\
	P(\bm{Z}_0 | \bm{D}) &= \prod_{n=1}^{M} \text{Cat}(\bm{z}^0_n | \bm{D}),
\end{align}
where $M$ is the number of data points available to train the transition model. Perhaps unsurprisingly, the transition mapping is modelled as a Categorical distribution parameterized by a 3d-tensor $\bm{B}$, where the prior over $\bm{B}$ is a Dirichlet distribution:
\begin{align}
	P(\bm{B}) &= \text{Dir}(\bm{B}; b), \\
	P(\bm{Z}_1 | \bm{Z}_0, \bm{A}_0, \bm{B}) &= \prod_{n=1}^{M} \text{Cat}(\bm{z}^1_n | \bm{z}^0_n, \bm{a}^0_n, \bm{B}).
\end{align}
Finally, merging the above four equations, leads to the full generative model for the transition model:
\begin{align}
	P(\bm{Z}_{0:1}, \bm{B}, \bm{D} | \bm{A}_0) = P(\bm{Z}_0 | \bm{D}) P(\bm{D}) P(\bm{Z}_1 | \bm{Z}_0, \bm{A}_0, \bm{B}) P(\bm{B}),
\end{align}
which corresponds to the Bayesian network depicted in Figure \ref{fig:aigm_gm}.

\subsubsection{Empirical prior, variational distribution and update equations}

Since we want to forget the data points indexed by $M^{'}$ and keep the data points indexed by $M^{''}$, we introduced an empirical prior like in Section \ref{sss:ep_and_vd}. Once inference has been performed, the parameters of the empirical prior will be used as prior parameters, effectively integrating the information provided by the data points to forget into the prior distribution. After this replacement occurs, the data points indexed by $M^{'}$ can be safely discarded. More precisely, the empirical prior is defined as follows:
\begin{align}
	P(\bm{Z}_{0:1}^{'}, \bm{B}, \bm{D} | \bm{A}_0^{'}) &\approx E(\bm{Z}_{0:1}^{'}, \bm{B}, \bm{D}) \delequal E(\bm{B}) E(\bm{D}) E(\bm{Z}_0^{'}) E(\bm{Z}_1^{'}),
\end{align}
where the individual factors are defined as follows:
\begin{align}
E(\bm{Z}_\tau^{'}) &= \prod_{n \in M^{'}} \text{Cat}(\bm{z}^\tau_n; \bm{\hat{r}}_{n\,\bigcdot}^\tau) \\
E(\bm{B}) &= \text{Dir}(\bm{B}; \bar{b}), \\
E(\bm{D}) &= \text{Dir}(\bm{D}; \bar{d}).
\end{align}
Additionally, we approximate the full posterior by a variational distribution as follows:
\begin{align}
	P(\bm{Z}_{0:1}, \bm{B}, \bm{D} | \bm{A}_0) &\approx Q(\bm{Z}_{0:1}, \bm{B}, \bm{D}) \delequal Q(\bm{B}) Q(\bm{D}) Q(\bm{Z}_0) Q(\bm{Z}_1),
\end{align}
where the individual factors are defined as follows:
\begin{align}
	Q(\bm{Z}_\tau) &= \prod_{n=1}^{M} \text{Cat}(\bm{z}^\tau_n; \bm{\hat{r}}_{n\,\bigcdot}^\tau) \\
	Q(\bm{B}) &= \text{Dir}(\bm{B}; \hat{b}), \\
	Q(\bm{D}) &= \text{Dir}(\bm{D}; \hat{d}).
\end{align}
Normally, we would derive the update equations for $\bm{Z}_0$, $\bm{Z}_1$, $\bm{B}$, and $\bm{D}$. However, we are only interested in learning the transition mapping. Thus, only the update equation for $\bm{B}$ has been derived. Additionally, as discussed at the beginning of Section \ref{ssec:tgm}, the parameters of $Q(\bm{Z}_0)$ and $Q(\bm{Z}_1)$ are initialized using the responsibilities obtained by performing inference in the perception model. As shown in Appendix E, the parameters of $E(\bm{B})$ should be computed as follows:
\begin{align}
	\bar{b}[a]_{kj} = b[a]_{kj} + \sum_{n \in M^{'}} [a = a_n^0] \bm{\hat{r}}_{nk}^0 \bm{\hat{r}}_{nj}^1,
\end{align}
where the parameter $b$ was initialized as a 3d-tensor filled with ones. Similarly, the parameters of $Q(\bm{B})$ should be computed as follows:
\begin{align}
	\hat{b}[a]_{kj} = \bar{b}[a]_{kj} + \sum_{n \in M^{''}} [a = a_n^0] \bm{\hat{r}}_{nk}^0 \bm{\hat{r}}_{nj}^1.
\end{align}

\begin{figure}
	\vspace{-0.15cm}
	\centering
	\includegraphics[width=0.9\textwidth]{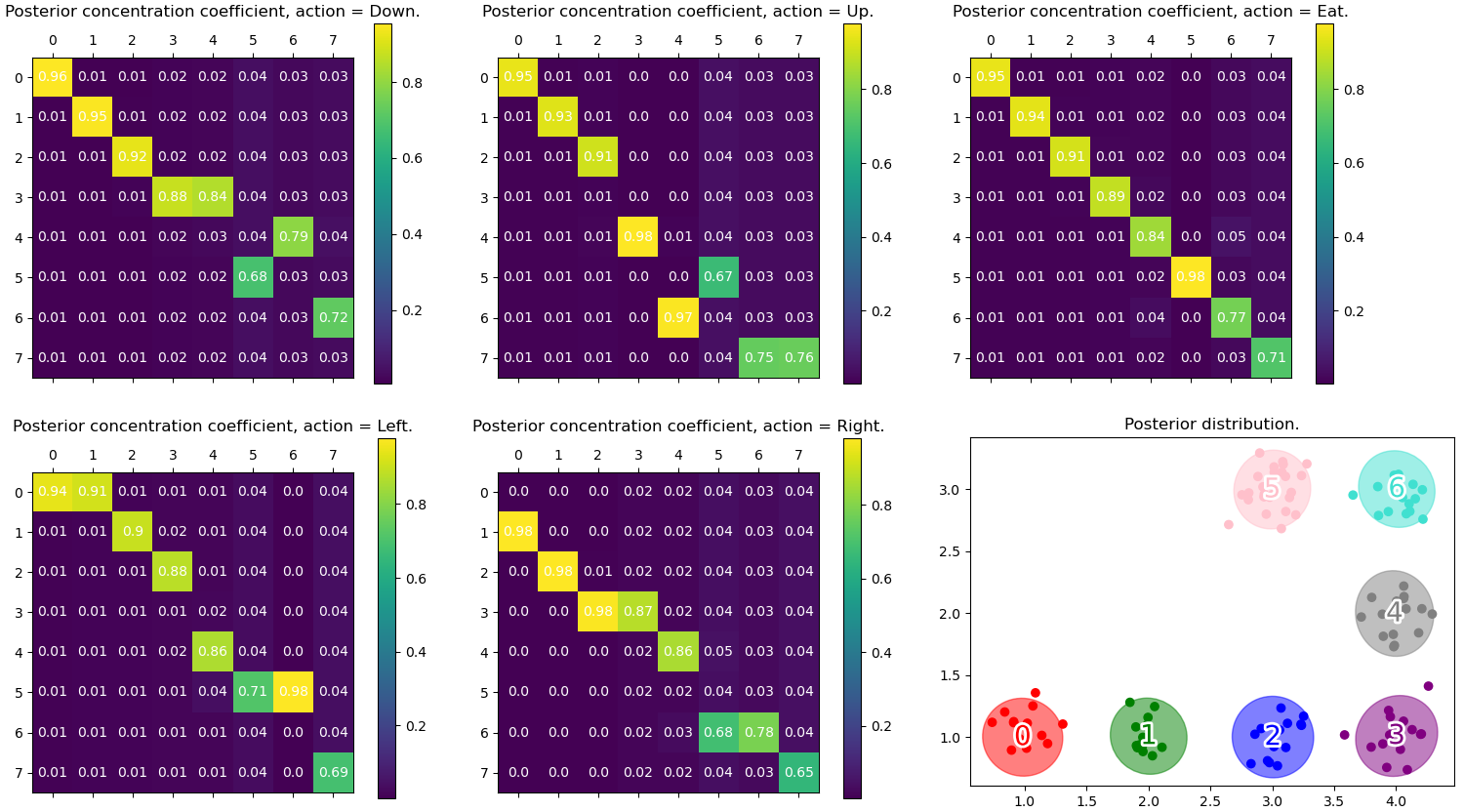}
	\vspace{-0.25cm}
	\caption{This figure depicts the transition matrices learnt from 5.000 data points, i.e., each matrix corresponds to a different action. Note, the ``eat" action allows the mouse to eat the cheese; if the mouse is not on the maze's terminal cell, this action is equivalent to idle. The bottom-right figure illustrates the Gaussian components learned by the model and the state index associated to each component. Note, state seven is above state six, but it is not displayed because the agent did not visit this state in the last 100 steps.}
	\label{fig:transition_matrices}
\end{figure}

Note, as the perception model learns the number of hidden states, the size of $\bar{b}$ and $\hat{b}$ change. Additionally, the probability of transitioning from the $k$-th component, to the $j$-th component, under action $a$, is proportional to the (probabilistic) number of times such transition a was observed in the data. Indeed, since latent variables are not directly observable, $\bm{\hat{r}}_{nk}^t$ does not have to be zero or one, instead, if $\bm{\hat{r}}_{nk}^0 = 0.5$ and $\bm{\hat{r}}_{nj}^1 = 0.5$, then the corresponding element in the sum will add $0.25$ to the (probabilistic) number of times such a transition was observed. Figure \ref{fig:transition_matrices} illustrates the learned $\bm{B}$ matrices. Note, the environment was a maze composed of eight states, seven of which are illustrated in the bottom-right image. The starting position was state zero, and the exit cell was state 5. Finally, in this environment, performing an action leading the agent to bump into a wall (e.g., go up in state two) results in the agent not moving.

\section{To forget or not to forget} \label{sec:forgetting}

In the previous sections, we described the perception and transition models. Importantly, we explained that each model is provided with two sets of data points, i.e., the data points that can be forgotten and the data points that need to be kept in memory. The forgettable data points are used to compute empirical priors, i.e., posterior distributions taking into account only the data points to forget. Importantly, empirical priors have the same functional form as the prior distributions and can therefore be used as prior beliefs when the inference process is over, effectively integrating the information provided by the forgettable data points into the prior. We now focus on how to decide which data points to forget and keep.

\subsection{Plasticity vs stability dilemma}

When learning, the brain needs to be plastic by allowing changes in synaptic connectivity, as well as in the strength of existing synaptic connections. However, these changes should ideally preserve the previously learned concepts to ensure some form of stability over time, and avoid forgetting useful information. This is the plasticity versus stability dilemma \citep{10.3389/fpsyg.2013.00504}. 

Our model is faced with a similar challenge, as in general, new components are progressively discovered as the agent explores the environment. These new components need to be learned without forgetting older components. For example, imagine a mouse moving through a maze and observing a noisy estimate of its position. As new maze cells are explored, new positions are sampled and new components will become visible. However, if the number of samples for a new component is too small, this component may be confused with nearby cells, until enough data points are available to differentiate between the cells.

\subsection{Flexible and fixed components}

\begin{figure}[H]
	\vspace{-0.15cm}
	\centering
	\includegraphics[width=0.9\textwidth]{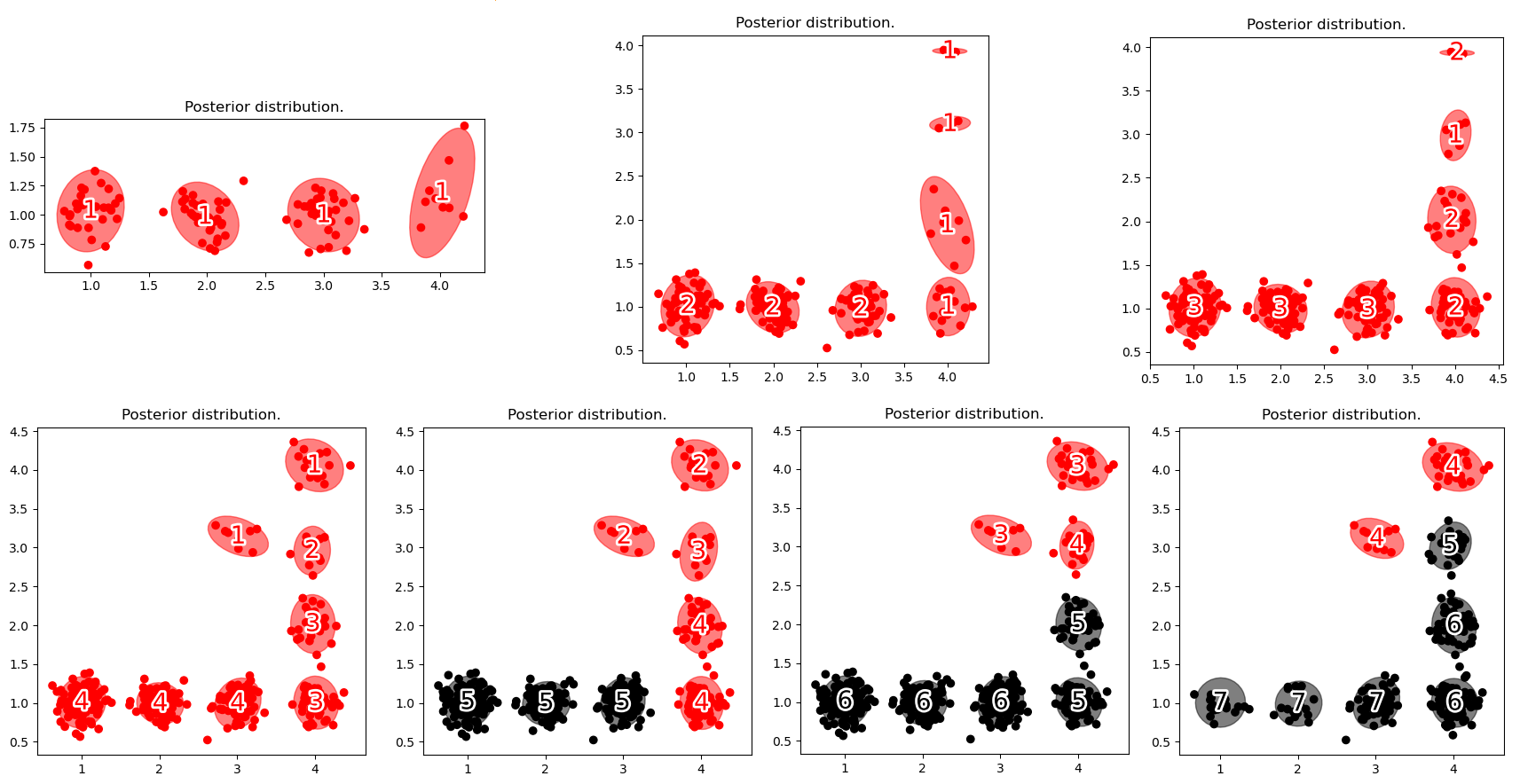}
	\vspace{-0.25cm}
	\caption{This figure depicts the Gaussian components discovered every 100 training steps. The number associated to each component counts how many times a component has been observed. When a component has been observed five times in a row or more, this component becomes fixed and is therefore depicted in black. In contrast, red components are flexible components.}
	\label{fig:flexible_and_fixed_components}
\end{figure}

In this section, we discuss how our model addresses the plasticity vs stability dilemma. More precisely, we assume that the Gaussian components can be either flexible (plastic) or fixed (stable). All components are initialized flexibly, and they become fixed if they persist over long periods of time. Figure \ref{fig:flexible_and_fixed_components} illustrates how components transform into fixed components.

Mathematically, every 100 training iterations, the variational Gaussian mixture identifies a set of Gaussian components $\mathcal{G}_t$, where $t \in \{100, 200, ...\}$. Given two Gaussian components with distributions $\mathcal{N}(X; \bm{\mu}_1, \bm{\Lambda}_1) \in \mathcal{G}_t$ and $\mathcal{N}(X; \bm{\mu}_2, \bm{\Lambda}_2) \in \mathcal{G}_{t+100}$, we say that $\mathcal{N}(X; \bm{\mu}_1, \bm{\Lambda}_1)$ at time $t$ becomes $\mathcal{N}(X; \bm{\mu}_2, \bm{\Lambda}_2)$ at time $t+100$, if:
\begin{align}
	\kl{\mathcal{N}(X; \bm{\mu}_1, \bm{\Lambda}_1)}{\mathcal{N}(X; \bm{\mu}_2, \bm{\Lambda}_2)} < \theta_{\text{kl}},
\end{align}
where $\theta_{\text{kl}}$ is the threshold under which two Gaussian components are considered identical, and in our simulation $\theta_{\text{kl}}$ was set to 0.5. In other words, a Gaussian component $\mathcal{N}(X; \bm{\mu}_1, \bm{\Lambda}_1) \in \mathcal{G}_t$ persists over time if there exists another component $\mathcal{N}(X; \bm{\mu}_2, \bm{\Lambda}_2) \in \mathcal{G}_{t+100}$ for which the KL-divergence is smaller than a predetermined threshold. If a Gaussian component persists more than $\theta_{\text{counts}}$ times in a row, the component becomes fixed. In our simulation, we chose $\theta_{\text{counts}} = 4$.

\vspace{-1cm}
\subsection{Which data points should be forgotten?}

\begin{figure}[H]
	\begin{center}
		\begin{tikzpicture}
			
			\node[latent, fill=blue!50!white] (x0) {$x_0$};
			\node[latent, right=of x0] (x1) {$x_1$};
			\node[latent, right=of x1] (x2) {$x_2$};
			\node[latent, right=of x2] (x3) {$x_3$};
			\node[latent, fill=blue!50!white, right=of x3] (x4) {$x_4$};
			\node[latent, right=of x4] (x5) {$x_5$};
			\node[latent, right=of x5] (x6) {$x_6$};
			\node[latent, right=of x6] (x7) {$x_7$};
			\node[latent, fill=blue!50!white, right=of x7] (x8) {$x_8$};
			
			\node[latent, fill=black!50!white, above=of x0] (r0) {$r_0$};
			\node[latent, fill=black!50!white, above=of x1] (r1) {$r_1$};
			\node[latent, fill=red!50!white, above=of x2] (r2) {$r_2$};
			\node[latent, fill=black!50!white, above=of x3] (r3) {$r_3$};
			\node[latent, fill=black!50!white, above=of x4] (r4) {$r_4$};
			\node[latent, fill=black!50!white, above=of x5] (r5) {$r_5$};
			\node[latent, fill=red!50!white, above=of x6] (r6) {$r_6$};
			\node[latent, fill=black!50!white, above=of x7] (r7) {$r_7$};
			\node[latent, fill=black!50!white, above=of x8] (r8) {$r_8$};
			
			\node[latent, above=of r0] (a0) {$a_0$};
			\node[latent, above=of r1] (a1) {$a_1$};
			\node[latent, above=of r2] (a2) {$a_2$};
			\node[latent, above=of r3] (a3) {$a_3$};
			\node[latent, above=of r4] (a4) {$a_4$};
			\node[latent, above=of r5] (a5) {$a_5$};
			\node[latent, above=of r6] (a6) {$a_6$};
			\node[latent, above=of r7] (a7) {$a_7$};
			
			\draw [decoration={brace, amplitude=6pt, raise=15pt, aspect=0.5, mirror}, decorate] (x0.west) -- (x0.east) node [below=18pt, pos=0.5] {Start of trial};
			
			\draw [decoration={brace, amplitude=6pt, raise=15pt, aspect=0.5, mirror}, decorate] (x4.west) -- (x4.east) node [below=18pt, pos=0.5] {Middle of trial};
			
			\draw [decoration={brace, amplitude=6pt, raise=15pt, aspect=0.5, mirror}, decorate] (x8.west) -- (x8.east) node [below=18pt, pos=0.5] {End of trial};
			
			\node[latent, scale=0.3, fill=blue!50!white, above=of r0, xshift=2.85cm, yshift=-2.05cm] (forget0) {};
			\draw (r0) -- (forget0);
			\draw (r1) -- (forget0);
			\draw (a0) -- (forget0);
			
			\node[latent, scale=0.3, above=of r1, xshift=2.85cm, yshift=-2.05cm] (keep0) {};
			\draw (r1) -- (keep0);
			\draw (r2) -- (keep0);
			\draw (a1) -- (keep0);
			
			\node[latent, scale=0.3, above=of r2, xshift=2.85cm, yshift=-2.05cm] (keep1) {};
			\draw (r2) -- (keep1);
			\draw (r3) -- (keep1);
			\draw (a2) -- (keep1);
			
			\node[latent, scale=0.3, fill=blue!50!white, above=of r3, xshift=2.85cm, yshift=-2.05cm] (forget1) {};
			\draw (r3) -- (forget1);
			\draw (r4) -- (forget1);
			\draw (a3) -- (forget1);
			
			\node[latent, scale=0.3, fill=blue!50!white, above=of r4, xshift=2.85cm, yshift=-2.05cm] (forget2) {};
			\draw (r4) -- (forget2);
			\draw (r5) -- (forget2);
			\draw (a4) -- (forget2);
			
			\node[latent, scale=0.3, above=of r5, xshift=2.85cm, yshift=-2.05cm] (keep2) {};
			\draw (r5) -- (keep2);
			\draw (r6) -- (keep2);
			\draw (a5) -- (keep2);
			
			\node[latent, scale=0.3, above=of r6, xshift=2.85cm, yshift=-2.05cm] (keep3) {};
			\draw (r6) -- (keep3);
			\draw (r7) -- (keep3);
			\draw (a6) -- (keep3);
			
			\node[latent, scale=0.3, fill=blue!50!white, above=of r7, xshift=2.85cm, yshift=-2.05cm] (forget3) {};
			\draw (r7) -- (forget3);
			\draw (r8) -- (forget3);
			\draw (a7) -- (forget3);
			
		\end{tikzpicture}
	\end{center}
	\vspace{-0.75cm}
	\caption{This figure illustrates the three kind of data points that can be forgotten. Note, $x_t$ corresponds to the observation at time $t$, $r_t$ corresponds to the responsibilities at time $t$, and $a_t$ corresponds to the action at time $t$. If $r_t$ has a black background it means that the corresponding observation $x_t$ is likely to come from a fixed Gaussian component. In contrast, if $r_t$ has a red background, then it likely comes from a flexible components. Observations that can be forgotten are depicted using a blue background, while observation to keep have a white background. Finally, small blue nodes connecting a triplet ($a_t$, $r_t$, $r_{t+1}$) correspond to transitions to forget, while small white nodes correspond to transition to keep.}
	\label{fig:forgettable_data_points}
\end{figure}
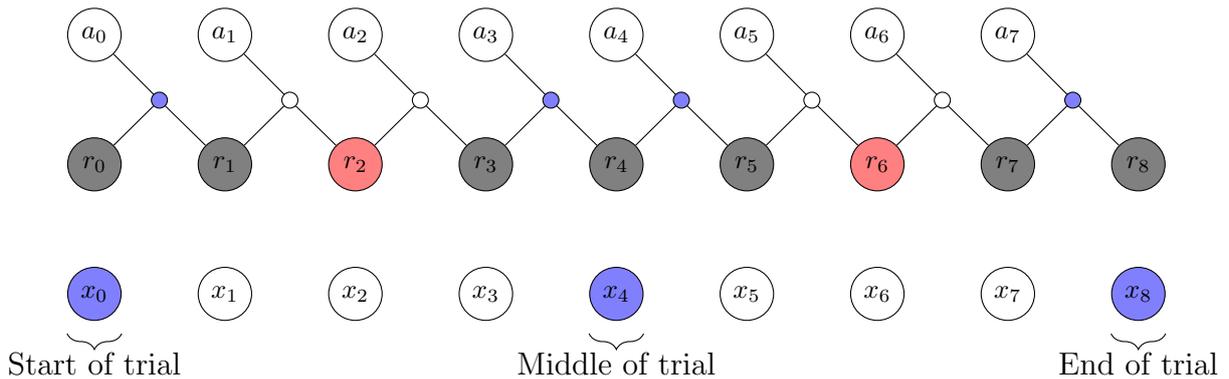

Figure \ref{fig:forgettable_data_points} illustrates the data available for a trial composed of seven actions and eight observations. For each observation, the variational Gaussian mixture can be used to compute the associated responsibilities. An observation $x_t$ can be forgotten if the previous $x_{t-1}$, current $x_t$, and next $x_{t+1}$ observations are all associated to a fixed component. Of course, if the observation is at the beginning of a trial, then the previous observation can be ignored. Similarly, if the observation is at the end of a trial, then the next observation can be ignored.

\begin{figure}[H]
	\begin{center}
		\begin{tikzpicture}
				
			\node[] (x0) {};
			\node[latent, right=of x0] (x1) {$x_1$};
			\node[latent, right=of x1] (x2) {$x_2$};
			\node[latent, right=of x2] (x3) {$x_3$};
			\node[latent, right=of x4] (x5) {$x_5$};
			\node[latent, right=of x5] (x6) {$x_6$};
			\node[latent, right=of x6] (x7) {$x_7$};
				
			\node[latent, fill=black!50!white, above=of x1] (r1) {$r_1$};
			\node[latent, fill=red!50!white, above=of x2] (r2) {$r_2$};
			\node[latent, fill=black!50!white, above=of x3] (r3) {$r_3$};
			\node[latent, fill=black!50!white, above=of x5] (r5) {$r_5$};
			\node[latent, fill=red!50!white, above=of x6] (r6) {$r_6$};
			\node[latent, fill=black!50!white, above=of x7] (r7) {$r_7$};
				
			\node[latent, above=of r1] (a1) {$a_1$};
			\node[latent, above=of r2] (a2) {$a_2$};
			\node[latent, above=of r5] (a5) {$a_5$};
			\node[latent, above=of r6] (a6) {$a_6$};
				
			\node[latent, scale=0.3, above=of r1, xshift=2.85cm, yshift=-2.05cm] (keep0) {};
			\draw (r1) -- (keep0);
			\draw (r2) -- (keep0);
			\draw (a1) -- (keep0);
				
			\node[latent, scale=0.3, above=of r2, xshift=2.85cm, yshift=-2.05cm] (keep1) {};
			\draw (r2) -- (keep1);
			\draw (r3) -- (keep1);
			\draw (a2) -- (keep1);
				
			\node[latent, scale=0.3, above=of r5, xshift=2.85cm, yshift=-2.05cm] (keep2) {};
			\draw (r5) -- (keep2);
			\draw (r6) -- (keep2);
			\draw (a5) -- (keep2);
				
			\node[latent, scale=0.3, above=of r6, xshift=2.85cm, yshift=-2.05cm] (keep3) {};
			\draw (r6) -- (keep3);
			\draw (r7) -- (keep3);
			\draw (a6) -- (keep3);
				
		\end{tikzpicture}
	\end{center}
	\vspace{-0.75cm}
	\caption{This figure shows the dataset after removing the forgettable data points in Figure \ref{fig:forgettable_data_points}.}
	\label{fig:after_forgeting_data_points}
\end{figure}
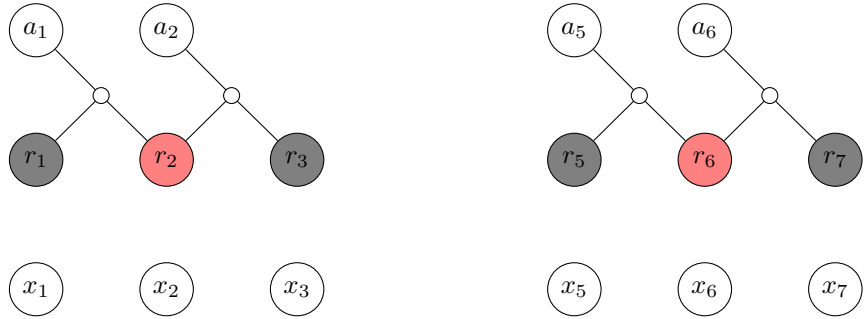

Following these rules in Figure \ref{fig:forgettable_data_points}, the observations $x_0$, $x_4$ and $x_8$ can be forgotten. Therefore, in this case $N^{'} = \{0, 4, 8\}$ and $N^{''} = \{1, 2, 3, 5, 6, 7\}$. If we forget these data points, we obtain the dataset illustrated in Figure \ref{fig:after_forgeting_data_points}. This sheds some light on the triplets ($a_t$, $r_t$, $r_{t+1}$) that will become unavailable, and that must therefore be added to the forgetting set of the transition model. Specifically, in this case $M^{'} = \{0, 3, 4, 7\}$ and $M^{''} = \{1, 2, 5, 6\}$. Thus, for example, the triplet ($a_7$, $r_7$, $r_8$) can be forgotten, which reflects the fact that $7 \in M^{'}$.

\section{Planning and decision making} \label{sec:planning}

In this section, we explain how Q-learning \citep{sutton2018reinforcement} can be adapted to work with a problem where there is uncertainty about the current states. First, we introduce the standard Q-learning algorithm, then we adapt this algorithm to work with beliefs over states.

\subsection{Q-learning}

The Q-learning algorithm aims to solve the reinforcement learning problem, where at each time step, an agent takes an action $a_t$ based on the current state $s_t$. Then, the environment produces a next state $s_{t+1}$ and the reward $r_{t+1}$ obtained by taking action $a_t$ in state $s_t$. The agent's goal is to maximize the discounted sum of future rewards:
\begin{align}
	G_t = \sum_{\tau = t + 1}^{T} \gamma^{\tau - t - 1} r_{\tau}, \label{eq:return_definit}
\end{align}
where $\gamma \in [0, 1]$ is the discount factor, and the time horizon $T$ must be larger than $t + 1$. In reinforcement learning, a policy $\pi$ defines the probability of taking each action in each state, i.e., $\pi(a|s)$. The expected value of a state $s$ under a policy $\pi$ is denoted $v_{\pi}(s)$, i.e.,
\begin{align}
	v_{\pi}(s) = \mathbb{E}_\pi [G_t \, | \, s_t = s] = \mathbb{E}_\pi \Bigg[\sum_{\tau = t + 1}^{T} \gamma^{\tau - t - 1} r_{\tau} \, \Bigg| \, s_t = s\Bigg],
\end{align}
where $v_{\pi}(s)$ are sometimes called the state values. Importantly, different policies $\pi$ lead to different state values, and there exists at least one policy $\pi^{*}$ with the highest state values:
\begin{align}
	v_{*}(s) = \max_{\pi} v_{\pi}(s),
\end{align}
where $v_{*}(s)$ are the optimal state values obtained by following one of the optimal policies $\pi^{*}$. Similarly, the expected value of taking action $a$ in state $s$, and then follow policy $\pi$ is denoted $q_{\pi}(a, s)$, i.e., 
\begin{align}
	q_{\pi}(a, s) = \mathbb{E}_\pi [G_t \, | \, a_t = a, s_t = s] = \mathbb{E}_\pi \Bigg[\sum_{\tau = t + 1}^{T} \gamma^{\tau - t - 1} r_{\tau} \, \Bigg| \, a_t = a, s_t = s\Bigg],
\end{align}
where $q_{\pi}(a, s)$ are sometime called the Q-values. Importantly, different policies $\pi$ lead to different Q-values, and there exists at least one policy $\pi^{*}$ with the highest Q-values:
\begin{align}
	q_{*}(a, s) = \max_{\pi} q_{\pi}(a, s),
\end{align}
where $q_{*}(a, s)$ are the optimal Q-values obtained by following one of the optimal policies $\pi^{*}$. As shown in Appendix F, the following relationship holds:
\begin{align}
	q_{*}(a_t, s_t) &= \sum_{s_{t+1}, r_{t+1}} P(s_{t+1}, r_{t+1} | s_t, a_t) \big[ r_{t + 1} + \gamma v_{*}(s_{t+1}) \big] \label{eq:optimal-q-values-and-state-values}\\
	&= \sum_{s_{t+1}, r_{t+1}} P(s_{t+1}, r_{t+1} | s_t, a_t) \big[ r_{t + 1} + \gamma \max_{a_{t+1}} q_{*}(a_{t+1}, s_{t+1}) \big], \label{eq:optimal-q-values-and-q-values}
\end{align}
where intuitively Equation \eqref{eq:optimal-q-values-and-state-values} says that the optimal Q-values is equal to the expected immediate reward, plus the discounted value of the future state. Additionally, by comparing Equations \eqref{eq:optimal-q-values-and-state-values} and \eqref{eq:optimal-q-values-and-q-values}, we see that the optimal value of a state is only as good as the value of the best action in that state, i.e.,
\begin{align}
	v_{*}(s_{t+1}) = \max_{a_{t+1}} q_{*}(a_{t+1}, s_{t+1}).
\end{align}
The goal of the Q-learning algorithm is to estimate the Q-values as given in Equation \eqref{eq:optimal-q-values-and-q-values}. The expectation over $P(r_{t+1} | s_{t+1}, s_t, a_t)$ is approximated using Monte-Carlo sampling:
\begin{align}
	q_{*}(a_t, s_t) &= \sum_{s_{t+1}, r_{t+1}} P(s_{t+1}, r_{t+1} | s_t, a_t) \big[ r_{t + 1} + \gamma \max_{a_{t+1}} q_{*}(a_{t+1}, s_{t+1}) \big] \\
	&\approx r_{t + 1} + \gamma \sum_{s_{t+1}} P(s_{t+1} | s_t, a_t) \big[ \max_{a_{t+1}} q_{*}(a_{t+1}, s_{t+1}) \big].
\end{align}
The Q-learning works by storing the Q-values in a table denoted $q(a_t, s_t)$, then ideally, the following update equation would be iterated:
\begin{align}
	q(a_t, s_t) &\leftarrow q(a_t, s_t) + \alpha \Bigg[\underbrace{r_{t + 1} + \gamma \sum_{s_{t+1}} P(s_{t+1} | s_t, a_t) \big[ \max_{a_{t+1}} q_{*}(a_{t+1}, s_{t+1}) \big]}_{\text{Q-values target}} - \,\, q(a_t, s_t)\Bigg],
\end{align}
where $\alpha$ is the learning rate. However, the $q_{*}(a_{t+1}, s_{t+1})$ is the quantity that Q-learning aims to learn and is therefore unknown. Thus, we approximate $q_{*}(a_{t+1}, s_{t+1})$ by our current best guess $q(a_{t+1}, s_{t+1})$, leading to the following update equation:
\begin{align}
	q(a_t, s_t) &\leftarrow q(a_t, s_t) + \alpha \Bigg[r_{t + 1} + \gamma \sum_{s_{t+1}} P(s_{t+1} | s_t, a_t) \big[ \max_{a_{t+1}} q(a_{t+1}, s_{t+1}) \big] - q(a_t, s_t)\Bigg]. \label{eq:temproal_diff_in_Q_learning}
\end{align}
Importantly, when no model of the environment is available, the expectation w.r.t. $P(s_{t+1} | s_t, a_t)$ can be approximated using a Monte-Carlo estimate.

\subsection{Beliefs based Q-learning}

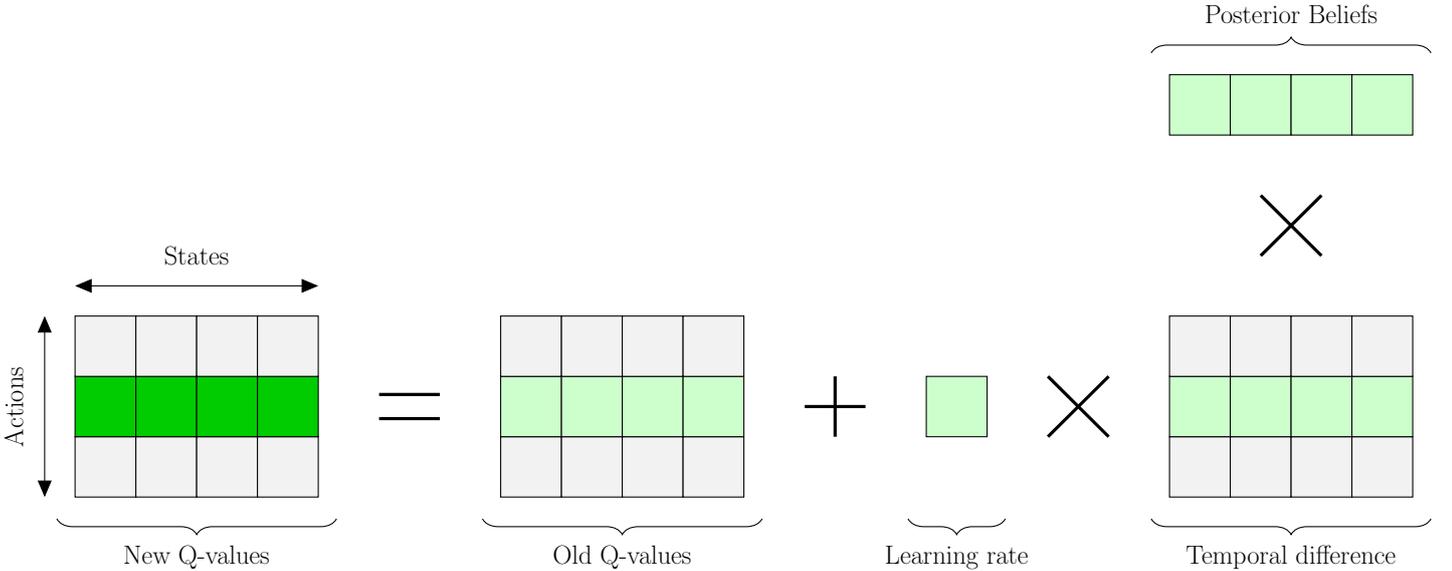
\begin{figure}[H]
	\centering
	\begin{tikzpicture}[scale=0.08, every node/.style={scale=0.08}]

		% New Q-values
		\draw[fill=gray!10!white] (0,0) rectangle (10, 10);
		\draw[fill=gray!10!white] (10,0) rectangle (20, 10);
		\draw[fill=gray!10!white] (20,0) rectangle (30, 10);
		\draw[fill=gray!10!white] (30,0) rectangle (40, 10);

		\draw[fill=green!80!black] (0,10) rectangle (10, 20);
		\draw[fill=green!80!black] (10,10) rectangle (20, 20);
		\draw[fill=green!80!black] (20,10) rectangle (30, 20);
		\draw[fill=green!80!black] (30,10) rectangle (40, 20);

		\draw[fill=gray!10!white] (0,20) rectangle (10, 30);
		\draw[fill=gray!10!white] (10,20) rectangle (20, 30);
		\draw[fill=gray!10!white] (20,20) rectangle (30, 30);
		\draw[fill=gray!10!white] (30,20) rectangle (40, 30);

		% Equal
		\draw[black, very thick] (50,13) rectangle (60, 13);
		\draw[black, very thick] (50,17) rectangle (60, 17);

		% Old Q-values
		\draw[fill=gray!10!white] (70,0) rectangle (80, 10);
		\draw[fill=gray!10!white] (80,0) rectangle (90, 10);
		\draw[fill=gray!10!white] (90,0) rectangle (100, 10);
		\draw[fill=gray!10!white] (100,0) rectangle (110, 10);

		\draw[fill=green!20!white] (70,10) rectangle (80, 20);
		\draw[fill=green!20!white] (80,10) rectangle (90, 20);
		\draw[fill=green!20!white] (90,10) rectangle (100, 20);
		\draw[fill=green!20!white] (100,10) rectangle (110, 20);

		\draw[fill=gray!10!white] (70,20) rectangle (80, 30);
		\draw[fill=gray!10!white] (80,20) rectangle (90, 30);
		\draw[fill=gray!10!white] (90,20) rectangle (100, 30);
		\draw[fill=gray!10!white] (100,20) rectangle (110, 30);
		
		% Plus
		\draw[black, very thick] (120,15) rectangle (130, 15);
		\draw[black, very thick] (125,10) rectangle (125, 20);

		% Learning rate
		\draw[fill=green!20!white] (140,10) rectangle (150, 20);

		% Multiplied by
		\draw[black, very thick] (160,20) -- (170, 10);
		\draw[black, very thick] (160,10) -- (170, 20);

		% Temporal difference Q-values
		\draw[fill=gray!10!white] (180,0) rectangle (190, 10);
		\draw[fill=gray!10!white] (190,0) rectangle (200, 10);
		\draw[fill=gray!10!white] (200,0) rectangle (210, 10);
		\draw[fill=gray!10!white] (210,0) rectangle (220, 10);

		\draw[fill=green!20!white] (180,10) rectangle (190, 20);
		\draw[fill=green!20!white] (190,10) rectangle (200, 20);
		\draw[fill=green!20!white] (200,10) rectangle (210, 20);
		\draw[fill=green!20!white] (210,10) rectangle (220, 20);

		\draw[fill=gray!10!white] (180,20) rectangle (190, 30);
		\draw[fill=gray!10!white] (190,20) rectangle (200, 30);
		\draw[fill=gray!10!white] (200,20) rectangle (210, 30);
		\draw[fill=gray!10!white] (210,20) rectangle (220, 30);

		% Multiplied by
		\draw[black, very thick] (195,40) -- (205, 50);
		\draw[black, very thick] (195,50) -- (205, 40);

		% Posterior beliefs
		\draw[fill=green!20!white] (180,60) rectangle (190, 70);
		\draw[fill=green!20!white] (190,60) rectangle (200, 70);
		\draw[fill=green!20!white] (200,60) rectangle (210, 70);
		\draw[fill=green!20!white] (210,60) rectangle (220, 70);
		
		% Labels
		\draw [decoration={brace, amplitude=6pt, raise=15pt, aspect=0.5, mirror}, decorate] (-3, 3) -- (43, 3) node [below=18pt, pos=0.5] {};
		\node[scale=5] at (20, -10) {\Huge New Q-values};

		\draw [decoration={brace, amplitude=6pt, raise=15pt, aspect=0.5, mirror}, decorate] (67, 3) -- (113, 3) node [below=18pt, pos=0.5] {};
		\node[scale=5] at (90, -10) {\Huge Old Q-values};

		\draw [decoration={brace, amplitude=6pt, raise=15pt, aspect=0.5, mirror}, decorate] (137, 3) -- (153, 3) node [below=18pt, pos=0.5] {};
		\node[scale=5] at (145, -10) {\Huge Learning rate};

		\draw [decoration={brace, amplitude=6pt, raise=15pt, aspect=0.5, mirror}, decorate] (177, 3) -- (223, 3) node [below=18pt, pos=0.5] {};
		\node[scale=5] at (200, -10) {\Huge Temporal difference};

		\draw [decoration={brace, amplitude=6pt, raise=15pt, aspect=0.5}, decorate] (177, 67) -- (223, 67) node [below=18pt, pos=0.5] {};
		\node[scale=5] at (200, 80) {\Huge Posterior Beliefs};

		% States and actions dimensions
		\draw[<->] (0, 35) -- (40, 35) node [below=18pt, pos=0.5] {};
		\node[scale=5] at (20, 40) {\Huge States};

		\draw[<->] (-5, 0) -- (-5, 30) node [below=18pt, pos=0.5] {};
		\node[rotate=90, scale=5] at (-10, 15) {\Huge Actions};

	\end{tikzpicture}
	\vspace{-0.25cm}
	\caption{This figure illustrates the computation of the updated Q-values. More precisely, the row of temporal difference corresponding to the action selected is multiplied element-wise with the posterior beliefs. The resulting product is multiplied with the learning rate and added to the old-Q-values.}
	\label{fig:Q-values}
\end{figure}

While Q-learning assumes that states are observable, the discrete states available to our planner are the latent variables $\bm{Z}_\tau$ over which the agent has posterior beliefs. The question then becomes: \textit{how should we adapt Q-learning to work with stochastic states?} We propose to scale the temporal difference in Equation \eqref{eq:temproal_diff_in_Q_learning} by the posterior probability of the state, more precisely:
\begin{align}
	q(a_t, s_t) &\leftarrow q(a_t, s_t) + \alpha Q(s_t) \Bigg[\underbrace{r_{t + 1} + \gamma \sum_{s_{t+1}} P(s_{t+1} | s_t, a_t) \big[ \max_{a_{t+1}} q(a_{t+1}, s_{t+1}) \big] - q(a_t, s_t)}_{\text{Temporal difference}}\Bigg], \label{eq:stochastic_q_values}
\end{align}
where $q(a_t, s_t)$ are the estimated Q-values of taking action $a_t$ in state $s_t$, while $Q(s_t)$ is the posterior probability of state $s_t$. Intuitively, the temporal difference can be understood as the error between the estimated Q-values and the target Q-values. Given an action, we can compute the temporal difference for each state, and by multiplying the temporal difference by the state probability, we effectively perform a form of credit assignment. Note, if all of the probability mass is concentrated in one state, then the above equation reduces to the standard Q-learning. Importantly, the above equation needs to be applied for all states $s_t$, Figure \ref{fig:Q-values} illustrates the computation of the updated Q-values. Finally, while Equation \eqref{eq:stochastic_q_values} used the notation from reinforcement learning, we can re-write this equation using our paper's notation:
\begin{align}
	q(a_t, \bm{z}_t) &\leftarrow q(a_t, \bm{z}_t) + \alpha Q(\bm{z}_t) \Bigg[r_{t + 1} + \gamma \sum_{\bm{z}_{t+1}} P(\bm{z}_{t+1} | \bm{z}_t, a_t) \big[ \max_{a_{t+1}} q(a_{t+1}, \bm{z}_{t+1}) \big] - q(a_t, \bm{z}_t)\Bigg],
\end{align}
where $s_\tau$ is now $z_\tau$, $Q(\bm{z}_t)$ is the posterior distribution obtained using the perception model from the current observation, and $P(\bm{z}_{t+1} | \bm{z}_t, a_t)$ are the transition probabilities obtained from the transition model.

\section{Experiments} \label{sec:experiments}

\begin{figure}[H]
	\centering
	\subfigure[]{
		\begin{tikzpicture}[scale=0.04, every node/.style={scale=0.04}]
			% Maze 8
			\draw[green!20!white, fill=green!20!white] (5,5.25) rectangle (45,14.75);
			\draw[green!20!white, fill=green!20!white] (35,45.25) rectangle (45,14.75);
			\draw[green!20!white, fill=green!20!white] (25,35.25) rectangle (35,25.25);
			\draw[black] (10,10) rectangle (40,10);
			\draw[black] (40,10) rectangle (40,30);
			\draw[black] (30,30) rectangle (40,30);
			\node at (0, 0) (x2) {\includegraphics[width=290px]{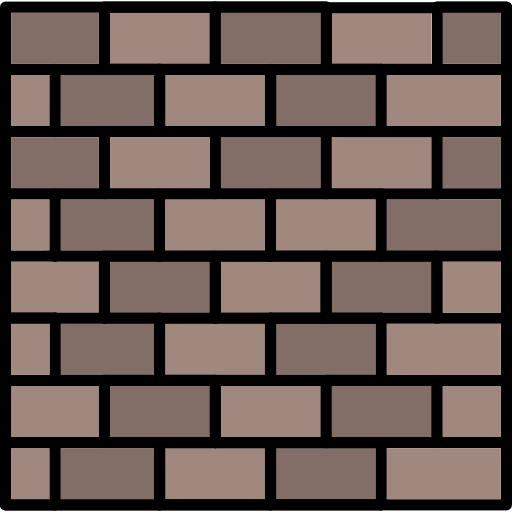}};
			\node at (10, 0) (x2) {\includegraphics[width=290px]{figures/wall.png}};
			\node at (20, 0) (x2) {\includegraphics[width=290px]{figures/wall.png}};
			\node at (30, 0) (x2) {\includegraphics[width=290px]{figures/wall.png}};
			\node at (40, 0) (x2) {\includegraphics[width=290px]{figures/wall.png}};
			\node at (50, 0) (x2) {\includegraphics[width=290px]{figures/wall.png}};
			\node at (0, 10) (x2) {\includegraphics[width=290px]{figures/wall.png}};
			\node at (10, 10) (x2) {\includegraphics[width=290px]{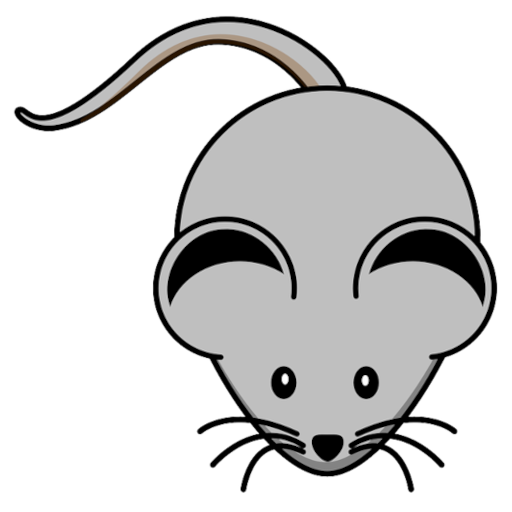}};
			\node at (50, 10) (x2) {\includegraphics[width=290px]{figures/wall.png}};
			\node at (0, 20) (x2) {\includegraphics[width=290px]{figures/wall.png}};
			\node at (10, 20) (x2) {\includegraphics[width=290px]{figures/wall.png}};
			\node at (20, 20) (x2) {\includegraphics[width=290px]{figures/wall.png}};
			\node at (30, 20) (x2) {\includegraphics[width=290px]{figures/wall.png}};
			\node at (50, 20) (x2) {\includegraphics[width=290px]{figures/wall.png}};
			\node at (20, 30) (x2) {\includegraphics[width=290px]{figures/wall.png}};
			\node at (30, 30) (x2) {\includegraphics[width=290px]{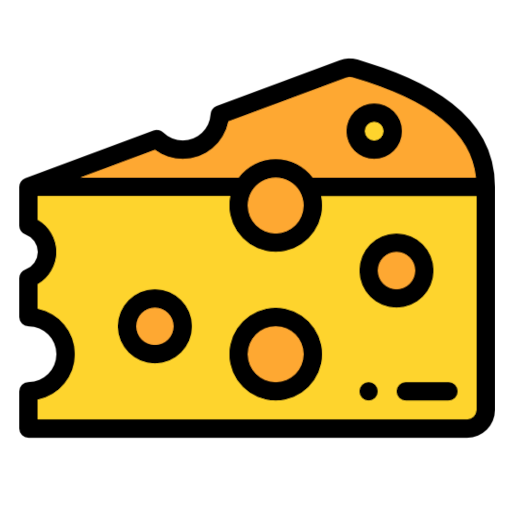}};
			\node at (50, 30) (x2) {\includegraphics[width=290px]{figures/wall.png}};
			\node at (20, 40) (x2) {\includegraphics[width=290px]{figures/wall.png}};
			\node at (30, 40) (x2) {\includegraphics[width=290px]{figures/wall.png}};
			\node at (50, 40) (x2) {\includegraphics[width=290px]{figures/wall.png}};
			\node at (30, 50) (x2) {\includegraphics[width=290px]{figures/wall.png}};
			\node at (40, 50) (x2) {\includegraphics[width=290px]{figures/wall.png}};
			\node at (50, 50) (x2) {\includegraphics[width=290px]{figures/wall.png}};
		\end{tikzpicture}\label{fig:maze_8}
	}
	\subfigure[]{
		\begin{tikzpicture}[scale=0.04, every node/.style={scale=0.04}]
			% Maze 10
			\draw[green!20!white, fill=green!20!white] (75,5.25) rectangle (125,14.75);
			\draw[green!20!white, fill=green!20!white] (105,25.25) rectangle (125,34.75);
			\draw[green!20!white, fill=green!20!white] (115,14.75) rectangle (125,25);
			\draw[red!20!white, fill=red!20!white] (75,25.25) rectangle (105,34.75);
			\draw[red!20!white, fill=red!20!white] (75,35) rectangle (125,94.75);
			\node at (70, 0) (x2) {\includegraphics[width=290px]{figures/wall.png}};
			\node at (80, 0) (x2) {\includegraphics[width=290px]{figures/wall.png}};
			\node at (90, 0) (x2) {\includegraphics[width=290px]{figures/wall.png}};
			\node at (100, 0) (x2) {\includegraphics[width=290px]{figures/wall.png}};
			\node at (110, 0) (x2) {\includegraphics[width=290px]{figures/wall.png}};
			\node at (120, 0) (x2) {\includegraphics[width=290px]{figures/wall.png}};
			\node at (130, 0) (x2) {\includegraphics[width=290px]{figures/wall.png}};
			\node at (70, 10) (x2) {\includegraphics[width=290px]{figures/wall.png}};
			\node at (80, 10) (x2) {\includegraphics[width=290px]{figures/mouse.png}};
			\node at (130, 10) (x2) {\includegraphics[width=290px]{figures/wall.png}};
			\node at (70, 20) (x2) {\includegraphics[width=290px]{figures/wall.png}};
			\node at (80, 20) (x2) {\includegraphics[width=290px]{figures/wall.png}};
			\node at (90, 20) (x2) {\includegraphics[width=290px]{figures/wall.png}};
			\node at (100, 20) (x2) {\includegraphics[width=290px]{figures/wall.png}};
			\node at (110, 20) (x2) {\includegraphics[width=290px]{figures/wall.png}};
			\node at (130, 20) (x2) {\includegraphics[width=290px]{figures/wall.png}};
			\node at (70, 30) (x2) {\includegraphics[width=290px]{figures/wall.png}};
			\node at (130, 30) (x2) {\includegraphics[width=290px]{figures/wall.png}};
			\node at (70, 40) (x2) {\includegraphics[width=290px]{figures/wall.png}};
			\node at (90, 40) (x2) {\includegraphics[width=290px]{figures/wall.png}};
			\node at (100, 40) (x2) {\includegraphics[width=290px]{figures/wall.png}};
			\node at (110, 40) (x2) {\includegraphics[width=290px]{figures/wall.png}};
			\node at (120, 40) (x2) {\includegraphics[width=290px]{figures/wall.png}};
			\node at (130, 40) (x2) {\includegraphics[width=290px]{figures/wall.png}};
			\node at (70, 50) (x2) {\includegraphics[width=290px]{figures/wall.png}};
			\node at (130, 50) (x2) {\includegraphics[width=290px]{figures/wall.png}};
			\node at (70, 60) (x2) {\includegraphics[width=290px]{figures/wall.png}};
			\node at (80, 60) (x2) {\includegraphics[width=290px]{figures/wall.png}};
			\node at (90, 60) (x2) {\includegraphics[width=290px]{figures/wall.png}};
			\node at (100, 60) (x2) {\includegraphics[width=290px]{figures/wall.png}};
			\node at (110, 60) (x2) {\includegraphics[width=290px]{figures/wall.png}};
			\node at (130, 60) (x2) {\includegraphics[width=290px]{figures/wall.png}};
			\node at (70, 70) (x2) {\includegraphics[width=290px]{figures/wall.png}};
			\node at (130, 70) (x2) {\includegraphics[width=290px]{figures/wall.png}};
			\node at (70, 80) (x2) {\includegraphics[width=290px]{figures/wall.png}};
			\node at (90, 80) (x2) {\includegraphics[width=290px]{figures/wall.png}};
			\node at (100, 80) (x2) {\includegraphics[width=290px]{figures/wall.png}};
			\node at (110, 80) (x2) {\includegraphics[width=290px]{figures/wall.png}};
			\node at (120, 80) (x2) {\includegraphics[width=290px]{figures/wall.png}};
			\node at (130, 80) (x2) {\includegraphics[width=290px]{figures/wall.png}};
			\node at (70, 90) (x2) {\includegraphics[width=290px]{figures/wall.png}};
			\node at (120, 90) (x2) {\includegraphics[width=290px]{figures/chesse.png}};
			\node at (130, 90) (x2) {\includegraphics[width=290px]{figures/wall.png}};
			\node at (70, 100) (x2) {\includegraphics[width=290px]{figures/wall.png}};
			\node at (80, 100) (x2) {\includegraphics[width=290px]{figures/wall.png}};
			\node at (90, 100) (x2) {\includegraphics[width=290px]{figures/wall.png}};
			\node at (100, 100) (x2) {\includegraphics[width=290px]{figures/wall.png}};
			\node at (110, 100) (x2) {\includegraphics[width=290px]{figures/wall.png}};
			\node at (120, 100) (x2) {\includegraphics[width=290px]{figures/wall.png}};
			\node at (130, 100) (x2) {\includegraphics[width=290px]{figures/wall.png}};
		\end{tikzpicture}\label{fig:maze_10}
	}
	\subfigure[]{
		\begin{tikzpicture}[scale=0.04, every node/.style={scale=0.04}]
			% Maze 20
			\draw[green!20!white, fill=green!20!white] (435,25.25) rectangle (475,34.75);
			\draw[green!20!white, fill=green!20!white] (485,5.25) rectangle (475,54.75);
			\draw[black] (440,30) rectangle (480,30);
			\draw[black] (480,10) rectangle (480,30);
			\node at (470, 0) (x2) {\includegraphics[width=290px]{figures/wall.png}};
			\node at (480, 0) (x2) {\includegraphics[width=290px]{figures/wall.png}};
			\node at (490, 0) (x2) {\includegraphics[width=290px]{figures/wall.png}};
			\node at (470, 10) (x2) {\includegraphics[width=290px]{figures/wall.png}};
			\node at (480, 10) (x2) {\includegraphics[width=290px]{figures/chesse.png}};
			\node at (490, 10) (x2) {\includegraphics[width=290px]{figures/wall.png}};
			\node at (430, 20) (x2) {\includegraphics[width=290px]{figures/wall.png}};
			\node at (440, 20) (x2) {\includegraphics[width=290px]{figures/wall.png}};
			\node at (450, 20) (x2) {\includegraphics[width=290px]{figures/wall.png}};
			\node at (460, 20) (x2) {\includegraphics[width=290px]{figures/wall.png}};
			\node at (470, 20) (x2) {\includegraphics[width=290px]{figures/wall.png}};
			\node at (490, 20) (x2) {\includegraphics[width=290px]{figures/wall.png}};
			\node at (430, 30) (x2) {\includegraphics[width=290px]{figures/wall.png}};
			\node at (440, 30) (x2) {\includegraphics[width=290px]{figures/mouse.png}};
			\node at (490, 30) (x2) {\includegraphics[width=290px]{figures/wall.png}};
			\node at (430, 40) (x2) {\includegraphics[width=290px]{figures/wall.png}};
			\node at (440, 40) (x2) {\includegraphics[width=290px]{figures/wall.png}};
			\node at (450, 40) (x2) {\includegraphics[width=290px]{figures/wall.png}};
			\node at (460, 40) (x2) {\includegraphics[width=290px]{figures/wall.png}};
			\node at (470, 40) (x2) {\includegraphics[width=290px]{figures/wall.png}};
			\node at (490, 40) (x2) {\includegraphics[width=290px]{figures/wall.png}};
			\node at (470, 50) (x2) {\includegraphics[width=290px]{figures/wall.png}};
			\node at (490, 50) (x2) {\includegraphics[width=290px]{figures/wall.png}};
			\node at (470, 60) (x2) {\includegraphics[width=290px]{figures/wall.png}};
			\node at (480, 60) (x2) {\includegraphics[width=290px]{figures/wall.png}};
			\node at (490, 60) (x2) {\includegraphics[width=290px]{figures/wall.png}};
		\end{tikzpicture}\label{fig:maze_20}
	}
	\subfigure[]{
		\begin{tikzpicture}[scale=0.04, every node/.style={scale=0.04}]
			% Maze 25
			\draw[green!20!white, fill=green!20!white] (505,5.25) rectangle (535,34.75);
			\draw[black] (510,10) rectangle (520,10);
			\draw[black] (520,10) rectangle (520,30);
			\draw[black] (520,30) rectangle (530,30);
			\node at (500, 0) (x2) {\includegraphics[width=290px]{figures/wall.png}};
			\node at (510, 0) (x2) {\includegraphics[width=290px]{figures/wall.png}};
			\node at (520, 0) (x2) {\includegraphics[width=290px]{figures/wall.png}};
			\node at (530, 0) (x2) {\includegraphics[width=290px]{figures/wall.png}};
			\node at (540, 0) (x2) {\includegraphics[width=290px]{figures/wall.png}};
			\node at (500, 10) (x2) {\includegraphics[width=290px]{figures/wall.png}};
			\node at (510, 10) (x2) {\includegraphics[width=290px]{figures/mouse.png}};
			\node at (540, 10) (x2) {\includegraphics[width=290px]{figures/wall.png}};
			\node at (500, 20) (x2) {\includegraphics[width=290px]{figures/wall.png}};
			\node at (540, 20) (x2) {\includegraphics[width=290px]{figures/wall.png}};
			\node at (500, 30) (x2) {\includegraphics[width=290px]{figures/wall.png}};
			\node at (530, 30) (x2) {\includegraphics[width=290px]{figures/chesse.png}};
			\node at (540, 30) (x2) {\includegraphics[width=290px]{figures/wall.png}};
			\node at (500, 40) (x2) {\includegraphics[width=290px]{figures/wall.png}};
			\node at (510, 40) (x2) {\includegraphics[width=290px]{figures/wall.png}};
			\node at (520, 40) (x2) {\includegraphics[width=290px]{figures/wall.png}};
			\node at (530, 40) (x2) {\includegraphics[width=290px]{figures/wall.png}};
			\node at (540, 40) (x2) {\includegraphics[width=290px]{figures/wall.png}};
		\end{tikzpicture}\label{fig:maze_25}
	}
	\subfigure[]{
		\begin{tikzpicture}[scale=0.04, every node/.style={scale=0.04}]
			% Maze 26
			\draw[green!20!white, fill=green!20!white] (5,105.25) rectangle (45,144.75);
			\draw[black] (10,110.25) rectangle (20,110.25);
			\draw[black] (20,120.25) rectangle (20,110.25);
			\draw[black] (20,120.25) rectangle (40,120.25);
			\draw[black] (40,140.25) rectangle (40,120.25);
			\node at (0, 100) (x2) {\includegraphics[width=290px]{figures/wall.png}};
			\node at (10, 100) (x2) {\includegraphics[width=290px]{figures/wall.png}};
			\node at (20, 100) (x2) {\includegraphics[width=290px]{figures/wall.png}};
			\node at (30, 100) (x2) {\includegraphics[width=290px]{figures/wall.png}};
			\node at (40, 100) (x2) {\includegraphics[width=290px]{figures/wall.png}};
			\node at (50, 100) (x2) {\includegraphics[width=290px]{figures/wall.png}};
			\node at (0, 110) (x2) {\includegraphics[width=290px]{figures/wall.png}};
			\node at (10, 110) (x2) {\includegraphics[width=290px]{figures/mouse.png}};
			\node at (50, 110) (x2) {\includegraphics[width=290px]{figures/wall.png}};
			\node at (0, 120) (x2) {\includegraphics[width=290px]{figures/wall.png}};
			\node at (50, 120) (x2) {\includegraphics[width=290px]{figures/wall.png}};
			\node at (0, 130) (x2) {\includegraphics[width=290px]{figures/wall.png}};
			\node at (50, 130) (x2) {\includegraphics[width=290px]{figures/wall.png}};
			\node at (0, 140) (x2) {\includegraphics[width=290px]{figures/wall.png}};
			\node at (40, 140) (x2) {\includegraphics[width=290px]{figures/chesse.png}};
			\node at (50, 140) (x2) {\includegraphics[width=290px]{figures/wall.png}};
			\node at (0, 150) (x2) {\includegraphics[width=290px]{figures/wall.png}};
			\node at (10, 150) (x2) {\includegraphics[width=290px]{figures/wall.png}};
			\node at (20, 150) (x2) {\includegraphics[width=290px]{figures/wall.png}};
			\node at (30, 150) (x2) {\includegraphics[width=290px]{figures/wall.png}};
			\node at (40, 150) (x2) {\includegraphics[width=290px]{figures/wall.png}};
			\node at (50, 150) (x2) {\includegraphics[width=290px]{figures/wall.png}};
		\end{tikzpicture}\label{fig:maze_26}
	}
	\subfigure[]{
		\begin{tikzpicture}[scale=0.04, every node/.style={scale=0.04}]
			% Maze 27
			\draw[green!20!white, fill=green!20!white] (5,5.25) rectangle (55,54.75);
			\draw[orange!30!white, fill=orange!30!white] (5,5.25) rectangle (15,35.75);
			\draw[orange!30!white, fill=orange!30!white] (5,44.75) rectangle (55,54.75);
			\draw[orange!30!white, fill=orange!30!white] (35,24.75) rectangle (45,54.75);
			\node at (0, 0) (x2) {\includegraphics[width=290px]{figures/wall.png}};
			\node at (10, 0) (x2) {\includegraphics[width=290px]{figures/wall.png}};
			\node at (20, 0) (x2) {\includegraphics[width=290px]{figures/wall.png}};
			\node at (30, 0) (x2) {\includegraphics[width=290px]{figures/wall.png}};
			\node at (40, 0) (x2) {\includegraphics[width=290px]{figures/wall.png}};
			\node at (50, 0) (x2) {\includegraphics[width=290px]{figures/wall.png}};
			\node at (60, 0) (x2) {\includegraphics[width=290px]{figures/wall.png}};
			\node at (0, 10) (x2) {\includegraphics[width=290px]{figures/wall.png}};
			\node at (10, 10) (x2) {\includegraphics[width=290px]{figures/mouse.png}};
			\node at (60, 10) (x2) {\includegraphics[width=290px]{figures/wall.png}};
			\node at (0, 20) (x2) {\includegraphics[width=290px]{figures/wall.png}};
			\node at (60, 20) (x2) {\includegraphics[width=290px]{figures/wall.png}};
			\node at (0, 30) (x2) {\includegraphics[width=290px]{figures/wall.png}};
			\node at (60, 30) (x2) {\includegraphics[width=290px]{figures/wall.png}};
			\node at (0, 40) (x2) {\includegraphics[width=290px]{figures/wall.png}};
			\node at (60, 40) (x2) {\includegraphics[width=290px]{figures/wall.png}};
			\node at (0, 50) (x2) {\includegraphics[width=290px]{figures/wall.png}};
			\node at (50, 50) (x2) {\includegraphics[width=290px]{figures/chesse.png}};
			\node at (60, 50) (x2) {\includegraphics[width=290px]{figures/wall.png}};
			\node at (0, 60) (x2) {\includegraphics[width=290px]{figures/wall.png}};
			\node at (10, 60) (x2) {\includegraphics[width=290px]{figures/wall.png}};
			\node at (20, 60) (x2) {\includegraphics[width=290px]{figures/wall.png}};
			\node at (30, 60) (x2) {\includegraphics[width=290px]{figures/wall.png}};
			\node at (40, 60) (x2) {\includegraphics[width=290px]{figures/wall.png}};
			\node at (50, 60) (x2) {\includegraphics[width=290px]{figures/wall.png}};
			\node at (60, 60) (x2) {\includegraphics[width=290px]{figures/wall.png}};
		\end{tikzpicture}\label{fig:maze_27}
	}
	\vspace{-0.25cm}
	\caption{This figure illustrates the mazes used throughout the experiment section. The starting position is represented by a mouse, while the goal state is represented by the cheese. Red cells correspond to positions for which no Gaussian components were learnt. In contrast, the model properly learnt Gaussian components for cells with green and orange backgrounds. However, components corresponding to orange cells became unstable as the number of states increased, and the model was unlearnt by the end of the simulation, i.e., orange cells are green cells that turned red. Finally, a line connects the starting position to the goal state if the agent successfully learnt to solve the task.}
	\label{fig:mazes}
\end{figure}

In this section, we validate our approach on a maze solving task. The agent is able to move ``up'', ``down'', ``left'' and ``right'', but if the action selected would lead the agent to bump into a wall, then the agent does not move. As illustrated in Figure \ref{fig:mazes}, the agent needs to navigate from the initial position (represented by the mouse) to the goal state (represented by the cheese). Then, when the goal state has been reached, the agent must perform the action ``eat''. We study the performance of our agent based on two criteria: (i) its ability to learn the structure of various mazes, and (ii) its ability to solve the mazes by getting to the goal state and eating the cheese.

First, we focus on the model's ability to learn the maze structure. By manual inspection of the learned components and transition matrices, we identified the cells which have been learnt properly (see green and orange cells in Figure \ref{fig:mazes}), and the cells that were not learned (see red cells in Figure \ref{fig:mazes}). Importantly, as the number of states increase, some components became unstable, effectively taking over the data points corresponding to neighbouring cells, the impacted cells are drawn in orange.

Second, we looked at which mazes can be solved by our approach. We found that mazes in Figure \ref{fig:maze_8}, \ref{fig:maze_20}, \ref{fig:maze_25}, and \ref{fig:maze_26}, can be solved by our approach. However, the agent was unable to solve the maze in Figure \ref{fig:maze_10}, because of a lack of exploration, leading to an inability to learn the entire maze structure. Similarly, the maze of Figure \ref{fig:maze_27} remained unsolved as the large number of states rendered the Gaussian components unstable, leading to some components being unlearned as time passes.

Another interesting question is how fast can the TGM agent learn to solve each maze. To answer this question, we rely on an empirical comparison between TGM and two reinforcement learning algorithms, i.e., Deep Q-network (DQN) and Advantage Actor Critic (A2C). Put simply, the DQN agent \citep{Mnih2015} learns the value of taking each action in each state, while A2C \citep{mnih2016asynchronous} is a policy-based approach equipped with a critic network. Note, DQN, A2C and PPO, all learn a mapping from states to actions (or values) without creating a model of the environment, i.e., they are model-free approaches.

\begin{figure}[H]
	\centering
	\subfigure[]{
		\includegraphics[width=200px]{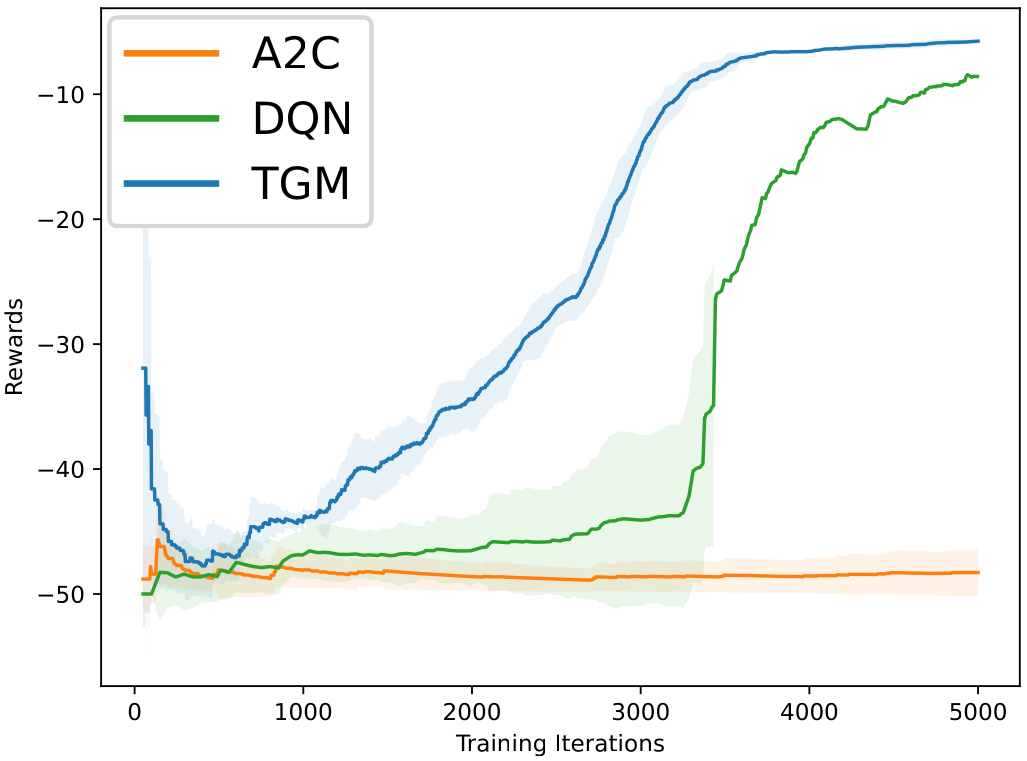} \label{fig:reward_maze_8}
	}
	\subfigure[]{
		\includegraphics[width=200px]{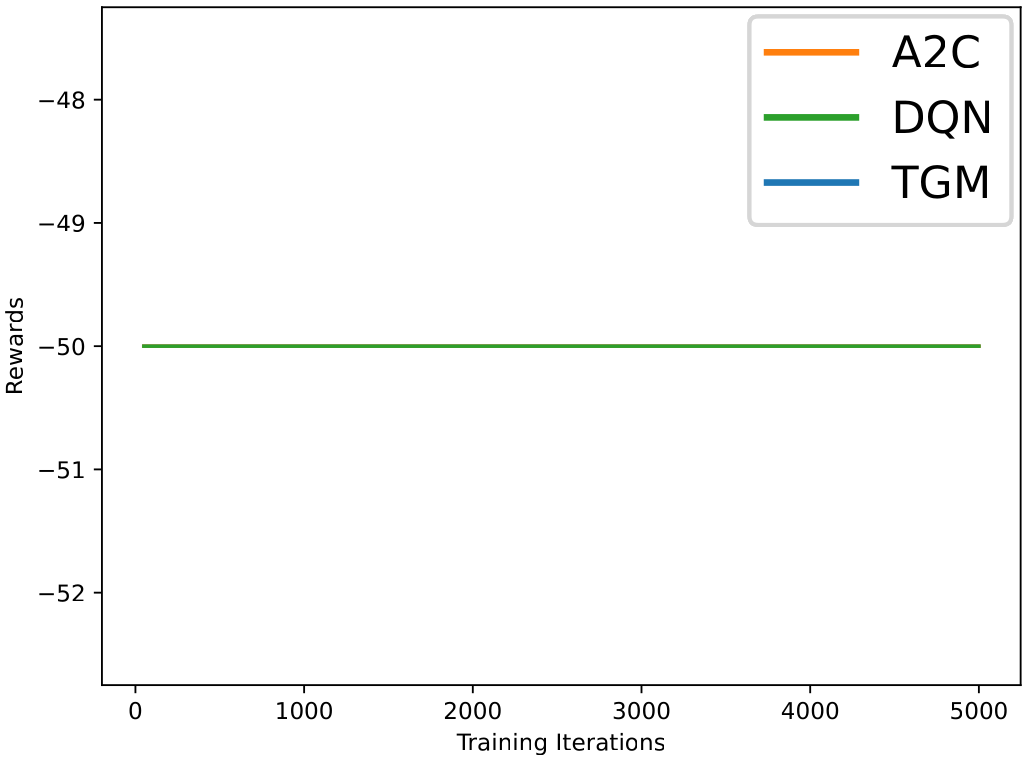} \label{fig:reward_maze_10}
	}
	\subfigure[]{
		\includegraphics[width=200px]{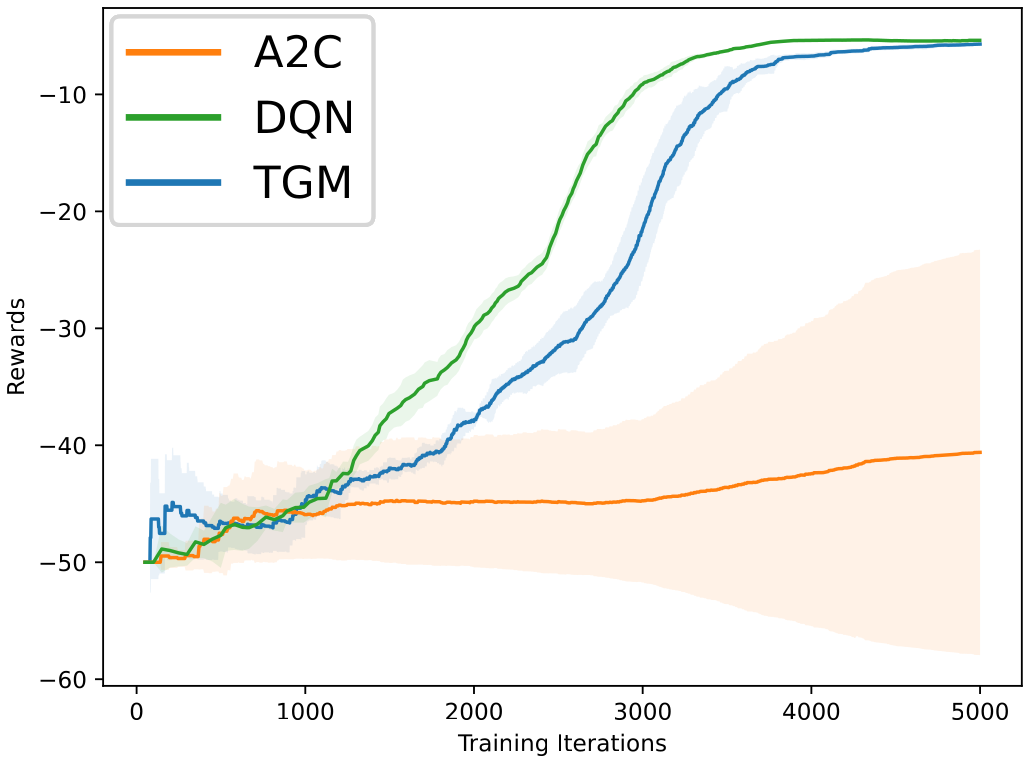} \label{fig:reward_maze_20}
	}
	\subfigure[]{
		\includegraphics[width=200px]{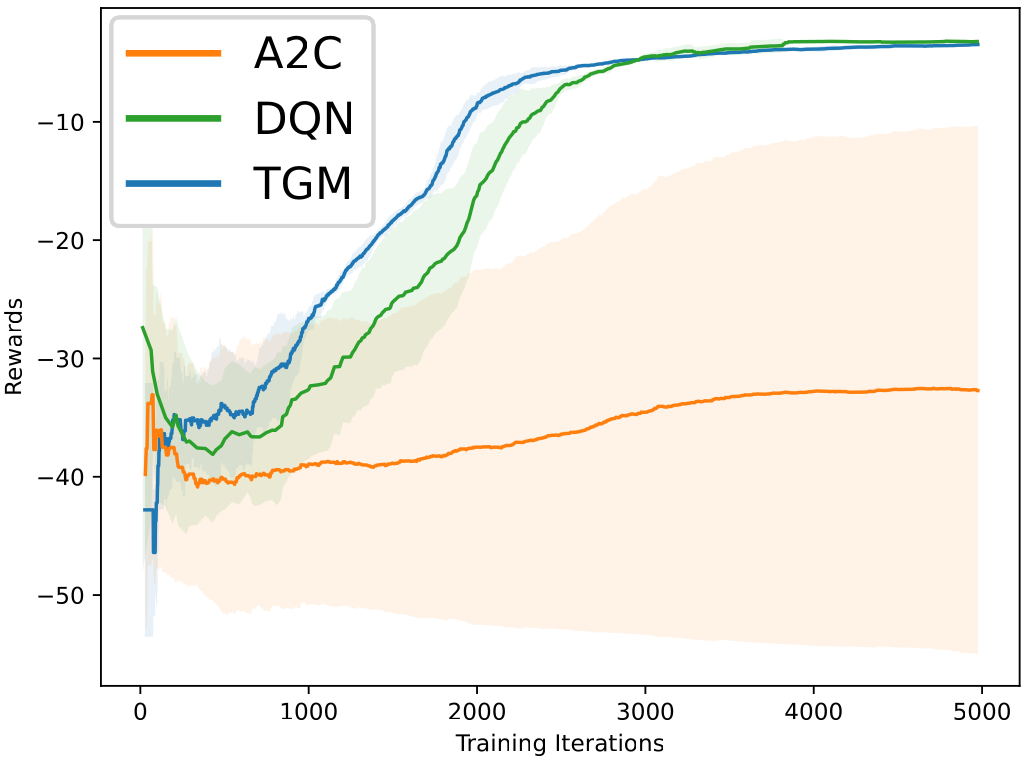} \label{fig:reward_maze_25}
	}
	\subfigure[]{
		\includegraphics[width=200px]{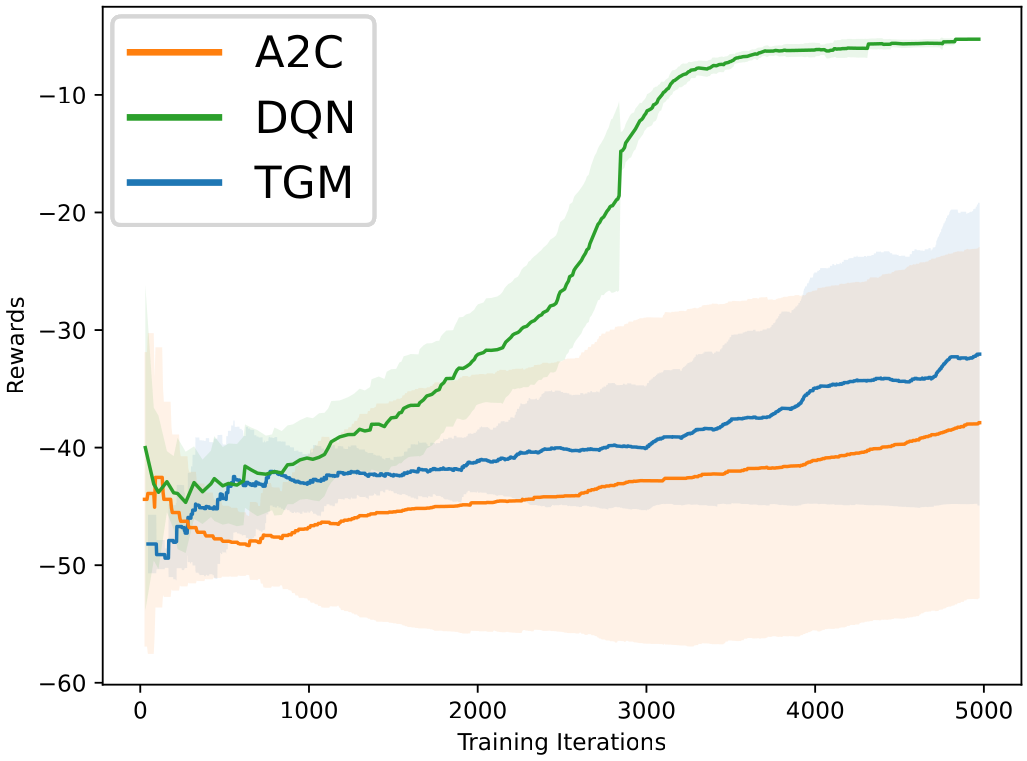} \label{fig:reward_maze_26}
	}
	\subfigure[]{
		\includegraphics[width=200px]{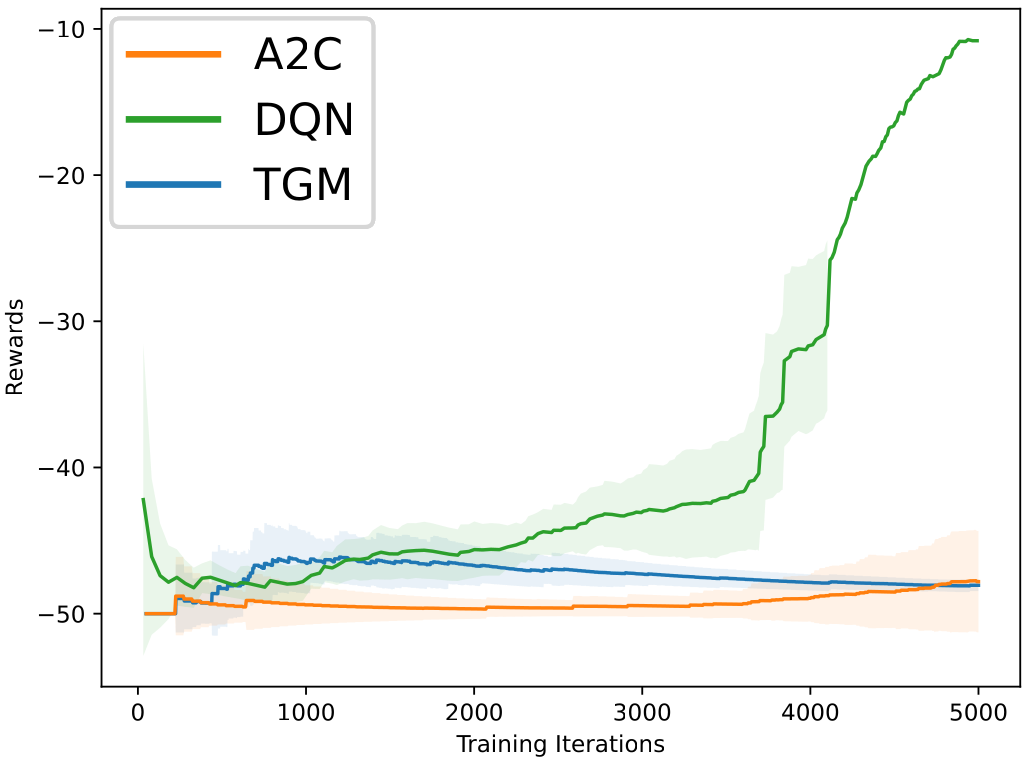} \label{fig:reward_maze_27}
	}
	\vspace{-0.25cm}
	\caption{This figure illustrates the average episodic reward gathered by the TGM, DQN and A2C agents on the corresponding mazes in Figure \ref{fig:mazes}. For example, the graph of Figure \ref{fig:reward_maze_8} corresponds to the maze in Figure \ref{fig:maze_8}. Similarly, the graph of Figure 13(x) corresponds to the maze in Figure \ref{fig:mazes}(x). Note, in Figure \ref{fig:reward_maze_10}, all three models performed the same, and therefore only DQN is visible. Finally, the shaded areas correspond to the standard deviation of each model.} \label{fig:reward_mazes}
\end{figure}

Figure \ref{fig:reward_mazes} illustrates the average episodic reward gathered by the TGM, DQN and A2C agents on the corresponding mazes of Figure \ref{fig:mazes}. Figure \ref{fig:reward_mazes} shows that TGM performs better than DQN and A2C in the maze of Figure \ref{fig:maze_8}. Interestingly, all three agents (i.e., TGM, DQN, and A2C) fail to solve the maze of Figure  \ref{fig:maze_10}. This suggests that despite a rather simple topology (i.e., a long corridor), this environment is challenging and would require a special exploration strategy. In other words, random selection of actions is extremely unlikely to enable the agent to reach the goal state, thus, the agent should actively seek to explore rarely visited parts of the environment.

Additionally, DQN seems to be learning faster that TGM on the maze of Figure \ref{fig:maze_20}, however, both agents reach a similar asymptotic mean reward. The opposite is true for the maze of Figure \ref{fig:maze_25}, where TGM learns faster than DQN but both agents reach a similar asymptotic mean reward. Note, on mazes \ref{fig:maze_20} and \ref{fig:maze_25}, A2C performs worse than both DQN and TGM.

Finally, DQN outperforms TGM and A2C on both mazes \ref{fig:maze_26} and \ref{fig:maze_27}. Importantly, TGM has a the small variance in Figure \ref{fig:reward_maze_27}, because the TGM agent always fails to solve maze \ref{fig:maze_27}. In contrast, the large variance in Figure \ref{fig:reward_maze_26} indicates that the TGM agent alternates between successfully solving the maze and failing to do so.

To conclude, we showed that the TGM agent is competitive with reinforcement learning benchmark such as DQN and A2C. More precisely, TGM outperforms A2C on all the tested mazes, except one where all agents failed to solve the task. This may be because TGM is more stable than A2C as illustrated in Figure \ref{fig:reward_maze_20} to \ref{fig:reward_maze_26}. Similarly, TGM performed better than DQN on one maze, similarly to DQN on three mazes, and worse than DQN on two mazes. However, while TGM sometimes performs worse than DQN, it learns an interpretable model of the environment, which can be desirable for some applications. Additionally, this model of the environment is composed of hidden states that may be useful to develop more advanced exploration strategies, which aim to explore rarely visited states.

\section{Conclusion} \label{sec:conclusion}

In this paper, we tackled the problem of learning the model structure from data, when the observations are continuous. More specifically, we aimed at identifying the number of clusters present in the data, and each cluster was associated to a latent state. As more data is gathered by the agent, new clusters are discovered, and the model needs to increase its number of states. In contrast, if the agent currently believes that there are more states, than the actual number of clusters in the data, then the number of states needs to decrease.

We proposed a variational Gaussian mixture model, in which the Gaussian mixture model is able to remove clusters not supported by the data, and the agent is constantly monitoring the data searching for new clusters. When new clusters are identified, the corresponding components are added to the mixture. These two mechanics allow the model to learn the number of components. While this is happening, the agent also learns the parameters of the individual components through variational inference.

However, inferring the current state is only part of the story, and the agent also needs to learn how different actions impact the temporal transition between states. This is done by taking advantage of a categorical-Dirichlet model, which effectively keeps track of the number of times the agent moved from a state to another when performing a specific action.

Once the agent can perform inference and predict the consequences of its action, the last requirement is to perform decision making and planning. We proposed a variant of the Q-learning algorithm, which accommodates stochastic states, i.e., having beliefs over states instead of observable states. This new algorithm enabled the agent to solve several mazes, by reaching the exit position and eating the cheese.

Experimentally, our approach was able to solve several mazes, but still has limitations. For example, the approach is susceptible to unlearning existing components, where several components fuse together into a single component. Additionally, since $\epsilon$-greedy is used to trade-off exploration and exploitation, the agent tends to struggle to explore remote parts of the maze. These limitations suggest that more research is required to increase the stability of the Gaussian components. Additionally, the agent would benefit from improving the exploration strategy, for example, the agent should probably focus on discovering rarely explored parts of the maze instead of randomly selecting action. Also, the environments being modelled were somewhat idealised, e.g., the organism follows a rather stereotyped trajectory, passing from one cluster to another.

Furthermore, one could investigate alternative approaches for learning the number of components such as the infinite Gaussian mixture model \citep{rasmussen1999infinite}. Another interesting direction of research would be to apply the temporal Gaussian mixture to study the latent representation learned in deep active inference \citep{DeepAIwithMCMC,catal2020learning,deconstructing_DAI,DeepAI,sancaktar2020endtoend,DAI_HR,DAI_HR2,DAI_POMDP}, or similarly to use the variational Gaussian mixture to study the representation learned by variational auto-encoders \citep{VAE,beta-VAE,Kingma2013,Rezende2014,bai2022gaussian,dilokthanakul2016deep}.

\newpage
\section*{Appendix A: notation and important properties}

In this paper, we made extensive use of several definitions and properties, which are summarized in this appendix. Appendix A.1 provides the definition of various probability distributions, Appendix A.2 presents properties related to probability theory, and Appendix A.3 highlights properties from linear algebra.

\subsection*{A.1: probability distributions}

The probability density function (PDF) of a multivariate Gaussian distribution with mean $\bm{\mu}$ and precision matrix $\bm{\Lambda}$ is given by:
\begin{align}
	\mathcal{N}(\bm{x}; \bm{\mu}, \bm{\Lambda}^{-1}) &= (2\pi)^{-\frac{K}{2}} |\bm{\Lambda}^{-1}|^{-\frac{1}{2}} \text{exp}\Big( -\frac{1}{2} (\bm{x} - \bm{\mu})^\top \bm{\Lambda} (\bm{x} - \bm{\mu}) \Big). \label{eq:gaussian_distribution}
\end{align}
The PDF of a Wishart distribution with $v$ degrees of freedom and scale matrix $\bm{W}$ is:
\begin{align}
	\mathcal{W}(\bm{\Lambda}; \bm{W}, v) &= 2^{-\frac{vK}{2}} |\bm{W}|^{-\frac{v}{2}} \Gamma_K(\tfrac{v}{2})^{-1} |\bm{\Lambda}|^{\frac{(v-K-1)}{2}} \text{exp}\Big( -\frac{1}{2} \text{tr}(\bm{W}^{-1}\bm{\Lambda}) \Big). \label{eq:wishart_distribution_def}
\end{align}
The PDF of a Dirichlet distribution with concentration parameters $d$ is defined as follows:
\begin{align}
	\text{Dir}(\bm{D}; d) = \frac{1}{B(d)}\prod_{k=1}^K \bm{D}^{d_k - 1}_k. \label{eq:dirichlet_distribution_def}
\end{align}
In the above equation, $\bm{D}$ was a vector. We define a Dirichlet distribution over a 3d-tensor $\bm{B}$, as a product of Dirichlet distributions over the columns of the matrices $\bm{B}[a]$. More precisely, the PDF of a Dirichlet distribution with concentration parameters $b$ is defined as:
\begin{align}
	\text{Dir}(\bm{B}; b) = \prod_{a=1}^A \prod_{j=1}^K \text{Dir}(\bm{B}[a]_{\bigcdot j}; b[a]_{\bigcdot j}) = \prod_{a=1}^A \prod_{j=1}^K \frac{1}{B(b[a]_{\bigcdot j})}\prod_{k=1}^K \bm{B}[a]^{b[a]_{kj} - 1}_{kj}.
\end{align}
The probability mass function (PMF) of a categorical distribution with parameters $\bm{D}$ is given by:
\begin{align}
	\text{Cat}(\bm{z}_n | \bm{D}) = \prod_{k=1}^K \bm{D}_{k}^{\bm{z}_{nk}}. \label{eq:categorical_distribution_def}
\end{align}
In the above equation, $\bm{D}$ is a vector, and the categorical distribution is a distribution over a single random variable. We extend the definition of categorical distributions to conditional distributions, by using 3d-tensor of parameters $\bm{B}$. More precisely, the PMF of a conditional categorical distribution with parameters $\bm{B}$ is:
\begin{align}
	\text{Cat}(\bm{z}^{\tau + 1}_n | \bm{z}^{\tau}_n, \bm{a}^{\tau}_n, \bm{B}) = \prod_{j=1}^K \prod_{k=1}^K \bm{B}[\bm{a}^{\tau}_n]_{jk}^{z^{\tau}_{nk}z^{\tau + 1}_{nj}}.
\end{align}

\subsection*{A.2: important properties of probability theory}

Given two sets of continuous random variables $\bm{X}$ and $\bm{Y}$, the sum-rule of probability states that:
\begin{align}
	P(\bm{Y}) = \int_{\bm{X}} P(\bm{X},\bm{Y}) \,\, d\bm{X},
\end{align}
and for discrete variables, the integral becomes a summation. The sum-rule can then be used to sum out random variables from a joint distribution. The second property is called the product-rule of probability, and can be used to split a joint distribution into conditional distributions. Given two sets of random variables $\bm{X}$ and $\bm{Y}$, the product-rule of probability states that:
\begin{align}
	P(\bm{X},\bm{Y}) = P(\bm{X}|\bm{Y}) P(\bm{Y}).
\end{align}
A corollary of the product-rule of probability is called Bayes theorem:
\begin{align}
	P(\bm{Y}|\bm{X}) = \frac{P(\bm{X}|\bm{Y}) P(\bm{Y})}{P(\bm{X})},
\end{align}
which states that the posterior $P(\bm{Y}|\bm{X})$ is equal to the likelihood $P(\bm{X}|\bm{Y})$ times the prior $P(\bm{Y})$ divided by the evidence $P(\bm{X})$. Next, if $\bm{a}$ and $\bm{b}$ are two real numbers, then a relevant property of logarithm is the following: 
\begin{align}
	\ln(\bm{a} \bm{b}) = \ln(\bm{a}) + \ln(\bm{b}).
\end{align}
Put simply, this allows us to turn the logarithm of a product into a sum of logarithms, and we will refer to the above equation as the ``log-property". Another property is called the linearity of expectation. Given a random variable $\bm{X}$, and two real numbers $\bm{a}$ and $\bm{b}$, the linearity of expectation states that:
\begin{align}
	\mathbb{E}[\bm{a}\bm{X} + \bm{b}] = \bm{a}\mathbb{E}[\bm{X}] + \bm{b},
\end{align}
where the expectation is w.r.t. the marginal distribution over $\bm{X}$, i.e., $P(\bm{X})$. Additionally, given two distributions $Q(X)$ and $P(X)$ over the same domain, the KL-divergence is an asymmetric distance that quantifies how different two distributions are from each other. The KL-divergence \citep{kullback1951information} is defined as:
\begin{align}
	\kl{Q(X)}{P(X)} = \mathbb{E}_{Q(X)}[\ln Q(X) - \ln P(X)].
\end{align}
Additionally, for discrete variables, the entropy of a conditional distribution decreases as more variables are given:
\begin{align}
	H[X|Y,Z] = -\sum_{x,y,z}P(x,y,z) \ln \frac{P(x,y,z)}{P(y,z)} \leq -\sum_{x,z}P(x,z) \ln \frac{P(x,z)}{P(z)} = H[X|Z]. \label{eq:conditional_entropy}
\end{align}
Moreover, according to Section of 10.2 of \citep{bishop2006pattern}, if $\bm{\Lambda} \sim Q(\bm{\Lambda}) = \mathcal{W}(\bm{\Lambda}; \bm{\hat{W}}, \hat{v})$ and $\bm{\mu} \sim Q(\bm{\mu}| \bm{\Lambda}) = \mathcal{N}(\bm{\mu}; \bm{\hat{m}}, (\hat{\beta}\bm{\Lambda})^{-1})$, then the following two properties holds:
\begin{align}
	\mathbb{E}_{Q(\bm{\Lambda})}[\ln|\bm{\Lambda}|] &= K \ln 2 + \ln |\bm{\hat{W}}| + \sum_{i=1}^K \psi (\tfrac{\hat{v}+1-i}{2}), \label{eq:wishart_ln_expectation} \\
	\mathbb{E}_{Q(\bm{\mu}, \bm{\Lambda})}[(\bm{x} - \bm{\mu})^\top\bm{\Lambda}(\bm{x} - \bm{\mu})] &= K \hat{\beta}^{-1} + \hat{v}(\bm{x} - \bm{\hat{m}})^\top\bm{\hat{W}}(\bm{x} - \bm{\hat{m}}), \label{eq:wishart_gaussian_expectation_of_quadratic_form}
\end{align}
where $\bm{\Lambda}$ and $\bm{\hat{W}}$ are $K \times K$ matrices, and $\psi(\,\cdot\,)$ is the digamma function. Also, if $\bm{D}$ is a random vector of size $K$ distributed according to a Dirichlet distribution, i.e., $\bm{D} \sim Q(\bm{D}) = \text{Dir}(\bm{D}; \hat{d})$, then:
\begin{align}
	\mathbb{E}_{Q(\bm{D})}[\ln \bm{D}_k] &= \psi (\hat{d}_k) - \psi (\textstyle\sum_{i=1}^K \hat{d}_i). \label{eq:dirichlet_ln_expectation}
\end{align}

\subsection*{A.3: important properties of linear algebra}

Given a symmetric matrix $X$, and two vectors $a$ and $b$, the following two properties hold:
\begin{align}
	(a - b)^\top X (a - b) &= a^\top X a -2 a^\top X b + b^\top X b \label{eq:gaussian_form_expansion} \\
	a^\top X b &= b^\top X a \label{eq:quadratic_form_symmetry}
\end{align}
Given a symmetric matrix $X$, $N$ vectors $a_i$, and a vector $b$, the following three properties hold:
\begin{align}
	\sum_{i=1}^N a_i^\top X b &= N \bar{a}^\top X b, \label{eq:avg_quadratic_1_form} \\
	\sum_{i=1}^N a_i b^\top &= N \bar{a} b^\top, \label{eq:avg_outter_product_1} \\
	\sum_{i=1}^N b a_i^\top &= N b \bar{a}^\top, \label{eq:avg_outter_product_2}
\end{align}
where $\bar{a} = \frac{1}{N} \sum_{i=1}^N a_i$.
\noindent Given a matrix $X$, and two vectors $a$ and $b$:
\begin{align}
	a^\top X b = tr(ba^\top X), \label{eq:trace_and_quadratic_form}
\end{align}
where $tr(X) = \sum_i X_{ii}$ is the trace $X$.
Given two matrices $A$ and $B$:
\begin{align}
	tr(A) + tr(B) = tr(A + B). \label{eq:trace_addition}
\end{align}
Given two vectors $a$ and $b$:
\begin{align}
	(a - b)(a - b)^\top = aa^\top + bb^\top - ab^\top - ba^\top, \label{eq:outter_product_neg} \\
	(a + b)(a + b)^\top = aa^\top + bb^\top + ab^\top + ba^\top. \label{eq:outter_product_pos}
\end{align}
Given two vectors $a$ and $b$, and two constants $c$ and $d$:
\begin{align}
	(ca)(db)^\top = cdab^\top. \label{eq:outter_product_with_constants}
\end{align}
Additionally, given a matrix $\bm{\Lambda}$ the following property holds:
\begin{align}
	|\bm{\Lambda}^{-1}| = |\bm{\Lambda}|^{-1}, \label{eq:determinant_of_inverse}
\end{align}
where $|\bm{\Lambda}|$ is the determinant of $\bm{\Lambda}$.

\newpage
\section*{Appendix B: Variational free energy proofs}

In this appendix, we explain how to compute the expectations of the variational free energy presented in Section \ref{sssec:vfe}. More precisely, we derive the following analytical solutions:
\begin{align*}
	\mathbb{E}_{Q(\bm{D})}[\ln Q(\bm{D})] &= - \ln B(\hat{d}) + \sum_{k=1}^K (\hat{d}_k - 1)\ln\bm{\tilde{D}}_k, \\
	\mathbb{E}_{Q(\bm{\mu}, \bm{\Lambda})}[\ln Q(\bm{\mu} | \bm{\Lambda})] &= \frac{1}{2} \sum_{k=1}^K K\ln \frac{\hat{\beta}_k}{2\pi} + \ln \bm{\tilde{\Lambda}}_k - K, \\
	\mathbb{E}_{Q(\bm{\Lambda})}[\ln Q(\bm{\Lambda})] &= \sum_{k=1}^K - \frac{\hat{v}_kK}{2}\ln 2 - \frac{\hat{v}_k}{2}\ln|\bm{\hat{W}}_k| - \ln \Gamma_K(\frac{\hat{v}_k}{2}) + \frac{\hat{v}_k - K - 1}{2} \ln \bm{\tilde{\Lambda}}_k - \frac{\hat{v}_kK}{2}, \\
	\mathbb{E}_{Q(\bm{Z})}[\ln Q(\bm{Z})] &= \sum_{n=1}^N \sum_{k=1}^K \bm{\hat{r}}_{nk} \ln \bm{\hat{r}}_{nk}, \\
	\mathbb{E}_{Q(\bm{D})}[\ln P(\bm{D})] &= - \ln B(d) + \sum_{k=1}^K (d_k - 1)\ln\bm{\tilde{D}}_k, \\
	\mathbb{E}_{Q(\bm{\mu}, \bm{\Lambda})}[\ln P(\bm{\mu} | \bm{\Lambda})] &= \frac{1}{2} \sum_{k=1}^K K\ln \frac{\beta_k}{2\pi} + \ln \bm{\tilde{\Lambda}}_k - \frac{\beta_k K}{\hat{\beta}_k} - \beta_k\hat{v}_k (\bm{m}_k - \hat{\bm{m}}_k)^\top \hat{\bm{W}}_k (\bm{m}_k - \hat{\bm{m}}_k), \\
	\mathbb{E}_{Q(\bm{Z}, \bm{D})}[\ln P(\bm{Z}|\bm{D})] &= \sum_{k=1}^K N_{k} \ln\bm{\tilde{D}}_k, \\
	\mathbb{E}_{Q(\bm{\Lambda})}[\ln P(\bm{\Lambda})] &= \\
	&\hspace{-1.35cm}\sum_{k=1}^K - \frac{v_kK}{2}\ln 2 - \frac{v_k}{2}\ln|\bm{W}_k| - \ln \Gamma_K(\frac{v_k}{2}) + \frac{v_k - K - 1}{2} \ln \bm{\tilde{\Lambda}}_k - \frac{\hat{v}_k}{2}tr(\bm{W}^{-1}_k\hat{\bm{W}}_k), \\
	\mathbb{E}_{Q(\bm{Z}, \bm{\mu}, \bm{\Lambda})}[\ln P(\bm{X} | \bm{Z}, \bm{\mu}, \bm{\Lambda})] &= \\
	\sum_{k=1}^K \frac{N_k}{2} &\Bigg[\ln \bm{\tilde{\Lambda}}_k - K\hat{\beta}_k^{-1} -\hat{v}_k tr(\bm{S}_k\bm{\hat{W}}_k) - \hat{v}_k(\bm{\bar{x}}_k - \bm{\hat{m}}_k)^\top \bm{\hat{W}}_k (\bm{\bar{x}}_k - \bm{\hat{m}}_k) -K \ln 2\pi \Bigg].
\end{align*}

\newpage
\subsection*{B.1 Derivation for $\boldmath{\mathbb{E}_{Q(\bm{D})}[\ln Q(\bm{D})]}$}

Let's first show that if $Q(\bm{D}) = \text{Dir}(\bm{D};\hat{d})$, then:
\begin{align*}
	\mathbb{E}_{Q(\bm{D})}[\ln Q(\bm{D})] &= - \ln B(\hat{d}) + \sum_{k=1}^K (\hat{d}_k - 1)\ln\bm{\tilde{D}}_k,
\end{align*}
where $\ln\bm{\tilde{D}}_k = \mathbb{E}_{Q(\bm{D})}[\ln \bm{D}_k] = \psi (\hat{d}_k) - \psi (\textstyle\sum_{i=1}^K \hat{d}_i)$.
\vspace{0.5cm}
\begin{mdframed}[style=proof]
	Using the definition of the Dirichlet distribution as given in Equation \eqref{eq:dirichlet_distribution_def}, the log-property, and the linearity of expectation, we have:
	\begin{align*}
		\mathbb{E}_{Q(\bm{D})}[\ln Q(\bm{D})] &= \mathbb{E}_{Q(\bm{D})}\Bigg[\ln \Big(\frac{1}{B(\hat{d})}\prod_{k=1}^K \bm{D}^{\hat{d}_k - 1}_k\Big)\Bigg] \\
		 &= \mathbb{E}_{Q(\bm{D})}\Bigg[-\ln B(\hat{d}) + \sum_{k=1}^K (\hat{d}_k - 1) \ln \bm{D}_k\Bigg] \\
		 &= -\ln B(\hat{d}) + \sum_{k=1}^K (\hat{d}_k - 1) \mathbb{E}_{Q(\bm{D})}[\ln \bm{D}_k].
	\end{align*}
	Then, using \eqref{eq:dirichlet_ln_expectation}, we get the final result:
	\begin{align*}
		\mathbb{E}_{Q(\bm{D})}[\ln Q(\bm{D})] &= -\ln B(\hat{d}) + \sum_{k=1}^K (\hat{d}_k - 1)\ln\bm{\tilde{D}}_k,
	\end{align*}
	where $\ln\bm{\tilde{D}}_k = \mathbb{E}_{Q(\bm{D})}[\ln \bm{D}_k] = \psi (\hat{d}_k) - \psi (\textstyle\sum_{i=1}^K \hat{d}_i)$.
\end{mdframed}

\newpage
\subsection*{B.2 Derivation for $\boldmath{\mathbb{E}_{Q(\bm{\mu}, \bm{\Lambda})}[\ln Q(\bm{\mu} | \bm{\Lambda})]}$}

Next, we prove that if $Q(\bm{\Lambda}) = \prod_{k=1}^K \mathcal{W}(\bm{\Lambda}_k; \bm{\hat{W}}_k, \hat{v}_k)$ and $Q(\bm{\mu} | \bm{\Lambda}) = \prod_{k=1}^K \mathcal{N}(\bm{\mu}_k; \bm{\hat{m}}_k, (\hat{\beta}_k\bm{\Lambda}_k)^{-1})$, then:
\begin{align*}
	\mathbb{E}_{Q(\bm{\mu}, \bm{\Lambda})}[\ln Q(\bm{\mu} | \bm{\Lambda})] &= \frac{1}{2} \sum_{k=1}^K K\ln \frac{\hat{\beta}_k}{2\pi} + \ln \bm{\tilde{\Lambda}}_k - K,
\end{align*}
where $\ln\bm{\tilde{\Lambda}}_k = \mathbb{E}_{Q(\bm{\Lambda_k})}[\ln|\bm{\Lambda}_k|] = K \ln 2 + \ln |\bm{\hat{W}}_k| + \sum_{i=1}^K \psi (\tfrac{\hat{v}_k+1-i}{2})$.
\vspace{0.5cm}
\begin{mdframed}[style=proof]
	Using the definition of the Gaussian distribution as given in Equation \eqref{eq:gaussian_distribution}, the log-property, and Equation \eqref{eq:determinant_of_inverse}:
	\begin{align*}
		\mathbb{E}_{Q(\bm{\mu}, \bm{\Lambda})}&[\ln Q(\bm{\mu} | \bm{\Lambda})] \\
		&= \mathbb{E}_{Q(\bm{\mu}, \bm{\Lambda})}\Bigg[\ln \prod_{k=1}^K \big(\tfrac{2\pi}{\hat{\beta}_k}\big)^{-\frac{K}{2}} |\bm{\Lambda}^{-1}_k|^{-\frac{1}{2}} \text{exp}\Big( -\frac{\hat{\beta}_k}{2} (\bm{\mu}_k - \bm{\hat{m}}_k)^\top \bm{\Lambda}_k (\bm{\mu}_k - \bm{\hat{m}}_k) \Big)\Bigg] \\
		&= \mathbb{E}_{Q(\bm{\mu}, \bm{\Lambda})}\Bigg[\sum_{k=1}^K \frac{K}{2}\ln \frac{\hat{\beta}_k}{2\pi} -\frac{1}{2}\ln |\bm{\Lambda}^{-1}_k| -\frac{\hat{\beta}_k}{2} (\bm{\mu}_k - \bm{\hat{m}}_k)^\top \bm{\Lambda}_k (\bm{\mu}_k - \bm{\hat{m}}_k) \Bigg] \\
		&= \mathbb{E}_{Q(\bm{\mu}, \bm{\Lambda})}\Bigg[\frac{1}{2} \sum_{k=1}^K K\ln \frac{\hat{\beta}_k}{2\pi} + \ln |\bm{\Lambda}_k| - \hat{\beta}_k (\bm{\mu}_k - \bm{\hat{m}}_k)^\top \bm{\Lambda}_k (\bm{\mu}_k - \bm{\hat{m}}_k) \Bigg].
	\end{align*}
	Then, using the linearity of expectation, as well as Equations \eqref{eq:wishart_ln_expectation} and \eqref{eq:wishart_gaussian_expectation_of_quadratic_form}:
	\begin{align*}
		\mathbb{E}_{Q(\bm{\mu}, \bm{\Lambda})}&[\ln Q(\bm{\mu} | \bm{\Lambda})] \\
		&= \frac{1}{2} \sum_{k=1}^K \Big(K\ln \frac{\hat{\beta}_k}{2\pi} + \mathbb{E}_{Q(\bm{\Lambda})}[\ln |\bm{\Lambda}_k|] - \hat{\beta}_k \mathbb{E}_{Q(\bm{\mu}, \bm{\Lambda})}[(\bm{\mu}_k - \bm{\hat{m}}_k)^\top \bm{\Lambda}_k (\bm{\mu}_k - \bm{\hat{m}}_k)]\Big) \\
		&= \frac{1}{2} \sum_{k=1}^K \Big(K\ln \frac{\hat{\beta}_k}{2\pi} + \ln \bm{\tilde{\Lambda}} - \hat{\beta}_k \big[K \hat{\beta}_k^{-1} + \hat{v}_k(\bm{\hat{m}}_k - \bm{\hat{m}}_k)^\top\bm{\hat{W}}_k(\bm{\hat{m}}_k - \bm{\hat{m}}_k)\big] \Big) \\
		&= \frac{1}{2} \sum_{k=1}^K K\ln \frac{\hat{\beta}_k}{2\pi} + \ln \bm{\tilde{\Lambda}} - K,
	\end{align*}
	where according to \eqref{eq:wishart_ln_expectation}, we have $\ln \bm{\tilde{\Lambda}} = \mathbb{E}_{Q(\bm{\Lambda}_k)}[\ln|\bm{\Lambda}|] = K \ln 2 + \ln |\bm{\hat{W}}_k| + \sum_{i=1}^K \psi (\tfrac{\hat{v}_k+1-i}{2})$.
\end{mdframed}

\newpage
\subsection*{B.3 Derivation for $\boldmath{\mathbb{E}_{Q(\bm{\Lambda})}[\ln Q(\bm{\Lambda})]}$}

Let us now demonstrate that if $Q(\bm{\Lambda}) = \prod_{k=1}^K \mathcal{W}(\bm{\Lambda}_k; \bm{\hat{W}}_k, \hat{v}_k)$, then:
\begin{align*}
	\mathbb{E}_{Q(\bm{\Lambda})}[\ln Q(\bm{\Lambda})] &= \sum_{k=1}^K - \frac{\hat{v}_kK}{2}\ln 2 - \frac{\hat{v}_k}{2}\ln|\bm{\hat{W}}_k| - \ln \Gamma_K(\frac{\hat{v}_k}{2}) + \frac{\hat{v}_k - K - 1}{2} \ln \bm{\tilde{\Lambda}}_k - \frac{\hat{v}_kK}{2},
\end{align*}
where $\ln\bm{\tilde{\Lambda}}_k = \mathbb{E}_{Q(\bm{\Lambda_k})}[\ln|\bm{\Lambda}_k|] = K \ln 2 + \ln |\bm{\hat{W}}_k| + \sum_{i=1}^K \psi (\tfrac{\hat{v}_k+1-i}{2})$.
\vspace{0.5cm}
\begin{mdframed}[style=proof]
	Using the definition of the Wishart distribution as given in Equation \eqref{eq:wishart_distribution_def}, the log-property, and the linearity of expectation:
	\begin{align*}
		\mathbb{E}_{Q(\bm{\Lambda})}&[\ln Q(\bm{\Lambda})] \\
		&= \mathbb{E}_{Q(\bm{\Lambda})}\Bigg[\ln \prod_{k=1}^K \mathcal{W}(\bm{\Lambda}_k; \bm{\hat{W}}_k, \hat{v}_k)\Bigg] \\
		&= \mathbb{E}_{Q(\bm{\Lambda})}\Bigg[\ln \prod_{k=1}^K 2^{-\frac{\hat{v}_kK}{2}} |\bm{\hat{W}}_k|^{-\frac{\hat{v}_k}{2}} \Gamma_K(\tfrac{\hat{v}_k}{2})^{-1} |\bm{\Lambda}_k|^{\frac{\hat{v}_k-K-1}{2}} \text{exp}\Big( -\frac{1}{2} \text{tr}(\bm{\hat{W}}^{-1}_k\bm{\Lambda}_k) \Big)\Bigg] \\
		&= \mathbb{E}_{Q(\bm{\Lambda})}\Bigg[\sum_{k=1}^K -\frac{\hat{v}_kK}{2} \ln 2 -\frac{\hat{v}_k}{2} \ln |\bm{\hat{W}}_k| - \ln \Gamma_K(\tfrac{\hat{v}_k}{2}) + \frac{\hat{v}_k-K-1}{2} \ln |\bm{\Lambda}_k| -\frac{1}{2} \text{tr}(\bm{\hat{W}}^{-1}_k\bm{\Lambda}_k) \Bigg] \\
		&= \sum_{k=1}^K -\frac{\hat{v}_kK}{2} \ln 2 -\frac{\hat{v}_k}{2} \ln |\bm{\hat{W}}_k| - \ln \Gamma_K(\tfrac{\hat{v}_k}{2}) + \frac{\hat{v}_k-K-1}{2} \mathbb{E}_{Q(\bm{\Lambda}_k)}[\ln |\bm{\Lambda}_k|] -\frac{1}{2} \mathbb{E}_{Q(\bm{\Lambda}_k)}[\text{tr}(\bm{\hat{W}}^{-1}_k\bm{\Lambda}_k)],
	\end{align*}
	where according to \eqref{eq:wishart_ln_expectation}, we have $\ln \bm{\tilde{\Lambda}} = \mathbb{E}_{Q(\bm{\Lambda}_k)}[\ln|\bm{\Lambda}|] = K \ln 2 + \ln |\bm{\hat{W}}_k| + \sum_{i=1}^K \psi (\tfrac{\hat{v}_k+1-i}{2})$. Then, focusing on the last term:
	\begin{align*}
		-\frac{1}{2} \mathbb{E}_{Q(\bm{\Lambda}_k)}[\text{tr}(\bm{\hat{W}}^{-1}_k\bm{\Lambda}_k)] &= -\frac{1}{2} \text{tr}(\bm{\hat{W}}^{-1}_k\mathbb{E}_{Q(\bm{\Lambda}_k)}[\bm{\Lambda}_k]) \\
		&= -\frac{1}{2} \text{tr}(\bm{\hat{W}}^{-1}_k \hat{v}_k\bm{\hat{W}}_k) \\
		&= -\frac{\hat{v}_k}{2} \text{tr}(\bm{\hat{W}}^{-1}_k\bm{\hat{W}}_k) \\
		&= -\frac{\hat{v}_kK}{2},
	\end{align*}
	which can be substituted back to get the final result:
	\begin{align*}
		\mathbb{E}_{Q(\bm{\Lambda})}[\ln Q(\bm{\Lambda})] = \sum_{k=1}^K -\frac{\hat{v}_kK}{2} \ln 2 -\frac{\hat{v}_k}{2} \ln |\bm{\hat{W}}_k| - \ln \Gamma_K(\tfrac{\hat{v}_k}{2}) + \frac{\hat{v}_k-K-1}{2} \ln \bm{\tilde{\Lambda}} -\frac{\hat{v}_kK}{2}.
	\end{align*}
\end{mdframed}

\newpage
\subsection*{B.4 Derivation for $\boldmath{\mathbb{E}_{Q(\bm{Z})}[\ln Q(\bm{Z})]}$}

In this section, we show that if $Q(\bm{Z}) = \prod_{n=1}^N \text{Cat}(\bm{z}_n;\bm{\hat{r}}_{n\bigcdot})$, then:
\begin{align*}
	\mathbb{E}_{Q(\bm{Z})}[\ln Q(\bm{Z})] &= \sum_{n=1}^N \sum_{k=1}^K \bm{\hat{r}}_{nk} \ln \bm{\hat{r}}_{nk}.
\end{align*}
\begin{mdframed}[style=proof]
	Using the definition of the Categorical distribution as given in Equation \eqref{eq:categorical_distribution_def}, the log-property, and the linearity of expectation:
	\vspace{-0.25cm}
	\begin{align*}
		\mathbb{E}_{Q(\bm{Z})}[\ln Q(\bm{Z})] &= \mathbb{E}_{Q(\bm{Z})}[\ln \prod_{n=1}^N \text{Cat}(\bm{z}_n | \bm{\hat{r}}_{n\bigcdot})] \\
		&= \mathbb{E}_{Q(\bm{Z})}[\ln \prod_{n=1}^N \prod_{k=1}^K \bm{\hat{r}}_{nk}^{\bm{z}_{nk}}] \\
		&= \mathbb{E}_{Q(\bm{Z})}[\sum_{n=1}^N \sum_{k=1}^K \bm{z}_{nk} \ln \bm{\hat{r}}_{nk}] \\
		&= \sum_{n=1}^N \sum_{k=1}^K \mathbb{E}_{Q(\bm{Z})}[\bm{z}_{nk}] \ln \bm{\hat{r}}_{nk},
	\end{align*}
	where the final result is recovered by realizing that $\mathbb{E}_{Q(\bm{Z})}[\bm{z}_{nk}] = \bm{\hat{r}}_{nk}$.
\end{mdframed}

\newpage
\subsection*{B.5 Derivation for $\boldmath{\mathbb{E}_{Q(\bm{D})}[\ln P(\bm{D})]}$}

Next, we prove that if $Q(\bm{D}) = \text{Dir}(\bm{D};\hat{d})$ and $P(\bm{D}) = \text{Dir}(\bm{D};d)$, then:
\begin{align*}
	\mathbb{E}_{Q(\bm{D})}[\ln P(\bm{D})] &= - \ln B(d) + \sum_{k=1}^K (d_k - 1)\ln\bm{\tilde{D}}_k,
\end{align*}
where $\ln\bm{\tilde{D}}_k = \mathbb{E}_{Q(\bm{D})}[\ln \bm{D}_k] = \psi (\hat{d}_k) - \psi (\textstyle\sum_{i=1}^K \hat{d}_i)$.
\vspace{0.5cm}
\begin{mdframed}[style=proof]
	Using the definition of the Dirichlet distribution as given in Equation \eqref{eq:dirichlet_distribution_def}, the log-property and the linearity of expectation:
	\begin{align*}
		\mathbb{E}_{Q(\bm{D})}[\ln P(\bm{D})] &= \mathbb{E}_{Q(\bm{D})}\bigg[\ln \Big(\frac{1}{B(d)}\prod_{k=1}^K \bm{D}^{d_k - 1}_k\Big)\bigg] \\
		&= \mathbb{E}_{Q(\bm{D})}\bigg[-\ln B(d) + \sum_{k=1}^K (d_k - 1) \ln \bm{D}_k\bigg] \\
		&= -\ln B(d) + \sum_{k=1}^K (d_k - 1) \mathbb{E}_{Q(\bm{D})}[\ln \bm{D}_k].
	\end{align*}
	Then, using \eqref{eq:dirichlet_ln_expectation}, we get the final result:
	\begin{align*}
		\mathbb{E}_{Q(\bm{D})}[\ln Q(\bm{D})] &= -\ln B(d) + \sum_{k=1}^K (d_k - 1)\ln\bm{\tilde{D}}_k,
	\end{align*}
	where $\ln\bm{\tilde{D}}_k = \mathbb{E}_{Q(\bm{D})}[\ln \bm{D}_k] = \psi (\hat{d}_k) - \psi (\textstyle\sum_{i=1}^K \hat{d}_i)$.
\end{mdframed}

\newpage
\subsection*{B.6 Derivation for $\boldmath{\mathbb{E}_{Q(\bm{\mu}, \bm{\Lambda})}[\ln P(\bm{\mu} | \bm{\Lambda})]}$}

Next, we demonstrate that if $Q(\bm{\Lambda}) = \prod_{k=1}^K \mathcal{W}(\bm{\Lambda}_k; \bm{\hat{W}}_k, \hat{v}_k)$, $Q(\bm{\mu} | \bm{\Lambda}) = \prod_{k=1}^K \mathcal{N}(\bm{\mu}_k; \bm{\hat{m}}_k, (\hat{\beta}_k\bm{\Lambda}_k)^{-1})$, and $P(\bm{\mu} | \bm{\Lambda}) = \prod_{k=1}^K \mathcal{N}(\bm{\mu}_k; \bm{m}_k, (\beta_k\bm{\Lambda}_k)^{-1})$, then:
\begin{align*}
	\mathbb{E}_{Q(\bm{\mu}, \bm{\Lambda})}[\ln P(\bm{\mu} | \bm{\Lambda})] &= \frac{1}{2} \sum_{k=1}^K K\ln \frac{\beta_k}{2\pi} + \ln \bm{\tilde{\Lambda}}_k - \frac{\beta_k K}{\hat{\beta}_k} - \beta_k\hat{v}_k (\bm{m}_k - \hat{\bm{m}}_k)^\top \hat{\bm{W}}_k (\bm{m}_k - \hat{\bm{m}}_k),
\end{align*}
where $\ln\bm{\tilde{\Lambda}}_k = \mathbb{E}_{Q(\bm{\Lambda_k})}[\ln|\bm{\Lambda}_k|] = K \ln 2 + \ln |\bm{\hat{W}}_k| + \sum_{i=1}^K \psi (\tfrac{\hat{v}_k+1-i}{2})$.
\vspace{0.5cm}
\begin{mdframed}[style=proof]
	Using the definition of the Gaussian distribution as given in Equation \eqref{eq:wishart_distribution_def}, the log-property and Equation \eqref{eq:determinant_of_inverse}:
	\begin{align*}
		\mathbb{E}_{Q(\bm{\mu}, \bm{\Lambda})}&[\ln P(\bm{\mu} | \bm{\Lambda})] \\
		&= \mathbb{E}_{Q(\bm{\mu}, \bm{\Lambda})}\Bigg[\ln \prod_{k=1}^K \big(\tfrac{2\pi}{\beta_k}\big)^{-\frac{K}{2}} |\bm{\Lambda}^{-1}_k|^{-\frac{1}{2}} \text{exp}\Big( -\frac{\beta_k}{2} (\bm{\mu}_k - \bm{m}_k)^\top \bm{\Lambda}_k (\bm{\mu}_k - \bm{m}_k) \Big)\Bigg] \\
		&= \mathbb{E}_{Q(\bm{\mu}, \bm{\Lambda})}\Bigg[\sum_{k=1}^K \frac{K}{2}\ln \frac{\beta_k}{2\pi} -\frac{1}{2}\ln |\bm{\Lambda}^{-1}_k| -\frac{\beta_k}{2} (\bm{\mu}_k - \bm{m}_k)^\top \bm{\Lambda}_k (\bm{\mu}_k - \bm{m}_k) \Bigg] \\
		&= \mathbb{E}_{Q(\bm{\mu}, \bm{\Lambda})}\Bigg[\frac{1}{2} \sum_{k=1}^K K\ln \frac{\beta_k}{2\pi} + \ln |\bm{\Lambda}_k| - \beta_k (\bm{\mu}_k - \bm{m}_k)^\top \bm{\Lambda}_k (\bm{\mu}_k - \bm{m}_k) \Bigg].
	\end{align*}
	Then, using the linearity of expectation, as well as Equations \eqref{eq:wishart_ln_expectation} and \eqref{eq:wishart_gaussian_expectation_of_quadratic_form}:
	\begin{align*}
		\mathbb{E}_{Q(\bm{\mu}, \bm{\Lambda})}&[\ln P(\bm{\mu} | \bm{\Lambda})] \\
		&= \frac{1}{2} \sum_{k=1}^K K\ln \frac{\beta_k}{2\pi} + \mathbb{E}_{Q(\bm{\Lambda})}[\ln |\bm{\Lambda}_k|] - \beta_k \mathbb{E}_{Q(\bm{\mu}, \bm{\Lambda})}[(\bm{\mu}_k - \bm{m}_k)^\top \bm{\Lambda}_k (\bm{\mu}_k - \bm{m}_k)] \\
		&= \frac{1}{2} \sum_{k=1}^K K\ln \frac{\beta_k}{2\pi} + \ln \bm{\tilde{\Lambda}} - \beta_k \big[K \hat{\beta}_k^{-1} + \hat{v}_k(\bm{m}_k - \bm{\hat{m}}_k)^\top\bm{\hat{W}}_k(\bm{m}_k - \bm{\hat{m}}_k)\big] \\
		&= \frac{1}{2} \sum_{k=1}^K K\ln \frac{\beta_k}{2\pi} + \ln \bm{\tilde{\Lambda}} - \frac{\beta_kK}{\hat{\beta}_k} - \beta_k\hat{v}_k(\bm{m}_k - \bm{\hat{m}}_k)^\top\bm{\hat{W}}_k(\bm{m}_k - \bm{\hat{m}}_k),
	\end{align*}
	where according to \eqref{eq:wishart_ln_expectation}, we have $\ln \bm{\tilde{\Lambda}} = \mathbb{E}_{Q(\bm{\Lambda}_k)}[\ln|\bm{\Lambda}|] = K \ln 2 + \ln |\bm{\hat{W}}_k| + \sum_{i=1}^K \psi (\tfrac{\hat{v}_k+1-i}{2})$.
\end{mdframed}

\newpage
\subsection*{B.7 Derivation for $\boldmath{\mathbb{E}_{Q(\bm{\Lambda})}[\ln P(\bm{\Lambda})]}$}

Let us now demonstrate that if $Q(\bm{\Lambda}) = \prod_{k=1}^K \mathcal{W}(\bm{\Lambda}_k; \bm{\hat{W}}_k, \hat{v}_k)$ and $P(\bm{\Lambda}) = \prod_{k=1}^K \mathcal{W}(\bm{\Lambda}_k; \bm{W}_k, v_k)$, then:
\begin{align*}
	\mathbb{E}_{Q(\bm{\Lambda})}[\ln P(\bm{\Lambda})] = \sum_{k=1}^K -\frac{v_kK}{2} \ln 2 -\frac{v_k}{2} \ln |\bm{W}_k| - \ln \Gamma_K(\tfrac{v_k}{2}) + \frac{(v_k-K-1)}{2} \ln \bm{\tilde{\Lambda}} -\frac{\hat{v}_k}{2} \text{tr}(\bm{W}^{-1}_k\bm{\hat{W}}_k),
\end{align*}
where $\ln\bm{\tilde{\Lambda}}_k = \mathbb{E}_{Q(\bm{\Lambda_k})}[\ln|\bm{\Lambda}_k|] = K \ln 2 + \ln |\bm{\hat{W}}_k| + \sum_{i=1}^K \psi (\tfrac{\hat{v}_k+1-i}{2})$.
\vspace{0.5cm}
\begin{mdframed}[style=proof]
	Using the definition of the Wishart distribution as given in \eqref{eq:wishart_distribution_def}, and using the log-property as well as the linearity of expectation:
	\begin{align*}
		\mathbb{E}_{Q(\bm{\Lambda})}&[\ln P(\bm{\Lambda})] \\
		&= \mathbb{E}_{Q(\bm{\Lambda})}\Bigg[\ln \prod_{k=1}^K \mathcal{W}(\bm{\Lambda}_k; \bm{W}_k, v_k)\Bigg] \\
		&= \mathbb{E}_{Q(\bm{\Lambda})}\Bigg[\ln \prod_{k=1}^K 2^{-\frac{v_kK}{2}} |\bm{W}_k|^{-\frac{v_k}{2}} \Gamma_K(\tfrac{v_k}{2})^{-1} |\bm{\Lambda}_k|^{\frac{v_k-K-1}{2}} \text{exp}\Big( -\frac{1}{2} \text{tr}(\bm{W}^{-1}_k\bm{\Lambda}_k) \Big)\Bigg] \\
		&= \mathbb{E}_{Q(\bm{\Lambda})}\Bigg[\sum_{k=1}^K -\frac{v_kK}{2} \ln 2 -\frac{v_k}{2} \ln |\bm{W}_k| - \ln \Gamma_K(\tfrac{v_k}{2}) + \frac{v_k-K-1}{2} \ln |\bm{\Lambda}_k| -\frac{1}{2} \text{tr}(\bm{W}^{-1}_k\bm{\Lambda}_k) \Bigg] \\
		&= \sum_{k=1}^K -\frac{v_kK}{2} \ln 2 -\frac{v_k}{2} \ln |\bm{W}_k| - \ln \Gamma_K(\tfrac{v_k}{2}) + \frac{v_k-K-1}{2} \mathbb{E}_{Q(\bm{\Lambda}_k)}[\ln |\bm{\Lambda}_k|] -\frac{1}{2} \mathbb{E}_{Q(\bm{\Lambda}_k)}[\text{tr}(\bm{W}^{-1}_k\bm{\Lambda}_k)].
	\end{align*}
	where according to \eqref{eq:wishart_ln_expectation}, we have $\ln \bm{\tilde{\Lambda}} = \mathbb{E}_{Q(\bm{\Lambda}_k)}[\ln|\bm{\Lambda}|] = K \ln 2 + \ln |\bm{\hat{W}}_k| + \sum_{i=1}^K \psi (\tfrac{\hat{v}_k+1-i}{2})$. Then, focusing on the last term:
	\begin{align*}
		-\frac{1}{2} \mathbb{E}_{Q(\bm{\Lambda}_k)}[\text{tr}(\bm{W}^{-1}_k\bm{\Lambda}_k)] &= -\frac{1}{2} \text{tr}(\bm{W}^{-1}_k\mathbb{E}_{Q(\bm{\Lambda}_k)}[\bm{\Lambda}_k]) \\
		&= -\frac{1}{2} \text{tr}(\bm{W}^{-1}_k \hat{v}_k\bm{\hat{W}}_k) \\
		&= -\frac{\hat{v}_k}{2} \text{tr}(\bm{W}^{-1}_k\bm{\hat{W}}_k),
	\end{align*}
	which can be substituted back to get the final result:
	\begin{align*}
		\mathbb{E}_{Q(\bm{\Lambda})}[\ln P(\bm{\Lambda})] = \sum_{k=1}^K -\frac{v_kK}{2} \ln 2 -\frac{v_k}{2} \ln |\bm{W}_k| - \ln \Gamma_K(\tfrac{v_k}{2}) + \frac{(v_k-K-1)}{2} \ln \bm{\tilde{\Lambda}} -\frac{\hat{v}_k}{2} \text{tr}(\bm{W}^{-1}_k\bm{\hat{W}}_k).
	\end{align*}
\end{mdframed}

\newpage
\subsection*{B.8 Derivation for $\boldmath{\mathbb{E}_{Q(\bm{Z}, \bm{D})}[\ln P(\bm{Z}|\bm{D})]}$}

Let us now prove that if $Q(\bm{Z}) = \prod_{n=1}^N \text{Cat}(\bm{z}_n;\bm{\hat{r}}_{n\bigcdot})$, $P(\bm{Z}|\bm{D}) = \prod_{n=1}^N \text{Cat}(\bm{z}_n | \bm{D})$, and $Q(\bm{D}) = \text{Dir}(\bm{D};\hat{d})$, then:
\vspace{-0.5cm}
\begin{align*}
	\mathbb{E}_{Q(\bm{Z}, \bm{D})}[\ln P(\bm{Z}|\bm{D})] &= \sum_{k=1}^K N_{k} \ln\bm{\tilde{D}}_k,
\end{align*}
where $N_{k} = \sum_{n=1}^N \bm{\hat{r}}_{nk}$ and $\ln\bm{\tilde{D}}_k = \mathbb{E}_{Q(\bm{D})}[\ln \bm{D}_k] = \psi (\hat{d}_k) - \psi (\textstyle\sum_{i=1}^K \hat{d}_i)$.\
\vspace{0.5cm}
\begin{mdframed}[style=proof]
	Using the definition of the Categorical distribution as given in \eqref{eq:categorical_distribution_def}, and using the log-property as well as the linearity of expectation:
	\vspace{-0.25cm}
	\begin{align*}
		\mathbb{E}_{Q(\bm{Z}, \bm{D})}[\ln P(\bm{Z}|\bm{D})] &= \mathbb{E}_{Q(\bm{Z}, \bm{D})}[\ln \prod_{n=1}^N \text{Cat}(\bm{z}_n | \bm{D})] \\
		&= \mathbb{E}_{Q(\bm{Z}, \bm{D})}[\ln \prod_{n=1}^N \prod_{k=1}^K \bm{D}_{k}^{\bm{z}_{nk}}] \\
		&= \mathbb{E}_{Q(\bm{Z}, \bm{D})}[\sum_{n=1}^N \sum_{k=1}^K \bm{z}_{nk} \ln \bm{D}_{k}] \\
		&= \sum_{n=1}^N \sum_{k=1}^K \mathbb{E}_{Q(\bm{Z})}[\bm{z}_{nk}] \mathbb{E}_{Q(\bm{D})}[\ln \bm{D}_{k}].
	\end{align*}
	Then, realizing that $\mathbb{E}_{Q(\bm{Z})}[\bm{z}_{nk}] = \bm{\hat{r}}_{nk}$ and recalling that $\ln\bm{\tilde{D}}_k = \mathbb{E}_{Q(\bm{D})}[\ln \bm{D}_{k}]$:
	\begin{align*}
		\mathbb{E}_{Q(\bm{Z}, \bm{D})}[\ln P(\bm{Z}|\bm{D})] &= \sum_{n=1}^N \sum_{k=1}^K \bm{\hat{r}}_{nk} \ln\bm{\tilde{D}}_k \\
		&= \sum_{k=1}^K \ln\bm{\tilde{D}}_k \sum_{n=1}^N \bm{\hat{r}}_{nk} \\
		&= \sum_{k=1}^K N_{k} \ln\bm{\tilde{D}}_k,
	\end{align*}
	where we defined $N_{k} = \sum_{n=1}^N \bm{\hat{r}}_{nk}$.
\end{mdframed}

\newpage
\subsection*{B.9 Derivation for $\boldmath{\mathbb{E}_{Q(\bm{Z}, \bm{\mu}, \bm{\Lambda})}[\ln P(\bm{X} | \bm{Z}, \bm{\mu}, \bm{\Lambda})]}$}

Finally, we demonstrate that if $Q(\bm{Z}) = \prod_{n=1}^N \text{Cat}(\bm{z}_n;\bm{\hat{r}}_{n\bigcdot})$, $Q(\bm{\Lambda}) = \prod_{k=1}^K \mathcal{W}(\bm{\Lambda}_k; \bm{\hat{W}}_k, \hat{v}_k)$, $P(\bm{X} | \bm{Z}, \bm{\mu}, \bm{\Lambda}) = \prod_{n=1}^N \prod_{k=1}^K \mathcal{N}(\bm{x}_n; \bm{\mu}_k, \bm{\Lambda}_k^{-1})^{\bm{z}_{nk}}$ and $Q(\bm{\mu} | \bm{\Lambda}) = \prod_{k=1}^K \mathcal{N}(\bm{\mu}_k; \bm{\hat{m}}_k, (\hat{\beta}_k\bm{\Lambda}_k)^{-1})$, then:
\begin{align*}
	\mathbb{E}_{Q(\bm{Z}, \bm{\mu}, \bm{\Lambda})}[&\ln P(\bm{X} | \bm{Z}, \bm{\mu}, \bm{\Lambda})] = \\
	&\hspace{-1cm}\sum_{k=1}^K \frac{N_k}{2} \Bigg[\ln \bm{\tilde{\Lambda}}_k - K\hat{\beta}_k^{-1} -\hat{v}_k tr(\bm{S}_k\bm{\hat{W}}_k) - \hat{v}_k(\bm{\bar{x}}_k - \bm{\hat{m}}_k)^\top \bm{\hat{W}}_k (\bm{\bar{x}}_k - \bm{\hat{m}}_k) -K \ln 2\pi \Bigg].
\end{align*}
\begin{mdframed}[style=proof]
	Using the definition of the Gaussian distribution as given in \eqref{eq:gaussian_distribution}, the log-property, and the linearity of expectation:
	\vspace{-0.25cm}
	\begin{align*}
		\mathbb{E}_{Q(\bm{Z}, \bm{\mu}, \bm{\Lambda})}&[\ln P(\bm{X} | \bm{Z}, \bm{\mu}, \bm{\Lambda})] \\
		&= \mathbb{E}_{Q(\bm{Z}, \bm{\mu}, \bm{\Lambda})}\bigg[\ln \prod_{n=1}^N \prod_{k=1}^K \mathcal{N}(\bm{x}_n; \bm{\mu}_k, \bm{\Lambda}_k^{-1})^{\bm{z}_{nk}}\bigg] \\
		&= \mathbb{E}_{Q(\bm{Z}, \bm{\mu}, \bm{\Lambda})}\bigg[\sum_{n=1}^N \sum_{k=1}^K \bm{z}_{nk} \ln \mathcal{N}(\bm{x}_n; \bm{\mu}_k, \bm{\Lambda}_k^{-1})\bigg] \\
		&= \mathbb{E}_{Q(\bm{Z}, \bm{\mu}, \bm{\Lambda})}\bigg[\sum_{n=1}^N \sum_{k=1}^K \bm{z}_{nk} \ln \bigg((2\pi)^{-\frac{K}{2}} |\bm{\Lambda}_k^{-1}|^{-\frac{1}{2}} \text{exp}\Big( -\frac{1}{2} (\bm{x}_n - \bm{\mu}_k)^\top \bm{\Lambda}_k (\bm{x}_n - \bm{\mu}_k) \Big)\bigg)\bigg] \\
		&= \mathbb{E}_{Q(\bm{Z}, \bm{\mu}, \bm{\Lambda})}\bigg[\sum_{n=1}^N \sum_{k=1}^K \bm{z}_{nk} \Big( -\frac{K}{2}\ln 2\pi -\frac{1}{2} \ln |\bm{\Lambda}_k^{-1}| -\frac{1}{2} (\bm{x}_n - \bm{\mu}_k)^\top \bm{\Lambda}_k (\bm{x}_n - \bm{\mu}_k)\Big)\Bigg].
	\end{align*}
	Then, realizing that $\mathbb{E}_{Q(\bm{Z})}[\bm{z}_{nk}] = \bm{\hat{r}}_{nk}$, and using \eqref{eq:wishart_gaussian_expectation_of_quadratic_form} and \eqref{eq:wishart_ln_expectation}, we have:
	\begin{align}
		\mathbb{E}_{Q(\bm{Z}, \bm{\mu}, \bm{\Lambda})}&[\ln P(\bm{X} | \bm{Z}, \bm{\mu}, \bm{\Lambda})] \nonumber \\
		&= \sum_{n=1}^N \sum_{k=1}^K \mathbb{E}_{Q(\bm{Z})}[\bm{z}_{nk}] \Bigg[ -\frac{K}{2}\ln 2\pi - \frac{1}{2} \mathbb{E}_{Q(\bm{\Lambda})}[\ln  |\bm{\Lambda}_k^{-1}|] -\frac{1}{2} \mathbb{E}_{Q(\bm{\mu}, \bm{\Lambda})}[(\bm{x}_n - \bm{\mu}_k)^\top \bm{\Lambda}_k (\bm{x}_n - \bm{\mu}_k)]\Bigg] \nonumber \\
		&= \sum_{n=1}^N \sum_{k=1}^K \frac{\bm{\hat{r}}_{nk}}{2} \Bigg[ -K\ln 2\pi + \mathbb{E}_{Q(\bm{\Lambda})}[\ln  |\bm{\Lambda}_k|] - K \hat{\beta}_k^{-1} - \hat{v}_k(\bm{x}_n - \bm{\hat{m}}_k)^\top\bm{\hat{W}}_k(\bm{x}_n - \bm{\hat{m}}_k) \Bigg] \nonumber \\
		&= \sum_{n=1}^N \sum_{k=1}^K \frac{\bm{\hat{r}}_{nk}}{2} \Bigg[ -K\ln 2\pi + \ln \bm{\tilde{\Lambda}} - K \hat{\beta}_k^{-1} - \hat{v}_k(\bm{x}_n - \bm{\hat{m}}_k)^\top\bm{\hat{W}}_k(\bm{x}_n - \bm{\hat{m}}_k) \Bigg] \nonumber \\
		&= \sum_{k=1}^K \frac{N_{k}}{2} \Bigg[ -K\ln 2\pi + \ln \bm{\tilde{\Lambda}} - K \hat{\beta}_k^{-1} \Bigg] - \sum_{k=1}^K \underbrace{\sum_{n=1}^N  \frac{\bm{\hat{r}}_{nk}}{2} \Bigg[ \hat{v}_k(\bm{x}_n - \bm{\hat{m}}_k)^\top\bm{\hat{W}}_k(\bm{x}_n - \bm{\hat{m}}_k) \Bigg]}_{Y_{1k}}, \label{eq:Y_1k_def}
	\end{align}
	where we used $\ln  |\bm{\Lambda}_k^{-1}| = -\ln |\bm{\Lambda}_k|$, and defined $\ln \bm{\tilde{\Lambda}} = \mathbb{E}_{Q(\bm{\Lambda}_k)}[\ln|\bm{\Lambda}|] = K \ln 2 + \ln |\bm{\hat{W}}_k| + \sum_{i=1}^K \psi (\tfrac{\hat{v}_k+1-i}{2})$ as well as $N_{k} = \sum_{n=1}^N \bm{\hat{r}}_{nk}$. Next, we focus on the second term in the above equation, for all $k$, we have:
	\begin{align*}
		Y_{1k} &= \sum_{n=1}^N \frac{\bm{\hat{r}}_{nk}}{2} \hat{v}_k \Bigg[ \bm{x}_n^\top\bm{\hat{W}}_k\bm{x}_n - \bm{x}_n^\top\bm{\hat{W}}_k\bm{\hat{m}}_k - \bm{\hat{m}}_k^\top\bm{\hat{W}}_k\bm{x}_n + \bm{\hat{m}}_k^\top\bm{\hat{W}}_k\bm{\hat{m}}_k \Bigg] \\
		&= \frac{\hat{v}_k}{2} \Bigg[ \sum_{n=1}^N \bm{\hat{r}}_{nk} \bm{x}_n^\top\bm{\hat{W}}_k\bm{x}_n - \sum_{n=1}^N \bm{\hat{r}}_{nk}\bm{x}_n^\top\bm{\hat{W}}_k\bm{\hat{m}}_k - \sum_{n=1}^N \bm{\hat{r}}_{nk}\bm{\hat{m}}_k^\top\bm{\hat{W}}_k\bm{x}_n + \sum_{n=1}^N \bm{\hat{r}}_{nk}\bm{\hat{m}}_k^\top\bm{\hat{W}}_k\bm{\hat{m}}_k \Bigg] \\
		&= \frac{\hat{v}_k}{2} \Bigg[ \sum_{n=1}^N \bm{\hat{r}}_{nk} \bm{x}_n^\top\bm{\hat{W}}_k\bm{x}_n - N_{k}\bm{\bar{x}}_k^\top\bm{\hat{W}}_k\bm{\hat{m}}_k - N_{k}\bm{\hat{m}}_k^\top\bm{\hat{W}}_k\bm{\bar{x}}_k + N_{k}\bm{\hat{m}}_k^\top\bm{\hat{W}}_k\bm{\hat{m}}_k \Bigg],
	\end{align*}
	where $\bm{\bar{x}}_k = \frac{1}{N_k}\sum_{n=1}^N \bm{\hat{r}}_{nk}\bm{x}_n$. By adding and subtracting $N_{k}\bm{\bar{x}}_k^\top\bm{\hat{W}}_k\bm{\bar{x}}_k$, we get:
	\begin{align*}
		Y_{1k} &= \frac{\hat{v}_k}{2} \sum_{n=1}^N \Big[ \bm{\hat{r}}_{nk} \bm{x}_n^\top\bm{\hat{W}}_k\bm{x}_n - N_{k}\bm{\bar{x}}_k^\top\bm{\hat{W}}_k\bm{\bar{x}}_k + N_{k}\bm{\bar{x}}_k^\top\bm{\hat{W}}_k\bm{\bar{x}}_k - N_{k}\bm{\bar{x}}_k^\top\bm{\hat{W}}_k\bm{\hat{m}}_k - N_{k}\bm{\hat{m}}_k^\top\bm{\hat{W}}_k\bm{\bar{x}}_k + N_{k}\bm{\hat{m}}_k^\top\bm{\hat{W}}_k\bm{\hat{m}}_k \Big] \\
		&= \frac{\hat{v}_k}{2} \Bigg[ \sum_{n=1}^N \bm{\hat{r}}_{nk} \bm{x}_n^\top\bm{\hat{W}}_k\bm{x}_n - N_{k}\bm{\bar{x}}_k^\top\bm{\hat{W}}_k\bm{\bar{x}}_k + N_{k}(\bm{\bar{x}}_k - \bm{\hat{m}}_k)^\top\bm{\hat{W}}_k(\bm{\bar{x}}_k - \bm{\hat{m}}_k)\Bigg] \\
		&= \frac{\hat{v}_k}{2} \underbrace{\sum_{n=1}^N \bm{\hat{r}}_{nk} (\bm{x}_n^\top\bm{\hat{W}}_k\bm{x}_n - \bm{\bar{x}}_k^\top\bm{\hat{W}}_k\bm{\bar{x}}_k)}_{Y_2} + \frac{N_{k}\hat{v}_k}{2} (\bm{\bar{x}}_k - \bm{\hat{m}}_k)^\top\bm{\hat{W}}_k(\bm{\bar{x}}_k - \bm{\hat{m}}_k).
	\end{align*}
	Focusing on the first term, we used \eqref{eq:trace_and_quadratic_form} and \eqref{eq:trace_addition} to get:
	\begin{align*}
		Y_2 &= \sum_{n=1}^N \bm{\hat{r}}_{nk} tr(\bm{x}_n\bm{x}_n^\top\bm{\hat{W}}_k - \bm{\bar{x}}_k\bm{\bar{x}}_k^\top\bm{\hat{W}}_k),
	\end{align*}
	we then add and subtract $\bm{\bar{x}}_k\bm{\bar{x}}_k^\top\bm{\hat{W}}_k - \bm{x}_n\bm{\bar{x}}_k^\top\bm{\hat{W}}_k - \bm{\bar{x}}_k\bm{x}_n^\top\bm{\hat{W}}_k$, and use \eqref{eq:outter_product_neg} to obtain:
	\begin{align*}
		Y_2 &= \sum_{n=1}^N \bm{\hat{r}}_{nk} tr((\bm{x}_n - \bm{\bar{x}}_k)(\bm{x}_n - \bm{\bar{x}}_k)^\top\bm{\hat{W}}_k - \bm{\bar{x}}_k\bm{\bar{x}}_k^\top\bm{\hat{W}}_k + \bm{x}_n\bm{\bar{x}}_k^\top\bm{\hat{W}}_k + \bm{\bar{x}}_k\bm{x}_n^\top\bm{\hat{W}}_k - \bm{\bar{x}}_k\bm{\bar{x}}_k^\top\bm{\hat{W}}_k).
	\end{align*}
	Moreover, by using \eqref{eq:trace_addition}, we have:
	\begin{align*}
		Y_2 &= \sum_{n=1}^N \bm{\hat{r}}_{nk} tr((\bm{x}_n - \bm{\bar{x}}_k)(\bm{x}_n - \bm{\bar{x}}_k)^\top\bm{\hat{W}}_k) + \sum_{n=1}^N \bm{\hat{r}}_{nk} tr(\bm{x}_n\bm{\bar{x}}_k^\top\bm{\hat{W}}_k + \bm{\bar{x}}_k\bm{x}_n^\top\bm{\hat{W}}_k - 2\bm{\bar{x}}_k\bm{\bar{x}}_k^\top\bm{\hat{W}}_k) \\
		&= N_k tr(\bm{S}_k\bm{\hat{W}}_k) + \underbrace{\sum_{n=1}^N \bm{\hat{r}}_{nk} tr(\bm{x}_n\bm{\bar{x}}_k^\top\bm{\hat{W}}_k + \bm{\bar{x}}_k\bm{x}_n^\top\bm{\hat{W}}_k - 2\bm{\bar{x}}_k\bm{\bar{x}}_k^\top\bm{\hat{W}}_k)}_{Y_3} \\
		&= N_k tr(\bm{S}_k\bm{\hat{W}}_k).
	\end{align*}
	where $\bm{S}_k = \frac{1}{N_k}\sum_{n=1}^N \bm{\hat{r}}_{nk}(\bm{x}_n - \bm{\bar{x}}_k)(\bm{x}_n - \bm{\bar{x}}_k)^\top$ and $Y_3 = 0$ because:
	\begin{align*}
		Y_3 &= \sum_{n=1}^N \bm{\hat{r}}_{nk} tr(\bm{x}_n\bm{\bar{x}}_k^\top\bm{\hat{W}}_k + \bm{\bar{x}}_k\bm{x}_n^\top\bm{\hat{W}}_k - 2\bm{\bar{x}}_k\bm{\bar{x}}_k^\top\bm{\hat{W}}_k) \\
		&= tr(\sum_{n=1}^N \bm{\hat{r}}_{nk} \bm{x}_n\bm{\bar{x}}_k^\top\bm{\hat{W}}_k + \sum_{n=1}^N \bm{\hat{r}}_{nk} \bm{\bar{x}}_k\bm{x}_n^\top\bm{\hat{W}}_k - \sum_{n=1}^N \bm{\hat{r}}_{nk} 2\bm{\bar{x}}_k\bm{\bar{x}}_k^\top\bm{\hat{W}}_k) \\
		&= tr(N_{k}\bm{\bar{x}}_k\bm{\bar{x}}_k^\top\bm{\hat{W}}_k + N_{k}\bm{\bar{x}}_k\bm{\bar{x}}_k^\top\bm{\hat{W}}_k - N_{k} 2\bm{\bar{x}}_k\bm{\bar{x}}_k^\top\bm{\hat{W}}_k) = 0.
	\end{align*}
	By substituting $Y_2$ into $Y_{1k}$, and $Y_{1k}$ into \eqref{eq:Y_1k_def}, we get the final result:
	\begin{align*}
		\mathbb{E}_{Q(\bm{Z}, \bm{\mu}, \bm{\Lambda})}&[\ln P(\bm{X} | \bm{Z}, \bm{\mu}, \bm{\Lambda})] \\ 
		&\hspace{-1.5cm}= \sum_{k=1}^K \frac{N_{k}}{2} \Bigg[ -K\ln 2\pi + \ln \bm{\tilde{\Lambda}} - K \hat{\beta}_k^{-1} \Bigg] - \sum_{k=1}^K \bigg[\frac{\hat{v}_k}{2} N_k tr(\bm{S}_k\bm{\hat{W}}_k) + \frac{N_{k}\hat{v}_k}{2} (\bm{\bar{x}}_k - \bm{\hat{m}}_k)^\top\bm{\hat{W}}_k(\bm{\bar{x}}_k - \bm{\hat{m}}_k)\bigg] \\
		&\hspace{-1.5cm}= \sum_{k=1}^K \frac{N_{k}}{2} \Bigg[ -K\ln 2\pi + \ln \bm{\tilde{\Lambda}} - K \hat{\beta}_k^{-1} - \hat{v}_k tr(\bm{S}_k\bm{\hat{W}}_k) - \hat{v}_k (\bm{\bar{x}}_k - \bm{\hat{m}}_k)^\top\bm{\hat{W}}_k(\bm{\bar{x}}_k - \bm{\hat{m}}_k) \Bigg]
	\end{align*}
\end{mdframed}

\newpage
\section*{Appendix C: Perception model, update equation of variational distribution}

In this appendix, we start with the general update equation of variational inference, i.e., Equation \eqref{eq:VI_general_solution}, and derive the specific update equations for $\bm{Z}$, $\bm{D}$, $\bm{\mu}$, and $\bm{\Lambda}$. Importantly, the update equations for $\bm{Z}$, $\bm{D}$, $\bm{\mu}$, and $\bm{\Lambda}$ rely on the generative model and variational distribution presented in Section \ref{sec:perception}.

\subsection*{C.1 Update equation for $\boldmath{Z}$}

Applying the general update equation given in \eqref{eq:VI_general_solution} to the random variable $\bm{Z}$, we get:
\begin{align*}
	\ln Q^*(\bm{Z}) = \mathbb{E}_{Q(\bm{D})}[\ln P(\bm{Z} | \bm{D})] + \mathbb{E}_{Q(\bm{\mu}, \bm{\Lambda})}[\ln P(\bm{X} | \bm{Z}, \bm{\mu}, \bm{\Lambda})] + c,
\end{align*}
where $c$ is a constant w.r.t. $\bm{Z}$. From the above equation, we can show that:
\begin{align}
	\ln Q^*(\bm{Z}) = \sum_{n=1}^N \sum_{k=1}^K \bm{z}_{nk} \ln \bm{\rho}_{nk} + c \label{eq:pre_solution}
\end{align}
where:
\begin{align*}
	\ln \bm{\rho}_{nk} &= \mathbb{E}_{Q(\bm{D})}[\ln \bm{D}_{k}] - \frac{K}{2} \ln 2 \pi + \frac{1}{2} \mathbb{E}_{Q(\bm{\Lambda}_k)}[\ln |\bm{\Lambda}_k|] - \frac{1}{2} \mathbb{E}_{Q(\bm{\mu}_k, \bm{\Lambda}_k)}[(\bm{x}_n - \bm{\mu}_k)^\top \bm{\Lambda}_k(\bm{x}_n - \bm{\mu}_k)].
\end{align*}
Importantly, the expectations in the above equation can be evaluated using \eqref{eq:wishart_ln_expectation}, \eqref{eq:wishart_gaussian_expectation_of_quadratic_form}, and \eqref{eq:dirichlet_ln_expectation}.
\vspace{0.5cm}
\begin{mdframed}[style=proof]
	Using \eqref{eq:VI_general_solution}, we get the update equation for $\bm{Z}$:
	\begin{align*}
		\ln Q^*(\bm{Z}) = \mathbb{E}_{Q(\bm{D})}[\ln P(\bm{Z} | \bm{D})] + \mathbb{E}_{Q(\bm{\mu}, \bm{\Lambda})}[\ln P(\bm{X} | \bm{Z}, \bm{\mu}, \bm{\Lambda})] + c.
	\end{align*}
	Focusing on the first term, and using the log-property, the linearity of expectation, and the fact that $P(\bm{Z} | \bm{D})$ is a product of categorical distributions, we get:
	\begin{align*}
		\mathbb{E}_{Q(\bm{D})}[\ln P(\bm{Z} | \bm{D})] &=  \mathbb{E}_{Q(\bm{D})}\bigg[\sum_{n=1}^N \sum_{k=1}^K \bm{z}_{nk} \ln \bm{D}_{k}\bigg] = \sum_{n=1}^N \sum_{k=1}^K \bm{z}_{nk} \mathbb{E}_{Q(\bm{D})}[\ln \bm{D}_{k}],
	\end{align*}
	where $\mathbb{E}_{Q(\bm{D})}[\ln \bm{D}_{k}]$ can be evaluated using \eqref{eq:dirichlet_ln_expectation}. Finally, using the log-property, the linearity of expectation and the definition of the Gaussian distribution, the second term can be re-arranged as follows:
	\begin{align}
		\mathbb{E}_{Q(\bm{\mu}, \bm{\Lambda})}[&\ln P(\bm{X} | \bm{Z}, \bm{\mu}, \bm{\Lambda})] \nonumber \\
		&= \mathbb{E}_{Q(\bm{\mu}_k, \bm{\Lambda}_k)}\bigg[\sum_{n=1}^N \sum_{k=1}^K \bm{z}_{nk} \ln \mathcal{N}(\bm{x}_n | \bm{\mu}_k, \bm{\Lambda}_k)\bigg] \nonumber \\
		&= \sum_{n=1}^N \sum_{k=1}^K \bm{z}_{nk} \mathbb{E}_{Q(\bm{\mu}_k, \bm{\Lambda}_k)}[- \frac{K}{2} \ln 2 \pi - \frac{1}{2} \ln |\bm{\Lambda}_k^{-1}| - \frac{1}{2} (\bm{x}_n - \bm{\mu}_k)^\top \bm{\Lambda}_k(\bm{x}_n - \bm{\mu}_k)] \nonumber \\
		&= \sum_{n=1}^N \sum_{k=1}^K \bm{z}_{nk} [- \frac{K}{2} \ln 2 \pi + \frac{1}{2} \mathbb{E}_{Q(\bm{\Lambda}_k)}[\ln |\bm{\Lambda}_k|] - \frac{1}{2} \mathbb{E}_{Q(\bm{\mu}_k, \bm{\Lambda}_k)}[(\bm{x}_n - \bm{\mu}_k)^\top \bm{\Lambda}_k(\bm{x}_n - \bm{\mu}_k)]], \label{eq:expectation_of_ln_gaussian}
	\end{align}
	where the expectations can be evaluated using \eqref{eq:wishart_ln_expectation} and \eqref{eq:wishart_gaussian_expectation_of_quadratic_form}. Substituting the two results above back into the update equation for $\bm{Z}$ and re-arranging, we get:
	\begin{align*}
		\ln Q^*(\bm{Z}) = \sum_{n=1}^N \sum_{k=1}^K \bm{z}_{nk} \ln \bm{\rho}_{nk} + c,
	\end{align*}
	where:
	\begin{align*}
		\ln \bm{\rho}_{nk} &= \mathbb{E}_{Q(\bm{D})}[\ln \bm{D}_{k}] - \frac{K}{2} \ln 2 \pi + \frac{1}{2} \mathbb{E}_{Q(\bm{\Lambda}_k)}[\ln |\bm{\Lambda}_k|] - \frac{1}{2} \mathbb{E}_{Q(\bm{\mu}_k, \bm{\Lambda}_k)}[(\bm{x}_n - \bm{\mu}_k)^\top \bm{\Lambda}_k(\bm{x}_n - \bm{\mu}_k)].
	\end{align*}
\end{mdframed}
\vspace{0.5cm}
\noindent Taking the exponential of both sides in \eqref{eq:pre_solution}, we obtain:
\begin{align*}
	Q^*(\bm{Z}) \propto \prod_{n=1}^N \prod_{k=1}^K  \bm{\rho}_{nk}^{\bm{z}_{nk}}.
\end{align*}
Note, for every value of $n$, $\bm{z}_{nk}$ are binary variables that sum to one, and $Q^*(\bm{Z})$ must be normalized, thus:
\begin{align*}
	Q^*(\bm{Z}) = \prod_{n=1}^N \prod_{k=1}^K  \bm{\hat{r}}_{nk}^{\bm{z}_{nk}} \quad \text{where:} \quad \bm{\hat{r}}_{nk} = \frac{\bm{\rho}_{nk}}{\sum_{k=1}^K \bm{\rho}_{nk}}.
\end{align*}
Importantly, we recognize that $Q^*(\bm{Z})$ is a product of categorical distributions parameterized by $\bm{\hat{r}}_{nk}$:
\begin{align*}
Q^*(\bm{Z}) = \prod_{n=1}^N \text{Cat}(\bm{z}_n ; \bm{\hat{r}}_{n\,\bigcdot}).
\end{align*}

\subsection*{C.2 Update equation for $\boldmath{D}$}

Using \eqref{eq:VI_general_solution}, we can obtain the following update equation for $\bm{D}$:
\begin{align*}
	\ln Q^*(\bm{D}) = \ln P(\bm{D}) + \mathbb{E}_{Q(\bm{Z})}[\ln P(\bm{Z} | \bm{D})] + c,
\end{align*}
from which it is possible to show that:
\begin{align*}
	\ln Q^*(\bm{D}) &= \ln \prod_{k=1}^K \bm{D}_k^{(d_{k} - 1 + \sum_{n=1}^N \bm{\hat{r}}_{nk})} + c,
\end{align*}
where $c$ is a constant w.r.t. $\bm{D}$.
\vspace{0.5cm}
\begin{mdframed}[style=proof]
	Using \eqref{eq:VI_general_solution}, we get:
	\begin{align*}
		\ln Q^*(\bm{D}) = \ln P(\bm{D}) + \mathbb{E}_{Q(\bm{Z})}[\ln P(\bm{Z} | \bm{D})] + c.
	\end{align*}
	Using the definition of the Dirichlet and categorical distributions, and the log-property:
	\begin{align*}
		\ln Q^*(\bm{D}) = \sum_{k=1}^K (d_{k} - 1) \ln \bm{D}_k + \mathbb{E}_{Q(\bm{Z})}\bigg[\sum_{n=1}^N \sum_{k=1}^K \bm{z}_{nk} \ln \bm{D}_k\bigg] + c.
	\end{align*}
	Using the linearity of the expectation, and re-arranging:
	\begin{align*}
		\ln Q^*(\bm{D}) &= \sum_{k=1}^K (d_{k} - 1) \ln \bm{D}_k + \sum_{n=1}^N \sum_{k=1}^K \mathbb{E}_{Q(\bm{Z})}[\bm{z}_{nk}] \ln \bm{D}_k + c \\
		&= \sum_{k=1}^K \Big( d_{k} - 1 + \sum_{n=1}^N \bm{\hat{r}}_{nk} \Big) \ln \bm{D}_k + c,
	\end{align*}
	where $\bm{\hat{r}}_{nk} = \mathbb{E}_{Q(\bm{Z})}[\bm{z}_{nk}]$. Finally, using the log-property, we get:
	\begin{align*}
		\ln Q^*(\bm{D}) &= \ln \prod_{k=1}^K \bm{D}_k^{(d_{k} - 1 + \sum_{n=1}^N \bm{\hat{r}}_{nk})} + c,
	\end{align*}
	which is a Dirichlet distribution with parameters: $\hat{d}_{k} = d_{k} + \sum_{n=1}^N \bm{\hat{r}}_{nk}$.
\end{mdframed}

\subsection*{C.3 Update equation for $\boldmath{\mu}$ and $\boldmath{\Lambda}$}

Using the general update equation given in \eqref{eq:VI_general_solution}, we get:
\begin{align*}
	\ln Q^*(\bm{\mu}, \bm{\Lambda}) = \ln P(\bm{\mu} | \bm{\Lambda}) + \ln P(\bm{\Lambda}) + \mathbb{E}_{Q(\bm{Z})}[\ln P(\bm{X} | \bm{Z}, \bm{\mu}, \bm{\Lambda})] + c,
\end{align*}
where $c$ is a constant w.r.t. $\bm{\mu}$ and $\bm{\Lambda}$. Recalling that $P(\bm{\mu} | \bm{\Lambda})$ is a product of Gaussian distributions, $P(\bm{\Lambda})$ is a product of Wishart distributions, and $P(\bm{X} | \bm{Z}, \bm{\mu}, \bm{\Lambda})$ is a mixture of Gaussian distributions, we can rearrange the above equation as follows:
\begin{align*}
	\ln Q^*(\bm{\mu}, \bm{\Lambda}) = \sum_{k=1}^K \Bigg[ \ln \mathcal{N}(\bm{\mu}_k; \bm{m}_k, (\beta_k\bm{\Lambda}_k)^{-1}) + \ln \mathcal{W}(\bm{\Lambda}_k; \bm{W}_k, v_k) + \sum_{n=1}^N \mathbb{E}_{Q(\bm{Z})}[\ln \mathcal{N}(\bm{x}_n; \bm{\mu}_k, \bm{\Lambda}_k^{-1})^{\bm{z}_{nk}}] \Bigg] + c,
\end{align*}
thus:
\begin{align*}
	\ln Q^*(\bm{\mu}_k, \bm{\Lambda}_k) = \ln \mathcal{N}(\bm{\mu}_k; \bm{m}_k, (\beta_k\bm{\Lambda}_k)^{-1}) + \ln \mathcal{W}(\bm{\Lambda}_k; \bm{W}_k, v_k) + \sum_{n=1}^N \mathbb{E}_{Q(\bm{Z})}[\bm{z}_{nk}]\ln \mathcal{N}(\bm{x}_n; \bm{\mu}_k, \bm{\Lambda}_k^{-1}) + c.
\end{align*}
Using that $\mathbb{E}_{Q(\bm{Z})}[\bm{z}_{nk}] = \bm{\hat{r}}_{nk}$, taking the exponential of both sides, and using the definition of Gaussian and Wishart distributions, as well as \eqref{eq:determinant_of_inverse}, we can rearrange as follows:
\begin{align}
	Q^*(\bm{\mu}_k, \bm{\Lambda}_k) \propto& \, \mathcal{N}(\bm{\mu}_k; \bm{m}_k, (\beta_k\bm{\Lambda}_k)^{-1}) \mathcal{W}(\bm{\Lambda}_k; \bm{W}_k, v_k)\prod_{n=1}^N \mathcal{N}(\bm{x}_n; \bm{\mu}_k, \bm{\Lambda}_k^{-1})^{\bm{\hat{r}}_{nk}} \nonumber \\
	\propto& \, |\bm{\Lambda}_k^{-1}|^{-\frac{1}{2}} |\bm{\Lambda}_k|^{\frac{(v_k + N_k) - D - 1}{2}}\exp\Big(-\frac{1}{2}Y_1\Big), \label{eq:posterior_eq_Y1}
\end{align}
where $N_k = \sum_{n=1}^N \bm{\hat{r}}_{nk}$, we identify $\hat{v}_k = v_k + N_k$ from $|\bm{\Lambda}_k|^{\frac{(v_k + N_k) - D - 1}{2}}$, and:
\begin{align*}
	Y_1 = tr(\bm{W}_k^{-1}\bm{\Lambda}_k) + (\bm{\mu}_k - \bm{m}_k)^\top \beta_k\bm{\Lambda}_k(\bm{\mu}_k - \bm{m}_k) + \sum_{n=1}^N \bm{\hat{r}}_{nk}(\bm{x}_n - \bm{\mu}_k)^\top\bm{\Lambda}_k(\bm{x}_n - \bm{\mu}_k).
\end{align*}
Using \eqref{eq:gaussian_form_expansion} to rearrange $Y_1$, we get:
\begin{align*}
	Y_1 = tr(\bm{W}_k^{-1}\bm{\Lambda}_k) + \bm{\mu}_k^\top \beta_k\bm{\Lambda}_k \bm{\mu}_k -2 \bm{\mu}_k^\top \beta_k\bm{\Lambda}_k \bm{m}_k + \bm{m}_k^\top \beta_k\bm{\Lambda}_k \bm{m}_k + \sum_{n=1}^N \bm{\hat{r}}_{nk}[\bm{x}_n^\top \bm{\Lambda}_k \bm{x}_n - 2 \bm{x}_n^\top \bm{\Lambda}_k \bm{\mu}_k + \bm{\mu}_k^\top\bm{\Lambda}_k\bm{\mu}_k].
\end{align*}
Using \eqref{eq:avg_quadratic_1_form}, we can simplify the second and third terms in the summation:
\begin{align*}
	Y_1 = tr(\bm{W}_k^{-1}\bm{\Lambda}_k) + \bm{\mu}_k^\top \beta_k\bm{\Lambda}_k \bm{\mu}_k -2 \bm{\mu}_k^\top \beta_k\bm{\Lambda}_k \bm{m}_k + \bm{m}_k^\top \beta_k\bm{\Lambda}_k \bm{m}_k + \sum_{n=1}^N [\bm{\hat{r}}_{nk}\bm{x}_n^\top \bm{\Lambda}_k \bm{x}_n] - 2 N_k \bm{\bar{x}}_k^\top \bm{\Lambda}_k \bm{\mu}_k + N_k\bm{\mu}_k^\top\bm{\Lambda}_k\bm{\mu}_k,
\end{align*}
where $\bm{\bar{x}}_k = \frac{1}{N_k}\sum_{n=1}^N \bm{\hat{r}}_{nk}\bm{x}_n$. Then, the second and last terms can be merged together. Similarly, using \eqref{eq:quadratic_form_symmetry} the third and before last terms can be combined, leading to:
\begin{align*}
	Y_1 =\bm{\mu}_k^\top (\beta_k + N_k) \bm{\Lambda}_k \bm{\mu}_k -2 \bm{\mu}_k^\top \bm{\Lambda}_k (\beta_k\bm{m}_k + N_k\bm{\bar{x}}_k) + tr(\bm{W}_k^{-1}\bm{\Lambda}_k) + \bm{m}_k^\top \beta_k\bm{\Lambda}_k \bm{m}_k + \sum_{n=1}^N \bm{\hat{r}}_{nk}\bm{x}_n^\top \bm{\Lambda}_k \bm{x}_n,
\end{align*}
where we identify $\hat{\beta}_k = \beta_k + N_k$ from $\bm{\mu}_k^\top (\beta_k + N_k) \bm{\Lambda}_k \bm{\mu}_k$. Note, we want $(\beta_k + N_k) \bm{\hat{m}}_k = \beta_k\bm{m}_k + N_k\bm{\bar{x}}_k$, for the quadratic form of a Gaussian distribution to emerge (see below), which implies that $\bm{\hat{m}}_k = \frac{\beta_k\bm{m}_k + N_k\bm{\bar{x}}_k}{\beta_k + N_k}$. Adding and subtracting $\bm{\hat{m}}_k^\top(\beta_k + N_k)\bm{\Lambda}_k \bm{\hat{m}}_k$:
\begin{align*}
	Y_1 = &\bm{\mu}_k^\top (\beta_k + N_k) \bm{\Lambda}_k \bm{\mu}_k -2 \bm{\mu}_k^\top (\beta_k + N_k) \bm{\Lambda}_k \bm{\hat{m}}_k + \bm{\hat{m}}_k^\top(\beta_k + N_k)\bm{\Lambda}_k \bm{\hat{m}}_k \\
	+ &tr(\bm{W}_k^{-1}\bm{\Lambda}_k) + \bm{m}_k^\top \beta_k\bm{\Lambda}_k \bm{m}_k - \bm{\hat{m}}_k^\top(\beta_k + N_k)\bm{\Lambda}_k \bm{\hat{m}}_k + \sum_{n=1}^N \bm{\hat{r}}_{nk} \bm{x}_n^\top \bm{\Lambda}_k \bm{x}_n.
\end{align*}
Using \eqref{eq:gaussian_form_expansion} backward, we obtain a Gaussian quadratic form, by combining the first three terms in the above equation:
\begin{align*}
	Y_1 = &(\bm{\mu}_k - \bm{\hat{m}}_k)^\top \hat{\beta}_k \bm{\Lambda}_k (\bm{\mu}_k - \bm{\hat{m}}_k) + tr(\bm{W}_k^{-1}\bm{\Lambda}_k) + \bm{m}_k^\top \beta_k\bm{\Lambda}_k \bm{m}_k - \bm{\hat{m}}_k^\top(\beta_k + N_k)\bm{\Lambda}_k \bm{\hat{m}}_k + \sum_{n=1}^N \bm{\hat{r}}_{nk} \bm{x}_n^\top \bm{\Lambda}_k \bm{x}_n.
\end{align*}
Substituting $Y_1$ back into the Equation \eqref{eq:posterior_eq_Y1}, and re-arranging:
\begin{align*}
	Q^*(\bm{\mu}_k, \bm{\Lambda}_k) \propto &|\bm{\Lambda}_k^{-1}|^{-\frac{1}{2}} \exp\Big(-\frac{1}{2}(\bm{\mu}_k - \bm{\hat{m}}_k)^\top \hat{\beta}_k \bm{\Lambda}_k (\bm{\mu}_k - \bm{\hat{m}}_k)\Big) \\
	&|\bm{\Lambda}_k|^{\frac{\hat{v}_k - D - 1}{2}}\exp\Big(-\frac{1}{2}[tr(\bm{W}_k^{-1}\bm{\Lambda}_k) + \bm{m}_k^\top \beta_k\bm{\Lambda}_k \bm{m}_k - \bm{\hat{m}}_k^\top(\beta_k + N_k)\bm{\Lambda}_k \bm{\hat{m}}_k + \sum_{n=1}^N \bm{\hat{r}}_{nk} \bm{x}_n^\top \bm{\Lambda}_k \bm{x}_n]\Big).
\end{align*}
Recognizing the definition of a Gaussian distribution:
\begin{align}
	Q^*(\bm{\mu}_k, \bm{\Lambda}_k) \propto & \mathcal{N}(\bm{\mu}_k|\bm{\hat{m}}_k, (\hat{\beta}_k\bm{\Lambda}_k)^{-1}) |\bm{\Lambda}_k|^{\frac{\hat{v}_k - D - 1}{2}}\exp\Big(-\frac{1}{2}Y_2\Big) \label{eq:posterior_Y2_MG}
\end{align}
where:
\begin{align*}
	Y_2 = tr(\bm{W}_k^{-1}\bm{\Lambda}_k) + \bm{m}_k^\top \beta_k\bm{\Lambda}_k \bm{m}_k - \bm{\hat{m}}_k^\top(\beta_k + N_k)\bm{\Lambda}_k \bm{\hat{m}}_k + \sum_{n=1}^N \bm{\hat{r}}_{nk} \bm{x}_n^\top \bm{\Lambda}_k \bm{x}_n.
\end{align*}
Using \eqref{eq:trace_and_quadratic_form}, we turn all the quadratic forms into expressions containing traces:
\begin{align*}
	Y_2 &= tr(\bm{W}_k^{-1}\bm{\Lambda}_k) + tr(\beta_k \bm{m}_k\bm{m}_k^\top \bm{\Lambda}_k) - tr((\beta_k + N_k)\bm{\hat{m}}_k\bm{\hat{m}}_k^\top \bm{\Lambda}_k) + \sum_{n=1}^N tr(\bm{\hat{r}}_{nk} \bm{x}_n \bm{x}_n^\top \bm{\Lambda}_k).
\end{align*}
Then, using \eqref{eq:trace_addition}, we merge all the traces as follows:
\begin{align}
	Y_2 &= tr([\bm{W}_k^{-1} + \beta_k \bm{m}_k\bm{m}_k^\top - (\beta_k + N_k)\bm{\hat{m}}_k\bm{\hat{m}}_k^\top + \sum_{n=1}^N \bm{\hat{r}}_{nk} \bm{x}_n \bm{x}_n^\top] \bm{\Lambda}_k), \label{eq:merged_traces_equation}
\end{align}
where we factored out $\bm{\Lambda}_k$. Adding and subtracting $\sum_{n = 1}^N \bm{\hat{r}}_{nk}[\bm{\bar{x}}_k\bm{\bar{x}}_k^\top - \bm{x}_n\bm{\bar{x}}_k^\top - \bm{\bar{x}}_k \bm{x}_n^\top]$, we get:
\begin{align}
	Y_2 &= tr([\bm{W}_k^{-1} + \sum_{n = 1}^N \bm{\hat{r}}_{nk}[\bm{x}_n \bm{x}_n^\top + \bm{\bar{x}}_k\bm{\bar{x}}_k^\top - \bm{x}_n\bm{\bar{x}}_k^\top - \bm{\bar{x}}_k \bm{x}_n^\top] + Y_3] \bm{\Lambda}_k), \label{eq:Y_2_before_prop_16}
\end{align}
where:
\begin{align}
	Y_3 &= \beta_k \bm{m}_k\bm{m}_k^\top - (\beta_k + N_k)\bm{\hat{m}}_k\bm{\hat{m}}_k^\top - \sum_{n = 1}^N \bm{\hat{r}}_{nk}[  \bm{\bar{x}}_k\bm{\bar{x}}_k^\top - \bm{x}_n\bm{\bar{x}}_k^\top - \bm{\bar{x}}_k \bm{x}_n^\top]. \label{eq:Y_2_before_prop_14_15}
\end{align}
Using \eqref{eq:outter_product_neg} backward in \eqref{eq:Y_2_before_prop_16}, we obtain:
\begin{align*}
	Y_2 &= tr([\bm{W}_k^{-1} + \sum_{n=1}^N \bm{\hat{r}}_{nk}(\bm{x}_n - \bm{\bar{x}}_k)(\bm{x}_n - \bm{\bar{x}}_k)^\top + Y_3] \bm{\Lambda}_k).
\end{align*}
Using \eqref{eq:avg_outter_product_1} and \eqref{eq:avg_outter_product_2}, we simply the summation in \eqref{eq:Y_2_before_prop_14_15}, leading to:
\begin{align*}
	Y_3 &= \beta_k \bm{m}_k\bm{m}_k^\top - \underbrace{(\beta_k + N_k)\bm{\hat{m}}_k\bm{\hat{m}}_k^\top}_{Y_4} + N_k\bm{\bar{x}}_k\bm{\bar{x}}_k^\top.
\end{align*}
Using the definition of $\bm{\hat{m}}_k$ and \eqref{eq:outter_product_with_constants}, we can re-write $Y_4$ as:
\begin{align*}
	Y_4 &= (\beta_k + N_k)\Big(\frac{\beta_k\bm{m}_k + N_k\bm{\bar{x}}_k}{\beta_k + N_k}\Big)\Big(\frac{\beta_k\bm{m}_k + N_k\bm{\bar{x}}_k}{\beta_k + N_k}\Big)^\top = \frac{1}{\beta_k + N_k}\Big(\beta_k\bm{m}_k + N_k\bm{\bar{x}}_k\Big)\Big(\beta_k\bm{m}_k + N_k\bm{\bar{x}}_k\Big)^\top.
\end{align*}
Using \eqref{eq:outter_product_pos}, the outer product can be substituted by a sum of for term, by re-arranging, we then get:
\begin{align*}
	Y_4 &= \frac{1}{\beta_k + N_k}\Big(\beta_k^2\bm{m}_k\bm{m}_k^\top + \beta_kN_k\bm{m}_k\bm{\bar{x}}_k^\top + \beta_kN_k\bm{\bar{x}}_k\bm{m}_k^\top + N_k^2\bm{\bar{x}}_k\bm{\bar{x}}_k^\top\Big) \\
	&= \frac{\beta_k^2}{\beta_k + N_k}\bm{m}_k\bm{m}_k^\top + \frac{\beta_kN_k}{\beta_k + N_k}\bm{m}_k\bm{\bar{x}}_k^\top + \frac{\beta_kN_k}{\beta_k + N_k}\bm{\bar{x}}_k\bm{m}_k^\top + \frac{N_k^2}{\beta_k + N_k}\bm{\bar{x}}_k\bm{\bar{x}}_k^\top.
\end{align*}
Substituting $Y_4$ in $Y_3$, and rearranging:
\begin{align*}
	Y_3 &= \beta_k \bm{m}_k\bm{m}_k^\top - \frac{\beta_k^2}{\beta_k + N_k}\bm{m}_k\bm{m}_k^\top - \frac{\beta_kN_k}{\beta_k + N_k}\bm{m}_k\bm{\bar{x}}_k^\top - \frac{\beta_kN_k}{\beta_k + N_k}\bm{\bar{x}}_k\bm{m}_k^\top - \frac{N_k^2}{\beta_k + N_k}\bm{\bar{x}}_k\bm{\bar{x}}_k^\top + N_k\bm{\bar{x}}_k\bm{\bar{x}}_k^\top \\
	&= \frac{\beta_k(\beta_k + N_k) - \beta_k^2}{\beta_k + N_k} \bm{m}_k\bm{m}_k^\top - \frac{\beta_kN_k}{\beta_k + N_k}\bm{m}_k\bm{\bar{x}}_k^\top - \frac{\beta_kN_k}{\beta_k + N_k}\bm{\bar{x}}_k\bm{m}_k^\top + \frac{N_k(\beta_k + N_k) - N_k^2}{\beta_k + N_k}\bm{\bar{x}}_k\bm{\bar{x}}_k^\top \\
	&= \frac{\beta_kN_k}{\beta_k + N_k} \bm{m}_k\bm{m}_k^\top - \frac{\beta_kN_k}{\beta_k + N_k}\bm{m}_k\bm{\bar{x}}_k^\top - \frac{\beta_kN_k}{\beta_k + N_k}\bm{\bar{x}}_k\bm{m}_k^\top + \frac{\beta_kN_k}{\beta_k + N_k}\bm{\bar{x}}_k\bm{\bar{x}}_k^\top \\
	&= \frac{\beta_kN_k}{\beta_k + N_k}[\bm{m}_k\bm{m}_k^\top - \bm{m}_k\bm{\bar{x}}_k^\top - \bm{\bar{x}}_k\bm{m}_k^\top + \bm{\bar{x}}_k\bm{\bar{x}}_k^\top].
\end{align*}
Using \eqref{eq:outter_product_neg} backward, we turn the four terms between the brackets into an outer product:
\begin{align*}
	Y_3 &=  \frac{\beta_kN_k}{\beta_k + N_k}(\bm{\bar{x}}_k - \bm{m}_k)(\bm{\bar{x}}_k - \bm{m}_k)^\top.
\end{align*}
Substituting $Y_3$ back into $Y_2$, and $Y_2$ back into \eqref{eq:posterior_Y2_MG}:
\begin{align*}
	Q^*(\bm{\mu}_k, \bm{\Lambda}_k) \propto & \mathcal{N}(\bm{\mu}_k|\bm{\hat{m}}_k, (\hat{\beta}_k\bm{\Lambda}_k)^{-1}) |\bm{\Lambda}_k|^{\frac{\hat{v}_k - D - 1}{2}}\exp\Big(-\frac{1}{2}tr(\bm{\hat{W}}_k^{-1}\bm{\Lambda}_k)\Big)
\end{align*}
where we identified:
\begin{align*}
	\bm{\hat{W}}_k^{-1} = \bm{W}_k^{-1} + \sum_{n=1}^N \bm{\hat{r}}_{nk}(\bm{x}_n - \bm{\bar{x}}_k)(\bm{x}_n - \bm{\bar{x}}_k)^\top + \frac{\beta_kN_k}{\beta_k + N_k}(\bm{\bar{x}}_k - \bm{m}_k)(\bm{\bar{x}}_k - \bm{m}_k)^\top.
\end{align*}
Recognizing the definition of a Wishart distribution, we obtain the final result:
\begin{align*}
	Q^*(\bm{\mu}_k, \bm{\Lambda}_k) \propto & \mathcal{N}(\bm{\mu}_k|\bm{\hat{m}}_k, (\hat{\beta}_k\bm{\Lambda}_k)^{-1})\mathcal{W}(\bm{\Lambda}_k; \bm{\hat{W}}_k, \hat{v}_k)
\end{align*}
where:
\begin{align*}
	\bm{\hat{W}}_k^{-1} &= \bm{W}_k^{-1} + \sum_{n=1}^N \bm{\hat{r}}_{nk}(\bm{x}_n - \bm{\bar{x}}_k)(\bm{x}_n - \bm{\bar{x}}_k)^\top + \frac{\beta_kN_k}{\beta_k + N_k}(\bm{\bar{x}}_k - \bm{m}_k)(\bm{\bar{x}}_k - \bm{m}_k)^\top, \\
	\bm{\hat{m}}_k &= \frac{\beta_k\bm{m}_k + N_k\bm{\bar{x}}_k}{\beta_k + N_k}, \\
	\hat{v}_k &= v_k + N_k, \\
	\hat{\beta}_k &= \beta_k + N_k.
\end{align*}

\newpage
\section*{Appendix D: Perception model, update equation of empirical prior}

In the previous section, we showed that the optimal variational posterior parameters can be written as:
\begin{align*}
	\hat{d}_{k} &= d_{k} + N_k, \\
	\bm{\hat{W}}_k^{-1} &= \bm{W}_k^{-1} + \sum_{n=1}^N \bm{\hat{r}}_{nk}(\bm{x}_n - \bm{\bar{x}}_k)(\bm{x}_n - \bm{\bar{x}}_k)^\top + \frac{\beta_kN_k}{\beta_k + N_k}(\bm{\bar{x}}_k - \bm{m}_k)(\bm{\bar{x}}_k - \bm{m}_k)^\top, \\
	\bm{\hat{m}}_k &= \frac{\beta_k\bm{m}_k + N_k\bm{\bar{x}}_k}{\beta_k + N_k}, \\
	\hat{v}_k &= v_k + N_k, \\
	\hat{\beta}_k &= \beta_k + N_k.
\end{align*}
We now derive update equations for the empirical prior's parameters, and a new set of update equations for the variational posterior that relies only on the empirical prior's parameters. More precisely, we show that:
\begin{align}
	\bar{v}_k &= v_k + N_k^{'},  & \hat{v}_k &= \bar{v}_k + N_k^{''}, \\
	\bar{\beta}_k &= \beta_k + N_k^{'}, & \hat{\beta}_k &= \bar{\beta}_k + N_k^{''}, \\
	\bar{d}_{k} &= d_{k} + N_k^{'}, & \hat{d}_{k} &= \bar{d}_{k} + N_k^{''},
\end{align}
where $N_k^{'}$ is the number of data points to forget attributed to the $k$-th component, and $N_k^{''}$ is the number of data points to keep attributed to the $k$-th component:
\begin{align}
	N_k^{'} &= \sum_{n \in N^{'}} \bm{\hat{r}}_{nk}, & N_k^{''} &= \sum_{n \in N^{''}} \bm{\hat{r}}_{nk}.
\end{align}
Additionally, we demonstrate that:
\begin{align}
	\bm{\bar{W}}_k^{-1} &= \bm{W}_k^{-1} + N_k^{'} \bm{S}_k^{'} + \frac{\beta_kN_k^{'}}{\beta_k + N_k^{'}}(\bm{\bar{x}}_k^{'} - \bm{m}_k)(\bm{\bar{x}}_k^{'} - \bm{m}_k)^\top, & \bm{\bar{m}}_k &= \frac{\beta_k\bm{m}_k + N_k^{'}\bm{\bar{x}}_k^{'}}{\beta_k + N_k^{'}}, \\
	\bm{\hat{W}}_k^{-1} &= \bm{\bar{W}}_k^{-1} + N_k^{''} \bm{S}_k^{''} + \frac{\bar{\beta}_k N_k^{''}}{\bar{\beta}_k + N_k^{''}}(\bm{\bar{x}}_k^{''} - \bm{\bar{m}}_k)(\bm{\bar{x}}_k^{''} - \bm{\bar{m}}_k)^\top, & \bm{\hat{m}}_k &= \frac{\bar{\beta}_k\bm{\bar{m}}_k + N_k^{''}\bm{\bar{x}}_k^{''}}{\bar{\beta}_k + N_k^{''}},
\end{align}
where the empirical mean $\bm{\bar{x}}_k^{'}$ and covariance $\bm{S}_k^{'}$ of the forgettable data points in the $k$-th, and the empirical mean $\bm{\bar{x}}_k^{''}$ and covariance $\bm{S}_k^{''}$ of the data points to keep in the $k$-th component are given by:
\begin{align}
	\bm{\bar{x}}_k^{'} &= \frac{1}{N_k^{'}}\sum_{n \in N^{'}} \bm{\hat{r}}_{nk}\bm{x}_n, & \bm{S}_k^{'} &= \frac{1}{N_k^{'}}\sum_{n \in N^{'}} \bm{\hat{r}}_{nk}(\bm{x}_n - \bm{\bar{x}}_k{'})(\bm{x}_n - \bm{\bar{x}}_k{'})^\top, \\
	\bm{\bar{x}}_k^{''} &= \frac{1}{N_k^{''}}\sum_{n \in N^{''}} \bm{\hat{r}}_{nk}\bm{x}_n, & \bm{S}_k^{''} &= \frac{1}{N_k^{''}}\sum_{n \in N^{''}} \bm{\hat{r}}_{nk}(\bm{x}_n - \bm{\bar{x}}_k^{''})(\bm{x}_n - \bm{\bar{x}}_k^{''})^\top.
\end{align}

\subsection*{D.1 Update equation for $\bar{v}_k$, $\hat{v}_k$, $\bar{\beta}_k$, $\hat{\beta}_k$, $\bar{d}_{k}$, and $\hat{d}_{k}$}

First, note that since $N^{'} \cup N^{''} = \{1, ..., N\}$, we have:
\begin{align}
	N_k = \sum_{n=1}^N \bm{\hat{r}}_{nk} = \sum_{n \in N^{'}} \bm{\hat{r}}_{nk} + \sum_{n \in N^{''}} \bm{\hat{r}}_{nk} = N_k^{'} + N_k^{''}. \label{eq:N_k_split_equation}
\end{align}
Remembering that:
\begin{align*}
	\hat{d}_{k} &= d_{k} + N_k \\
	\hat{v}_k &= v_k + N_k, \\
	\hat{\beta}_k &= \beta_k + N_k,
\end{align*}
and using \eqref{eq:N_k_split_equation}, we get:
\begin{align*}
	\hat{d}_{k} &= \underbrace{d_{k} + N_k^{'}}_{\bar{d}_{k}} + N_k^{''} \\
	\hat{v}_k &= \underbrace{v_k + N_k^{'}}_{\bar{v}_{k}} + N_k^{''}, \\
	\hat{\beta}_k &= \underbrace{\beta_k + N_k^{'}}_{\bar{\beta}_{k}} + N_k^{''},
\end{align*}
which leads to the expected result:
\begin{align}
	\bar{v}_k &= v_k + N_k^{'},  & \hat{v}_k &= \bar{v}_k + N_k^{''}, \\
	\bar{\beta}_k &= \beta_k + N_k^{'}, & \hat{\beta}_k &= \bar{\beta}_k + N_k^{''}, \\
	\bar{d}_{k} &= d_{k} + N_k^{'}, & \hat{d}_{k} &= \bar{d}_{k} + N_k^{''}.
\end{align}

\subsection*{D.2 Update equation for $\bm{\bar{m}}_k$, and $\bm{\hat{m}}_k$}

We first note that:
\begin{align}
	N_k \bm{\bar{x}}_k = N_k^{'} \bm{\bar{x}}_k^{'} + N_k^{''} \bm{\bar{x}}_k^{''}.
\end{align}

\begin{mdframed}[style=proof]
Indeed:
\begin{align}
	\bm{\bar{x}}_k &= \frac{1}{N_k}\sum_{n=1}^N \bm{\hat{r}}_{nk}\bm{x}_n,
\end{align}
thus, remembering that $N^{'} \cup N^{''} = \{1, ..., N\}$, we have:
\begin{align}
	N_k \bm{\bar{x}}_k &= \sum_{n=1}^N \bm{\hat{r}}_{nk}\bm{x}_n, \\
	&= \sum_{n \in N^{'}} \bm{\hat{r}}_{nk}\bm{x}_n + \sum_{n \in N^{''}} \bm{\hat{r}}_{nk}\bm{x}_n.
\end{align}
Then, dividing and multiplying the first and second term by $N_k^{'}$ and $N_k^{''}$, respectively, we get:
\begin{align}
	N_k \bm{\bar{x}}_k &= \frac{N_k^{'}}{N_k^{'}} \sum_{n \in N^{'}} \bm{\hat{r}}_{nk}\bm{x}_n + \frac{N_k^{''}}{N_k^{''}}\sum_{n \in N^{''}} \bm{\hat{r}}_{nk}\bm{x}_n \\
	&= N_k^{'} \bm{\bar{x}}_k^{'} + N_k^{''} \bm{\bar{x}}_k^{''}.
\end{align}
\end{mdframed}
Now, remember that the optimal $\bm{\hat{m}}_k$ is:
\begin{align}
	\bm{\hat{m}}_k = \frac{\beta_k\bm{m}_k + N_k\bm{\bar{x}}_k}{\beta_k + N_k},
\end{align}
and using $N_k \bm{\bar{x}}_k = N_k^{'} \bm{\bar{x}}_k^{'} + N_k^{''} \bm{\bar{x}}_k^{''}$ and $N_k = N_k^{'} + N_k^{''}$, we obtain:
\begin{align}
	\bm{\hat{m}}_k = \frac{\beta_k\bm{m}_k + N_k^{'} \bm{\bar{x}}_k^{'} + N_k^{''} \bm{\bar{x}}_k^{''}}{\beta_k + N_k^{'} + N_k^{''}},
\end{align}
where we recognize $\bar{\beta}_k = \beta_k + N_k^{'}$, and let $\bar{\beta}_k \bm{\bar{m}}_k = \beta_k\bm{m}_k + N_k^{'} \bm{\bar{x}}_k^{'}$. Thus, the above equation can be re-written as:
\begin{align}
	\bm{\hat{m}}_k = \frac{\bar{\beta}_k \bm{\bar{m}}_k + N_k^{''} \bm{\bar{x}}_k^{''}}{\bar{\beta}_k + N_k^{''}}, \label{eq:update_equation_for_m_hat}
\end{align}
moreover, since $\bar{\beta}_k \bm{\bar{m}}_k = \beta_k\bm{m}_k + N_k^{'} \bm{\bar{x}}_k^{'}$, we have:
\begin{align}
	\bm{\bar{m}}_k = \frac{\beta_k\bm{m}_k + N_k^{'} \bm{\bar{x}}_k^{'}}{\bar{\beta}_k} = \frac{\beta_k\bm{m}_k + N_k^{'} \bm{\bar{x}}_k^{'}}{\beta_k + N_k^{'}}.\label{eq:update_equation_for_m_bar}
\end{align}
Note, \eqref{eq:update_equation_for_m_hat} and \eqref{eq:update_equation_for_m_bar} are the update equations of $\bm{\hat{m}}_k$ and $\bm{\bar{m}}_k$, respectively.

\subsection*{D.3 Update equation for $\bm{\bar{W}}_k^{-1}$, and $\bm{\hat{W}}_k^{-1}$}

In this section, we show that the optimal updates for $\bm{\bar{W}}_k^{-1}$, and $\bm{\hat{W}}_k^{-1}$ are:
\begin{align}
	\bm{\bar{W}}_k^{-1} &= \bm{W}_k^{-1} + N_k^{'} \bm{S}_k^{'} + \frac{\beta_kN_k^{'}}{\beta_k + N_k^{'}}(\bm{\bar{x}}_k^{'} - \bm{m}_k)(\bm{\bar{x}}_k^{'} - \bm{m}_k)^\top, \\
	\bm{\hat{W}}_k^{-1} &= \bm{\bar{W}}_k^{-1} + N_k^{''} \bm{S}_k^{''} + \frac{\bar{\beta}_k N_k^{''}}{\bar{\beta}_k + N_k^{''}}(\bm{\bar{x}}_k^{''} - \bm{\bar{m}}_k)(\bm{\bar{x}}_k^{''} - \bm{\bar{m}}_k)^\top,
\end{align}
where:
\begin{align}
	\bm{S}_k^{'} &= \frac{1}{N_k^{'}}\sum_{n \in N^{'}} \bm{\hat{r}}_{nk}(\bm{x}_n - \bm{\bar{x}}_k{'})(\bm{x}_n - \bm{\bar{x}}_k{'})^\top, \\
	\bm{S}_k^{''} &= \frac{1}{N_k^{''}}\sum_{n \in N^{''}} \bm{\hat{r}}_{nk}(\bm{x}_n - \bm{\bar{x}}_k^{''})(\bm{x}_n - \bm{\bar{x}}_k^{''})^\top.
\end{align}

\begin{mdframed}[style=proof]
	From equation \eqref{eq:merged_traces_equation}, by remembering that $N^{'} \cup N^{''} = \{1, ..., N\}$:
	\begin{align}
		Y_2 = \text{tr}([\bm{W}^{-1}_k + \beta_k \bm{m}_k \bm{m}_k^\top - (\beta_k + N_k) \hat{\bm{m}}_k \hat{\bm{m}}_k^\top + \sum_{n \in N^{'}} \hat{\bm{r}}_{nk} \bm{x}_n \bm{x}_n^\top + \sum_{n \in N^{''}} \hat{\bm{r}}_{nk} \bm{x}_n \bm{x}_n^\top]\bm{\Lambda}_k).
	\end{align}
	Adding and subtracting $\sum_{n \in N^{'}} \hat{\bm{r}}_{nk} [\bar{\bm{x}}^{'}_k \bar{\bm{x}}^{' \top}_k - \bar{\bm{x}}^{'}_k \bm{x}_n^\top - \bm{x}_n \bar{\bm{x}}^{' \top}_k]$ and $\sum_{n \in N^{''}} \hat{\bm{r}}_{nk} [\bar{\bm{x}}^{''}_k \bar{\bm{x}}^{'' \top}_k - \bar{\bm{x}}^{''}_k \bm{x}_n^\top - \bm{x}_n \bar{\bm{x}}^{'' \top}_k]$:
	\begin{align}
		Y_2 = \text{tr}([& \nonumber \\
		&\bm{W}^{-1}_k + \beta_k \bm{m}_k \bm{m}_k^\top - (\beta_k + N_k) \hat{\bm{m}}_k \hat{\bm{m}}_k^\top \nonumber \\
		+ &\sum_{n \in N^{'}} \hat{\bm{r}}_{nk} [\bm{x}_n \bm{x}_n^\top + \bar{\bm{x}}^{'}_k \bar{\bm{x}}^{' \top}_k - \bar{\bm{x}}^{'}_k \bm{x}_n^\top - \bm{x}_n \bar{\bm{x}}^{' \top}_k] + \sum_{n \in N^{''}} \hat{\bm{r}}_{nk} [\bm{x}_n \bm{x}_n^\top + \bar{\bm{x}}^{''}_k \bar{\bm{x}}^{'' \top}_k - \bar{\bm{x}}^{''}_k \bm{x}_n^\top - \bm{x}_n \bar{\bm{x}}^{'' \top}_k] \nonumber \\ 
		- &\sum_{n \in N^{'}} \hat{\bm{r}}_{nk} [\bar{\bm{x}}^{'}_k \bar{\bm{x}}^{' \top}_k - \bar{\bm{x}}^{'}_k \bm{x}_n^\top - \bm{x}_n \bar{\bm{x}}^{' \top}_k] - \sum_{n \in N^{''}} \hat{\bm{r}}_{nk} [\bar{\bm{x}}^{''}_k \bar{\bm{x}}^{'' \top}_k - \bar{\bm{x}}^{''}_k \bm{x}_n^\top - \bm{x}_n \bar{\bm{x}}^{'' \top}_k] \nonumber \\
		]\bm{\Lambda}_k).\qquad \,\,\, &
	\end{align}
	Using \eqref{eq:outter_product_neg}:
	\begin{align}
	Y_2 = \text{tr}([& \nonumber \\
	&\bm{W}^{-1}_k + \beta_k \bm{m}_k \bm{m}_k^\top - (\beta_k + N_k) \hat{\bm{m}}_k \hat{\bm{m}}_k^\top \nonumber \\
	+ &\sum_{n \in N^{'}} \hat{\bm{r}}_{nk} (\bm{x}_n - \bar{\bm{x}}^{'}_k)(\bm{x}_n - \bar{\bm{x}}^{'}_k)^\top + \sum_{n \in N^{''}} \hat{\bm{r}}_{nk} (\bm{x}_n - \bar{\bm{x}}^{''}_k)(\bm{x}_n - \bar{\bm{x}}^{''}_k)^\top \nonumber \\ 
	- &\sum_{n \in N^{'}} \hat{\bm{r}}_{nk} [\bar{\bm{x}}^{'}_k \bar{\bm{x}}^{' \top}_k - \bar{\bm{x}}^{'}_k \bm{x}_n^\top - \bm{x}_n \bar{\bm{x}}^{' \top}_k] - \sum_{n \in N^{''}} \hat{\bm{r}}_{nk} [\bar{\bm{x}}^{''}_k \bar{\bm{x}}^{'' \top}_k - \bar{\bm{x}}^{''}_k \bm{x}_n^\top - \bm{x}_n \bar{\bm{x}}^{'' \top}_k] \nonumber \\
	]\bm{\Lambda}_k),\qquad \,\,\, &
	\end{align}
	and the definition of $S_k^{'}$ and $S_k^{''}$, we get:
	\begin{align}
		Y_2 = \text{tr}([\bm{W}^{-1}_k + N_k^{'}S_k^{'} + N_k^{''}S_k^{''} + Y_3]\bm{\Lambda}_k),
	\end{align}
	where:
	\begin{align}
	Y_3 = &\beta_k \bm{m}_k \bm{m}_k^\top - (\beta_k + N_k) \hat{\bm{m}}_k \hat{\bm{m}}_k^\top \nonumber \\
	&- \sum_{n \in N^{'}} \hat{\bm{r}}_{nk} [\bar{\bm{x}}^{'}_k \bar{\bm{x}}^{' \top}_k - \bar{\bm{x}}^{'}_k \bm{x}_n^\top - \bm{x}_n \bar{\bm{x}}^{' \top}_k] - \sum_{n \in N^{''}} \hat{\bm{r}}_{nk} [\bar{\bm{x}}^{''}_k \bar{\bm{x}}^{'' \top}_k - \bar{\bm{x}}^{''}_k \bm{x}_n^\top - \bm{x}_n \bar{\bm{x}}^{'' \top}_k].
	\end{align}
	Note that $\sum_{n \in N^{'}} \hat{\bm{r}}_{nk} [\bar{\bm{x}}^{'}_k \bar{\bm{x}}^{' \top}_k - \bar{\bm{x}}^{'}_k \bm{x}_n^\top - \bm{x}_n \bar{\bm{x}}^{' \top}_k] = - N^{'}_k\bar{\bm{x}}^{'}_k \bar{\bm{x}}^{' \top}_k$ and $\sum_{n \in N^{''}} \hat{\bm{r}}_{nk} [\bar{\bm{x}}^{''}_k \bar{\bm{x}}^{'' \top}_k - \bar{\bm{x}}^{''}_k \bm{x}_n^\top - \bm{x}_n \bar{\bm{x}}^{'' \top}_k] = - N^{''}_k\bar{\bm{x}}^{''}_k \bar{\bm{x}}^{'' \top}_k$, thus:
	\begin{align}
	Y_3 = &\beta_k \bm{m}_k \bm{m}_k^\top - \underbrace{(\beta_k + N_k) \hat{\bm{m}}_k \hat{\bm{m}}_k^\top}_{Y_4} + N^{'}_k\bar{\bm{x}}^{'}_k \bar{\bm{x}}^{' \top}_k + N^{''}_k\bar{\bm{x}}^{''}_k \bar{\bm{x}}^{'' \top}_k.
	\end{align}
	Focusing on $Y_4$, we have:
	\begin{align}
	Y_4 &= (\beta_k + N_k) \hat{\bm{m}}_k \hat{\bm{m}}_k^\top \\
	&= \frac{1}{\beta_k + N_k} (\beta_k \bm{m}_k + N_k \bar{\bm{x}}_k)(\beta_k \bm{m}_k + N_k \bar{\bm{x}}_k)^\top \\
	&= \frac{1}{\beta_k + N_k} (\beta_k \bm{m}_k + N^{'}_k \bar{\bm{x}}^{'}_k + N^{''}_k \bar{\bm{x}}^{''}_k)(\beta_k \bm{m}_k + N^{'}_k \bar{\bm{x}}^{'}_k + N^{''}_k \bar{\bm{x}}^{''}_k)^\top \\
	&= \frac{1}{\bar{\beta}_k + N^{''}_k} (\bar{\beta}_k \bar{\bm{m}}_k + N^{''}_k \bar{\bm{x}}^{''}_k)(\bar{\beta}_k \bar{\bm{m}}_k + N^{''}_k \bar{\bm{x}}^{''}_k)^\top \\
	&= \frac{\bar{\beta}_k^2}{\bar{\beta}_k + N^{''}_k} \bar{\bm{m}}_k\bar{\bm{m}}_k^\top + \frac{\bar{\beta}_k N^{''}_k}{\bar{\beta}_k + N^{''}_k} \bar{\bm{m}}_k \bar{\bm{x}}^{'' \top}_k + \frac{\bar{\beta}_k N^{''}_k}{\bar{\beta}_k + N^{''}_k} \bar{\bm{x}}^{''}_k \bar{\bm{m}}_k^\top + \frac{N^{'' 2}_k}{\bar{\beta}_k + N^{''}_k} \bar{\bm{x}}^{''}_k \bar{\bm{x}}^{'' \top}_k.
	\end{align}
	Adding and subtracting $\bar{\beta}_k \bar{\bm{m}}_k\bar{\bm{m}}_k^\top + N^{''}_k \bar{\bm{x}}^{''}_k \bar{\bm{x}}^{'' \top}_k$, we obtain:
	\begin{align}
	Y_4 &= \frac{\bar{\beta}_k^2}{\bar{\beta}_k + N^{''}_k} \bar{\bm{m}}_k\bar{\bm{m}}_k^\top + \frac{\bar{\beta}_k N^{''}_k}{\bar{\beta}_k + N^{''}_k} \bar{\bm{m}}_k \bar{\bm{x}}^{'' \top}_k + \frac{\bar{\beta}_k N^{''}_k}{\bar{\beta}_k + N^{''}_k} \bar{\bm{x}}^{''}_k \bar{\bm{m}}_k^\top + \frac{N^{'' 2}_k}{\bar{\beta}_k + N^{''}_k} \bar{\bm{x}}^{''}_k \bar{\bm{x}}^{'' \top}_k \nonumber \\
	&- \bar{\beta}_k \bar{\bm{m}}_k\bar{\bm{m}}_k^\top - N^{''}_k \bar{\bm{x}}^{''}_k \bar{\bm{x}}^{'' \top}_k + \bar{\beta}_k \bar{\bm{m}}_k\bar{\bm{m}}_k^\top + N^{''}_k \bar{\bm{x}}^{''}_k \bar{\bm{x}}^{'' \top}_k \\
	&= \frac{\bar{\beta}_k^2 - \bar{\beta}_k(\bar{\beta}_k + N^{''}_k)}{\bar{\beta}_k + N^{''}_k} \bar{\bm{m}}_k\bar{\bm{m}}_k^\top + \frac{\bar{\beta}_k N^{''}_k}{\bar{\beta}_k + N^{''}_k} \bar{\bm{m}}_k \bar{\bm{x}}^{'' \top}_k + \frac{\bar{\beta}_k N^{''}_k}{\bar{\beta}_k + N^{''}_k} \bar{\bm{x}}^{''}_k \bar{\bm{m}}_k^\top + \frac{N^{'' 2}_k - N^{''}_k(\bar{\beta}_k + N^{''}_k)}{\bar{\beta}_k + N^{''}_k} \bar{\bm{x}}^{''}_k \bar{\bm{x}}^{'' \top}_k \nonumber \\
	&+ \bar{\beta}_k \bar{\bm{m}}_k\bar{\bm{m}}_k^\top + N^{''}_k \bar{\bm{x}}^{''}_k \bar{\bm{x}}^{'' \top}_k \\
	&= \frac{\bar{\beta}_kN^{''}_k}{\bar{\beta}_k + N^{''}_k} [- \bar{\bm{m}}_k\bar{\bm{m}}_k^\top + \bar{\bm{m}}_k \bar{\bm{x}}^{'' \top}_k + \bar{\bm{x}}^{''}_k \bar{\bm{m}}_k^\top - \bar{\bm{x}}^{''}_k \bar{\bm{x}}^{'' \top}_k ] + \bar{\beta}_k \bar{\bm{m}}_k\bar{\bm{m}}_k^\top + N^{''}_k \bar{\bm{x}}^{''}_k \bar{\bm{x}}^{'' \top}_k.
	\end{align}
	Substituting $Y_4$ into $Y_3$, we get:
	\begin{align}
		Y_3 &= \beta_k \bm{m}_k \bm{m}_k^\top + \frac{\bar{\beta}_kN^{''}_k}{\bar{\beta}_k + N^{''}_k} [\bar{\bm{m}}_k\bar{\bm{m}}_k^\top - \bar{\bm{m}}_k \bar{\bm{x}}^{'' \top}_k - \bar{\bm{x}}^{''}_k \bar{\bm{m}}_k^\top + \bar{\bm{x}}^{''}_k \bar{\bm{x}}^{'' \top}_k ] - \bar{\beta}_k \bar{\bm{m}}_k\bar{\bm{m}}_k^\top + N^{'}_k\bar{\bm{x}}^{'}_k \bar{\bm{x}}^{' \top}_k \nonumber \\
		&- N^{''}_k \bar{\bm{x}}^{''}_k \bar{\bm{x}}^{'' \top}_k + N^{''}_k\bar{\bm{x}}^{''}_k \bar{\bm{x}}^{'' \top}_k \\
		&= \frac{\bar{\beta}_kN^{''}_k}{\bar{\beta}_k + N^{''}_k} (\bar{\bm{x}}^{''}_k - \bar{\bm{m}}_k)(\bar{\bm{x}}^{''}_k - \bar{\bm{m}}_k)^\top + \beta_k \bm{m}_k \bm{m}_k^\top - \underbrace{\bar{\beta}_k \bar{\bm{m}}_k\bar{\bm{m}}_k^\top}_{Y_5} + N^{'}_k\bar{\bm{x}}^{'}_k \bar{\bm{x}}^{' \top}_k.
	\end{align}
	Focusing on $Y_5$, and following a similar logic as for $Y_4$:
	\begin{align}
		Y_5 &= (\beta_k + N^{'}_k) \bar{\bm{m}}_k \bar{\bm{m}}_k^\top \\
		&= \frac{1}{\beta_k + N^{'}_k} (\beta_k \bm{m}_k + N^{'}_k \bar{\bm{x}}^{'}_k)(\beta_k \bm{m}_k + N^{'}_k \bar{\bm{x}}^{'}_k)^\top \\
		&= \frac{\beta_k^2}{\beta_k + N^{'}_k} \bm{m}_k\bm{m}_k^\top + \frac{\beta_k N^{'}_k}{\beta_k + N^{'}_k} \bm{m}_k \bar{\bm{x}}^{' \top}_k + \frac{\beta_k N^{'}_k}{\beta_k + N^{'}_k} \bar{\bm{x}}^{'}_k \bm{m}_k^\top + \frac{N^{' 2}_k}{\beta_k + N^{'}_k} \bar{\bm{x}}^{'}_k \bar{\bm{x}}^{' \top}_k.
	\end{align}
	Substituting $Y_5$ into $Y_3$, we get:
	\begin{align}
	Y_3 &= \beta_k \bm{m}_k \bm{m}_k^\top - \frac{\beta_k^2}{\beta_k + N^{'}_k} \bm{m}_k\bm{m}_k^\top - \frac{\beta_k N^{'}_k}{\beta_k + N^{'}_k} \bm{m}_k \bar{\bm{x}}^{' \top}_k - \frac{\beta_k N^{'}_k}{\beta_k + N^{'}_k} \bar{\bm{x}}^{'}_k \bm{m}_k^\top - \frac{N^{' 2}_k}{\beta_k + N^{'}_k} \bar{\bm{x}}^{'}_k \bar{\bm{x}}^{' \top}_k + N^{'}_k\bar{\bm{x}}^{'}_k \bar{\bm{x}}^{' \top}_k \nonumber \\
	&+ \frac{\bar{\beta}_kN^{''}_k}{\bar{\beta}_k + N^{''}_k} (\bar{\bm{x}}^{''}_k - \bar{\bm{m}}_k)(\bar{\bm{x}}^{''}_k - \bar{\bm{m}}_k)^\top \\
	&= \frac{\beta_k(\beta_k + N^{'}_k) - \beta_k^2}{\beta_k + N^{'}_k} \bm{m}_k\bm{m}_k^\top - \frac{\beta_k N^{'}_k}{\beta_k + N^{'}_k} \bm{m}_k \bar{\bm{x}}^{' \top}_k - \frac{\beta_k N^{'}_k}{\beta_k + N^{'}_k} \bar{\bm{x}}^{'}_k \bm{m}_k^\top + \frac{N^{'}_k(\beta_k + N^{'}_k) - N^{' 2}_k}{\beta_k + N^{'}_k} \bar{\bm{x}}^{'}_k \bar{\bm{x}}^{' \top}_k \nonumber \\
	&+ \frac{\bar{\beta}_kN^{''}_k}{\bar{\beta}_k + N^{''}_k} (\bar{\bm{x}}^{''}_k - \bar{\bm{m}}_k)(\bar{\bm{x}}^{''}_k - \bar{\bm{m}}_k)^\top \\
	&= \frac{\beta_kN^{'}_k}{\beta_k + N^{'}_k} [\bm{m}_k\bm{m}_k^\top - \bm{m}_k \bar{\bm{x}}^{' \top}_k - \bar{\bm{x}}^{'}_k \bm{m}_k^\top + \bar{\bm{x}}^{'}_k \bar{\bm{x}}^{' \top}_k] + \frac{\bar{\beta}_kN^{''}_k}{\bar{\beta}_k + N^{''}_k} (\bar{\bm{x}}^{''}_k - \bar{\bm{m}}_k)(\bar{\bm{x}}^{''}_k - \bar{\bm{m}}_k)^\top \\
	&= \frac{\beta_kN^{'}_k}{\beta_k + N^{'}_k} (\bar{\bm{x}}^{'}_k - \bm{m}_k)(\bar{\bm{x}}^{'}_k - \bm{m}_k)^\top + \frac{\bar{\beta}_kN^{''}_k}{\bar{\beta}_k + N^{''}_k} (\bar{\bm{x}}^{''}_k - \bar{\bm{m}}_k)(\bar{\bm{x}}^{''}_k - \bar{\bm{m}}_k)^\top.
	\end{align}
	Finally, substituting $Y_3$ into $Y_2$, we get:
	\begin{align}
	Y_2 = \text{tr}([& \nonumber \\
	&\underbrace{\underbrace{\bm{W}^{-1}_k + N_k^{'}S_k^{'} + \frac{\beta_kN^{'}_k}{\beta_k + N^{'}_k} (\bar{\bm{x}}^{'}_k - \bm{m}_k)(\bar{\bm{x}}^{'}_k - \bm{m}_k)^\top}_{\bar{\bm{W}}^{-1}_k} + N_k^{''}S_k^{''} + \frac{\bar{\beta}_kN^{''}_k}{\bar{\beta}_k + N^{''}_k} (\bar{\bm{x}}^{''}_k - \bar{\bm{m}}_k)(\bar{\bm{x}}^{''}_k - \bar{\bm{m}}_k)^\top}_{\hat{\bm{W}}^{-1}_k} \nonumber \\
	]\bm{\Lambda}_k).\qquad \,\,\, &
	\end{align}
\end{mdframed}

\newpage
\section*{Appendix E: Transition Model}

In this appendix, we start with the general update equation of variational inference, i.e., Equation \eqref{eq:VI_general_solution}, and derive the specific update equation for $\bm{B}$. Then, we re-arranged the resulting equation to obtain the update equation of the empirical prior. Importantly, the update equation for $\bm{B}$ depends on the generative model and variational distribution of Section \ref{sec:transition}.

\subsection*{E.1 Variational distribution, update equation for $\boldmath{B}$}

Once again, by using \eqref{eq:VI_general_solution}, we get the following update equation for $\bm{B}$:
\begin{align}
	\ln Q^*(\bm{B}) = \ln P(\bm{B}) + \mathbb{E}_{Q(\bm{Z}_1, \bm{Z}_0)}[\ln P(\bm{Z}_1 | \bm{Z}_0, \bm{A}_0, \bm{B})] + c,
\end{align}
from which we can show that:
\begin{align}
	\ln Q^*(\bm{B}) &= \ln \prod_{a=1}^A \prod_{j=1}^K \prod_{k=1}^K \bm{B}[a]_{kj}^{(b[a]_{kj} - 1 + \sum_{n=1}^M [a = a_n^0] \bm{\hat{r}}_{nk}^\tau \bm{\hat{r}}_{nk}^{\tau + 1})} + c,
\end{align}
where $c$ is a constant w.r.t. $\bm{B}$.
\vspace{0.5cm}
\begin{mdframed}[style=proof]
	Using \eqref{eq:VI_general_solution}, we get:
	\begin{align}
		\ln Q^*(\bm{B}) = \ln P(\bm{B}) + \mathbb{E}_{Q(\bm{Z}_1, \bm{Z}_0)}[\ln P(\bm{Z}_1 | \bm{Z}_0, \bm{A}_0, \bm{B})] + c.
	\end{align}
	Using the definition of the Dirichlet and categorical distributions, and the log-property:
	\begin{align}
		\ln Q^*(\bm{B}) = \sum_{a=1}^A \sum_{j=1}^K \sum_{k=1}^K (b[a]_{kj} - 1) \ln \bm{B}[a]_{kj} + \mathbb{E}_{Q(\bm{Z}_1, \bm{Z}_0)}[\sum_{n=1}^M \sum_{j=1}^K \sum_{k=1}^K \bm{z}_{nk}^0 \bm{z}_{nj}^1 \ln \bm{B}[a_n^0]_{kj}] + c.
	\end{align}
	Using the linearity of the expectation, and re-arranging:
	\begin{align}
		\ln Q^*(\bm{B}) &= \sum_{a=1}^A \sum_{j=1}^K \sum_{k=1}^K (b[a]_{kj} - 1) \ln \bm{B}[a]_{kj} + \sum_{n=1}^M \sum_{j=1}^K \sum_{k=1}^K \mathbb{E}_{Q(\bm{Z}_0)}[\bm{z}_{nk}^0] \mathbb{E}_{Q(\bm{Z}_1)}[\bm{z}_{nj}^1] \ln \bm{B}[a_n^0]_{kj} + c \\
		&= \sum_{a=1}^A \sum_{j=1}^K \sum_{k=1}^K (b[a]_{kj} - 1) \ln \bm{B}[a]_{kj} + \sum_{a=1}^A \sum_{j=1}^K \sum_{k=1}^K \sum_{n=1}^M [a = a_n^0] \bm{\hat{r}}_{nk}^0\bm{\hat{r}}_{nk}^1\ln \bm{B}[a_n^0]_{kj} + c \\
		&= \sum_{a=1}^A \sum_{j=1}^K \sum_{k=1}^K \Big(b[a]_{kj} - 1 + \sum_{n=1}^M [a = a_n^0] \bm{\hat{r}}_{nk}^0\bm{\hat{r}}_{nk}^1 \Big) \ln \bm{B}[a]_{kj} + c,
	\end{align}
	where $\bm{\hat{r}}_{nk}^0 = \mathbb{E}_{Q(\bm{Z}_0)}[\bm{z}_{nk}^0]$, $\bm{\hat{r}}_{nk}^1 = \mathbb{E}_{Q(\bm{Z}_1)}[\bm{z}_{nj}^1]$, and $[\,\bigcdot\,]$ is an indicator function equals to one if the condition is satisfied and zero otherwise. Finally, using the log-property:
	\begin{align}
		\ln Q^*(\bm{B}) &= \ln \prod_{a=1}^A \prod_{j=1}^K \prod_{k=1}^K \bm{B}[a]_{kj}^{(b[a]_{kj} - 1 + \sum_{n=1}^M [a = a_n^0] \bm{\hat{r}}_{nk}^0\bm{\hat{r}}_{nk}^1 )} + c,
	\end{align}
	which is a Dirichlet distribution with parameters: $\hat{b}[a]_{kj} = b[a]_{kj} + \sum_{n=1}^M [a = a_n^0] \bm{\hat{r}}_{nk}^0\bm{\hat{r}}_{nk}^1$.
\end{mdframed}

\subsection*{E.2 Empirical prior, update equation for $\boldmath{B}$}

In the previous section, we showed that the optimal variational posterior can be written as:
\begin{align}
	\ln Q^*(\bm{B}) &= \ln \prod_{a=1}^A \prod_{j=1}^K \prod_{k=1}^K \bm{B}[a]_{kj}^{(b[a]_{kj} - 1 + \sum_{n=1}^M [a = a_n^0] \bm{\hat{r}}_{nk}^0\bm{\hat{r}}_{nk}^1 )} + c,
\end{align}
which is a Dirichlet distribution with parameters: $\hat{b}[a]_{kj} = b[a]_{kj} + \sum_{n=1}^M [a = a_n^0] \bm{\hat{r}}_{nk}^0\bm{\hat{r}}_{nk}^1$. In this section, we re-arrange this result to obtain an update equation for the empirical prior parameters, and another update equation that expresses the variational posterior parameters as a function of the empirical prior parameters. More precisely, we show that the parameters of $E^*(\bm{B})$ should be computed as follows:
\begin{align}
	\bar{b}[a]_{kj} = b[a]_{kj} + \sum_{n \in M^{'}} [a = a_n^0] \bm{\hat{r}}_{nk}^0 \bm{\hat{r}}_{nj}^1.
\end{align}
Similarly, the parameters of $Q^*(\bm{B})$ should be computed as follows:
\begin{align}
	\hat{b}[a]_{kj} = \bar{b}[a]_{kj} + \sum_{n \in M^{''}} [a = a_n^0] \bm{\hat{r}}_{nk}^0 \bm{\hat{r}}_{nj}^1.
\end{align}
\begin{mdframed}[style=proof]
	Remember that we want to forget the data points indexed by $M^{'}$ and keep the data points indexed by $M^{''}$. Importantly, since $M^{'} \cup M^{''} = \{1, ..., M\}$, we can re-arrange the update equation for the variational parameters as follows:
	\begin{align}
		\hat{b}[a]_{kj} &= b[a]_{kj} + \sum_{n=1}^M [a = a_n^0] \bm{\hat{r}}_{nk}^0\bm{\hat{r}}_{nk}^1 \\
		&= \underbrace{b[a]_{kj} + \sum_{n \in M^{'}} [a = a_n^0] \bm{\hat{r}}_{nk}^0 \bm{\hat{r}}_{nj}^1}_{\bar{b}[a]_{kj}} + \sum_{n \in M^{''}} [a = a_n^0] \bm{\hat{r}}_{nk}^0 \bm{\hat{r}}_{nj}^1,
	\end{align}
	where because the parameters of the empirical prior should take into account the information of the data points indexed by $M^{'}$, we have:
	\begin{align}
		\bar{b}[a]_{kj} = b[a]_{kj} + \sum_{n \in M^{'}} [a = a_n^0] \bm{\hat{r}}_{nk}^0 \bm{\hat{r}}_{nj}^1,
	\end{align}
	and the parameters of the variational posterior can be written as:
	\begin{align}
		\hat{b}[a]_{kj} = \bar{b}[a]_{kj} + \sum_{n \in M^{''}} [a = a_n^0] \bm{\hat{r}}_{nk}^0 \bm{\hat{r}}_{nj}^1.
	\end{align}
\end{mdframed}

\newpage
\section*{Appendix F: Q-learning}

In this appendix, we prove the following relationships:
\begin{align}
	q_{*}(a, s) &= \sum_{s_{t+1}, r_{t+1}} P(s_{t+1}, r_{t+1} | s_t, a_t) \big[ r_{t + 1} + \gamma v_{*}(s_{t+1}) \big] \\
	&= \sum_{s_{t+1}, r_{t+1}} P(s_{t+1}, r_{t+1} | s_t, a_t) \big[ r_{t + 1} + \gamma \max_{a_{t+1}} q_{*}(a_{t+1}, s_{t+1}) \big].
\end{align}
\begin{mdframed}[style=proof]
Let us begin with the definition of $q_{*}(a, s)$ and $q_{\pi}(a, s)$:
\begin{align}
	q_{*}(a, s) &= \max_{\pi} q_{\pi}(a, s) \\
	&= \max_{\pi} \mathbb{E}_\pi [G_t \, | \, a_t = a, s_t = s],
\end{align}
then using the definition of the return $G_t$ from Equation \eqref{eq:return_definit}, we obtain:
\begin{align}
	q_{*}(a, s) &= \max_{\pi} \mathbb{E}_\pi \Bigg[\sum_{\tau = t + 1}^{T} \gamma^{\tau - t - 1} r_{\tau} \, \Bigg| \, a_t = a, s_t = s\Bigg].
\end{align}
Re-arranging the elements in the summation, we get:
\begin{align}
	q_{*}(a, s) &= \max_{\pi} \mathbb{E}_\pi \Bigg[r_{t + 1} + \gamma \sum_{\tau = t + 2}^{T} \gamma^{\tau - t - 2} r_{\tau} \, \Bigg| \, a_t = a, s_t = s\Bigg] \\
	&= \max_{\pi} \mathbb{E}_\pi [r_{t + 1} + \gamma G_{t+1} \, | \, a_t = a, s_t = s],
\end{align}
where we used the return $G_{t+1}$ definition backward. Next, we extract a factor from the expectation to obtain:
\begin{align}
	q_{*}(a, s) &= \max_{\pi} \sum_{s_{t+1}, r_{t+1}} P(s_{t+1}, r_{t+1} | s_t, a_t) \Bigg[ \mathbb{E}_\pi [r_{t + 1} + \gamma G_{t+1} \, | \, s_{t+1}] \Bigg],
\end{align}
which can be re-arranged as follow, by realizing that $r_{t + 1}$ is a constant w.r.t. the $\mathbb{E}_\pi [\, \bigcdot \, | \, s_{t+1}]$ and does not depend on $\pi$:
\begin{align}
	q_{*}(a, s) &= \sum_{s_{t+1}, r_{t+1}} P(s_{t+1}, r_{t+1} | s_t, a_t) \Bigg[ r_{t + 1} + \gamma \max_{\pi} \mathbb{E}_\pi [G_{t+1} \, | \, s_{t+1}] \Bigg]. \label{before_first_relationship}
\end{align}
Importantly, recognizing the definition of $v_{*}(s_{t+1})$, we get our first relationship:
\begin{align}
	q_{*}(a, s) &= \sum_{s_{t+1}, r_{t+1}} P(s_{t+1}, r_{t+1} | s_t, a_t) \big[ r_{t + 1} + \gamma v_{*}(s_{t+1}) \big]. \\
\end{align}
Restarting from \eqref{before_first_relationship}, we extract another factor from the expectation leading to:
\begin{align}
	q_{*}(a, s) &= \sum_{s_{t+1}, r_{t+1}} P(s_{t+1}, r_{t+1} | s_t, a_t) \Bigg[ r_{t + 1} + \gamma \max_{\pi} \sum_{a_{t+1}} \pi(a_{t+1}|s_{t+1}) \Big[\mathbb{E}_\pi [G_{t+1} \, | \, a_{t+1}, s_{t+1}] \Big] \Bigg],
\end{align}
where we recognize the definition of $q_{\pi}(a_{t+1}, s_{t+1})$:
\begin{align}
	q_{*}(a, s) &= \sum_{s_{t+1}, r_{t+1}} P(s_{t+1}, r_{t+1} | s_t, a_t) \Bigg[ r_{t + 1} + \gamma \max_{\pi} \underbrace{\sum_{a_{t+1}} \pi(a_{t+1}|s_{t+1}) q_{\pi}(a_{t+1}, s_{t+1})}_{f(\pi)} \Bigg]. \label{eq:before_condition_eq}
\end{align}
Focusing on the expression within the maximum operator, we see that $f(\pi)$ can only be maximized if:
\begin{align}
	 \forall a_{t+1},\,\, \Big( q_{\pi}(a_{t+1}, s_{t+1}) \neq \max_{a} q_{\pi}(a, s_{t+1})\Big) \Rightarrow \pi(a_{t+1}|s_{t+1}) = 0. \label{eq:maximum_condition}
\end{align}
Indeed, if there exists an action $a_{t+1}$ such that $q_{\pi}(a_{t+1}, s_{t+1}) \neq \max_a q_{\pi}(a, s_{t+1})$ and $\pi(a_{t+1}|s_{t+1}) \neq 0$, then $f(\pi)$ can be increased by reducing $\pi(a_{t+1}|s_{t+1})$ to zero and increasing $\pi(a_{t+1}^{'}|s_{t+1})$ for an action $a_{t+1}^{'}$ where $q_{\pi}(a_{t+1}^{'}, s_{t+1}) = \max_a q_{\pi}(a, s_{t+1})$. The condition describes in \eqref{eq:maximum_condition} together with \eqref{eq:before_condition_eq}, implies that:
\begin{align}
	q_{*}(a, s) &= \sum_{s_{t+1}, r_{t+1}} P(s_{t+1}, r_{t+1} | s_t, a_t) \Bigg[ r_{t + 1} + \gamma \max_{\pi} \max_{a_{t+1}} q_{\pi}(a_{t+1}, s_{t+1}) \Bigg],
\end{align}
where by swapping the maximum operators, we get:
\begin{align}
	q_{*}(a, s) &= \sum_{s_{t+1}, r_{t+1}} P(s_{t+1}, r_{t+1} | s_t, a_t) \Bigg[ r_{t + 1} + \gamma \max_{a_{t+1}} \max_{\pi} q_{\pi}(a_{t+1}, s_{t+1}) \Bigg],
\end{align}
and by recognizing the definition of $q_{*}(a_{t+1}, s_{t+1})$, we get the second relationship:
\begin{align}
	q_{*}(a, s) &= \sum_{s_{t+1}, r_{t+1}} P(s_{t+1}, r_{t+1} | s_t, a_t) \big[ r_{t + 1} + \gamma \max_{a_{t+1}} q_{*}(a_{t+1}, s_{t+1}) \big].
\end{align}
\end{mdframed}

\end{document}